\documentclass{article}

\usepackage[main, final]{neurips_2026}

\usepackage[utf8]{inputenc} 
\usepackage[T1]{fontenc}    
\usepackage{hyperref}       
\usepackage{url}            
\usepackage{booktabs}       
\usepackage{amsfonts}       
\usepackage{nicefrac}       
\usepackage{microtype}      
\usepackage{xcolor}         
\usepackage{outlines}
\usepackage{tikz}
\usepackage{float}
\usepackage{algorithm}
\usepackage[noend]{algpseudocode}
\usepackage{todonotes}
\usepackage{amsmath}
\usepackage{mathtools}
\usepackage{subcaption}
\newcommand{\galois}[2]{\mathrel{\mathchoice%
  {\begin{array}{@{}c@{}}\xleftharpoonup{\;\;\;\;#2\;\;\;\;}\\[-0.75em]\xrightharpoondown[\;\;\;\;#1\;\;\;\;]{}\end{array}}%
  {\begin{array}{@{}c@{}}\xleftharpoonup{\;#2\;}\\[-0.75em]\xrightharpoondown[\;#1\;]{}\end{array}}%
  {\begin{array}{@{}c@{}}\xleftharpoonup{\;#2\;}\\[-0.75em]\xrightharpoondown[\;#1\;]{}\end{array}}%
  {\begin{array}{@{}c@{}}\xleftharpoonup{\;#2\;}\\[-0.75em]\xrightharpoondown[\;#1\;]{}\end{array}}%
}}
\usetikzlibrary{matrix, arrows.meta, positioning, fit, calc, shapes.geometric, backgrounds, decorations.pathreplacing}

\title{Lattice Deduction Transformers}

\author{%
  Liam Davis \\
  Amherst College \\
  \texttt{ljdavis27@amherst.edu} \\
  \And 
  Leopold Haller \\
  Axiom \\
  \texttt{leo@axiommath.ai} \\
  \And 
  Alberto Alfarano \\
  Axiom \\
  \texttt{alberto@axiommath.ai} \\
  \And 
  Mark Santolucito \\
  Barnard College, Columbia University \\
  \texttt{msantolu@barnard.edu}
}

\begin{document}

\maketitle

\begin{abstract}
We introduce the Lattice Deduction Transformer (LDT), a recurrent
transformer that approximates logically sound deduction by
projecting its latent state through a lattice between forward passes.
We train on-policy in a process that mirrors deduction in a
search-based constraint solver and supervise training via a
domain-agnostic approximation of the
set of solution candidates based on abstract interpretation. An $800$K-parameter
LDT outperforms prior small recurrent reasoners on Sudoku-Extreme with under
$2\%$ of their estimated training FLOPs, solving $99.999\%$ of the
$422{,}786$ test puzzles. The model is nearly always sound, in that it gives a wrong answer for $4$ of
$1.27$M puzzle evaluations, and otherwise returns a correct answer or
abstains. A $1.8$M-parameter LDT variant reaches $99.8\%$ accuracy on Maze-Hard.
  Our code is available at \href{https://github.com/lcrh/lattice-deduction-transformers}{\textcolor{blue}{github.com/lcrh/lattice-deduction-transformers}}.
\end{abstract}

\begin{figure}[htbp]
  \centering
  \begin{subfigure}[b]{0.575\linewidth}
    \centering
    \resizebox{\linewidth}{!}{\begin{tikzpicture}[
    font=\large,
    block/.style={rectangle, rounded corners=4pt, draw, thick, text centered,
                  minimum width=4cm, minimum height=0.75cm},
    attnblock/.style={block, fill=orange!30,    draw=orange!75!black},
    normblock/.style={block, fill=yellow!28,    draw=yellow!55!brown},
    mlpblock/.style={block, fill=green!22,      draw=green!55!black},
    latticeblock/.style={block, fill=blue!10,   draw=blue!55},
    branchblock/.style={rectangle, rounded corners=4pt, draw, thick,
                        fill=red!8, draw=red!60!black,
                        minimum width=1.5cm, minimum height=0.6cm},
    tfmbox/.style={rectangle, rounded corners=8pt, draw, thick,
                   fill=black!2, draw=black!40, inner sep=0.45cm},
    loopbox/.style={rectangle, rounded corners=10pt, draw=black!45, thick, dashed,
                    inner sep=0.55cm},
    arrow/.style={-{Stealth[length=2.5mm]}, thick, black!70},
    iter/.style={-{Stealth[length=3mm]}, very thick, black!75},
    symb/.style={font=\Large, text=black!70},
    mathsymb/.style={font=\huge}
]

\node[normblock] (n1)   {LayerNorm};
\node[attnblock] (attn) [above=0.4cm of n1]    {Multi-Head Attention};
\node[normblock] (n2)   [above=0.4cm of attn]  {LayerNorm};
\node[mlpblock]  (mlp)  [above=0.4cm of n2]    {Feed Forward};

\begin{scope}[on background layer]
  \node[tfmbox, fit=(n1)(attn)(n2)(mlp)] (tfm) {};
\end{scope}

\node[symb, right=0.05cm of tfm.east, anchor=west] (label_4x) {$4\times$};

\node[latticeblock] (lat_in)  [below=1.1cm of tfm] {Lattice encoding \textcolor{red!70!black}{$x$}};
\node[latticeblock] (lat_out) [above=1.1cm of tfm] {Lattice encoding \textcolor{red!70!black}{$x'$}};

\draw[arrow] (lat_in)    -- (tfm.south);
\draw[arrow] (n1)        -- (attn);
\draw[arrow] (attn)      -- (n2);
\draw[arrow] (n2)        -- (mlp);
\draw[arrow] (tfm.north) -- (lat_out);

\begin{scope}[on background layer]
  \node[loopbox, fit=(tfm)(label_4x), inner sep=0.25cm] (loop) {};
\end{scope}

\pgfmathsetmacro{\StapleDx}{1.0}   
\pgfmathsetmacro{\StapleDy}{0.4}   
\coordinate (s_top_in)  at ($(loop.north)+(0, \StapleDy cm)$);
\coordinate (s_top_out) at ($(loop.north -| loop.east)+(\StapleDx cm, \StapleDy cm)$);
\coordinate (s_bot_out) at ($(loop.south -| loop.east)+(\StapleDx cm, -\StapleDy cm)$);
\coordinate (s_bot_in)  at ($(loop.south)+(0, -\StapleDy cm)$);
\draw[arrow]
  (loop.north) -- (s_top_in)
  -- (s_top_out)
  -- (s_bot_out)
  -- (s_bot_in)
  -- (loop.south);
\node[symb, anchor=west] (label_16x)
  at ($(s_top_out)!0.5!(s_bot_out)+(0.15cm,0)$) {$16\times$};

\coordinate (iter_top) at ($(lat_out.east -| loop.east)+(3.2cm,0)$);
\coordinate (iter_bot) at ($(lat_in.east  -| loop.east)+(3.2cm,0)$);
\coordinate (iter_mid) at ($(iter_top)!0.5!(iter_bot)$);
\draw[iter]
  (lat_out.east) -- (iter_top)
  -- (iter_bot)
  -- (lat_in.east);
\node[branchblock] (branch) at (iter_mid) {Branch};

\draw[arrow] ($(lat_in.south)+(0,-0.6cm)$) -- (lat_in.south);
\draw[arrow] (lat_out.north) -- ($(lat_out.north)+(0,0.6cm)$);

\pgfmathsetmacro{\HasseDx}{6.6}
\coordinate (h_top_right) at ($(lat_out.north -| loop.west)+(-0.8cm,0)$);
\coordinate (h_bot_right) at ($(lat_in.south  -| loop.west)+(-0.8cm,0)$);
\coordinate (h_top_left)  at ($(h_top_right)+(-{\HasseDx}cm,0)$);
\coordinate (h_bot_left)  at ($(h_bot_right)+(-{\HasseDx}cm,0)$);

\coordinate (h_center)  at ($(h_bot_left)!0.5!(h_top_right)$);

\coordinate (h_top_pos) at ($(h_top_right)!0.08!(h_bot_right)$);
\coordinate (h_bot_pos) at ($(h_bot_right)!0.08!(h_top_right)$);

\tikzset{lattnode/.style={circle, draw=black!45, thin, minimum size=7.4mm,
                          inner sep=0.5pt, fill=white, font=\small},
         statenode/.style={lattnode, draw=red!70!black, very thick, fill=red!12},
         edge/.style={black!22, line width=0.6pt}}

\node[lattnode] (top_sym) at ($(h_center |- h_top_pos)$) {1234};
\node[lattnode] (bot_sym) at ($(h_center |- h_bot_pos)$) {$\emptyset$};

\coordinate (L1y) at ($(top_sym)!0.25!(bot_sym)$);
\coordinate (L2y) at ($(top_sym)!0.5!(bot_sym)$);
\coordinate (L3y) at ($(top_sym)!0.75!(bot_sym)$);

\pgfmathsetmacro{\LXa}{1.32}    
\pgfmathsetmacro{\LXb}{1.32}    

\pgfmathsetmacro{\DyCurve}{0.13}
\pgfmathsetmacro{\DyStack}{0.45}

\node[lattnode] (l1a) at ($(L1y)+(-1.5*\LXa cm, -\DyCurve cm)$) {123};
\node[lattnode] (l1b) at ($(L1y)+(-0.5*\LXa cm,  \DyCurve cm)$) {124};
\node[lattnode] (l1c) at ($(L1y)+( 0.5*\LXa cm,  \DyCurve cm)$) {134};
\node[statenode] (l1d) at ($(L1y)+( 1.5*\LXa cm, -\DyCurve cm)$) {234};

\node[lattnode] (l2a) at ($(L2y)+(-2.5*\LXb cm, 0)$)            {12};
\node[lattnode] (l2b) at ($(L2y)+(-1.5*\LXb cm, 0)$)            {13};
\node[lattnode] (l2c) at ($(L2y)+( 0,            \DyStack cm)$) {14};
\node[lattnode] (l2d) at ($(L2y)+( 0,           -\DyStack cm)$) {23};
\node[lattnode] (l2e) at ($(L2y)+( 1.5*\LXb cm, 0)$)            {24};
\node[lattnode] (l2f) at ($(L2y)+( 2.5*\LXb cm, 0)$)            {34};

\node[lattnode] (l3a) at ($(L3y)+(-1.5*\LXa cm,  \DyCurve cm)$) {1};
\node[lattnode] (l3b) at ($(L3y)+(-0.5*\LXa cm, -\DyCurve cm)$) {2};
\node[statenode, fill=red!30] (l3c) at ($(L3y)+( 0.5*\LXa cm, -\DyCurve cm)$) {3};
\node[lattnode] (l3d) at ($(L3y)+( 1.5*\LXa cm,  \DyCurve cm)$) {4};

\draw[edge] (top_sym) -- (l1a);
\draw[edge] (top_sym) -- (l1b);
\draw[edge] (top_sym) -- (l1c);
\draw[edge] (top_sym) -- (l1d);

\draw[edge] (l1a) -- (l2a);   
\draw[edge] (l1a) -- (l2b);   
\draw[edge] (l1a) -- (l2d);   
\draw[edge] (l1b) -- (l2a);   
\draw[edge] (l1b) -- (l2c);   
\draw[edge] (l1b) -- (l2e);   
\draw[edge] (l1c) -- (l2b);   
\draw[edge] (l1c) -- (l2c);   
\draw[edge] (l1c) -- (l2f);   
\draw[edge] (l1d) -- (l2d);   
\draw[edge] (l1d) -- (l2e);   
\draw[edge] (l1d) -- (l2f);   

\draw[edge] (l2a) -- (l3a);   
\draw[edge] (l2a) -- (l3b);   
\draw[edge] (l2b) -- (l3a);   
\draw[edge] (l2b) -- (l3c);   
\draw[edge] (l2c) -- (l3a);   
\draw[edge] (l2c) -- (l3d);   
\draw[edge] (l2d) -- (l3b);   
\draw[edge] (l2d) -- (l3c);   
\draw[edge] (l2e) -- (l3b);   
\draw[edge] (l2e) -- (l3d);   
\draw[edge] (l2f) -- (l3c);   
\draw[edge] (l2f) -- (l3d);   

\draw[edge] (l3a) -- (bot_sym);
\draw[edge] (l3b) -- (bot_sym);
\draw[edge] (l3c) -- (bot_sym);
\draw[edge] (l3d) -- (bot_sym);

\draw[-{Stealth[length=3mm]}, line width=1.4pt, red!70!black]
  (l1d.south west) to[bend left=12] (l3c.north east);
\node[font=\Large, text=red!70!black, anchor=west] at ($(l1d.east)+(0.05,0.1)$) {$x$};
\node[font=\Large, text=red!70!black, anchor=east] at ($(l3c.west)+(-0.02,-0.18)$) {$x'$};

\end{tikzpicture}}\\[-0.065cm]
    {\fontfamily{DejaVuSans-TLF}\fontsize{5.76}{7}\selectfont
      Deduction on a lattice of candidate sets, alternated with branching}\\[0.245cm]
    \subcaption{}
  \end{subfigure}\hfill
  \begin{subfigure}[b]{0.41\linewidth}
    \centering
    \includegraphics[height=4.6cm]{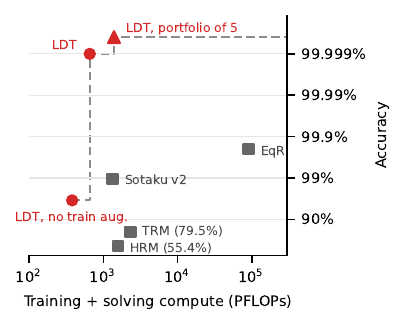}
    \subcaption{}
  \end{subfigure}
  \caption{(a) The Lattice Deduction Transformer (LDT) performs deduction on
    a lattice using a recurrent transformer which is alternated with search
    via stochastic branching. (b) Sudoku-Extreme accuracy against the
    estimated FLOPs of training plus solving the $422{,}786$-puzzle test set.}
  \label{fig:ldt-overview}
\end{figure}

\section{Introduction}
\label{sec:intro}

Reasoning in large language models has become largely synonymous with the
chain-of-thought paradigm~\cite{wei2022chain} and its test-time-compute
extensions~\cite{wang2023selfconsistency, snell2025ttc}, in which the model
generates intermediate reasoning tokens before producing a final answer.
In contrast, automated logical reasoning procedures operate by
iterative refinement of a partial information state.
While the latent representations of neural networks can exhibit useful
algebraic regularities, such as the vector arithmetic of word embeddings
(``king'' $-$ ``man'' $+$ ``woman'' $\approx$
``queen''~\cite{mikolov2013linguistic}), partial information states require
operations that approximate logical primitives such as conjunction and
disjunction. Lattices used in abstract interpretation provide these
primitives as greatest lower bounds (meet, $\sqcap$) and least upper bounds
(join, $\sqcup$). For example, a lattice-based latent representation could
represent ``king'' $\sqcup$ ``queen'' $\approx$ ``royalty'' and
``royalty'' $\sqcap$ ``woman'' $\approx$ ``queen''.
We train a recurrent transformer whose output is an element of
such a lattice, so that symbolic reasoning procedures can operate directly
on its state.

We consider reasoning from the perspective of \textit{abstract
interpretation}~\cite{CousotCousot77-1, CousotCousot79}, which models
information states as elements of a lattice with a top element $\top$
representing no information (every solution still possible) and a
bottom element $\bot$ representing inconsistency (no solution
remains). Deduction is a process that descends from $\top$ toward more
informative states; a state that collapses to $\bot$ signals that no
valid solution remains. Satisfiability procedures such as
DPLL~\cite{davis1960dp, davis1962dpll}, CDCL~\cite{marquessilva2009cdcl} and
St{\aa}lmarck's procedure~\cite{10.1007/3-540-49519-3_7,
leonov2025optimizationsstaalmarckprocedure} can be viewed as procedures that
combine lattice-based deduction with search~\cite{dsilva2013acdcl,
thakurReps2012stalmarck}. Their deduction steps are sound: they may fail to
derive a true fact, but never derive a false one. LDT replaces these
hand-written steps with a learned transformer and does not inherit this
guarantee; for LDT, soundness is an empirical property that we measure.

We use a recurrent transformer whose latent state is
projected to and from a lattice representation between forward
passes; we call this the Lattice Deduction Transformer (LDT). The
interpretability of the abstract domain lets us leverage symbolic
search processes at both training and inference time. On hard problems with logical
structure, LDT exceeds comparable reasoning transformers with smaller models
and less training. On some domains it further achieves empirical soundness
given a sufficient training budget: the search either returns a correct
answer or abstains.

We make four contributions. First, an architecture that projects a
recurrent transformer's latent state onto an abstract lattice between
forward passes, turning each forward pass into an approximately sound deduction step.
Second, an on-policy training procedure where lattice states evolve
under the model's own forward pass, keeping training states
in-distribution with search states encountered during inference. Third,
the \emph{alpha operator}, which aggregates the valid solutions still
consistent with the current lattice state. This places our training in
a regime we are not aware of in prior work: on-policy yet still
supervised by a domain-agnostic, state-dependent target. Fourth, an experimental
characterization of the train/test compute trade-off in learned, approximately sound
inference, showing that additional training shortens inference search
by orders of magnitude.

The rest of the paper is organized as follows: Section~\ref{sec:related-work}
reviews related work. Section~\ref{sec:method} introduces the lattices LDT
reasons over and details its architecture, training and inference. Section~\ref{sec:eval} provides an evaluation of the Lattice
Deduction Transformer against comparable reasoning transformers. Finally, we conclude and discuss future directions in
section~\ref{sec:conclusion}.

\section{Related Work}
\label{sec:related-work}

Our approach draws from research on small recurrent reasoning models. Adaptive Computation Time~\cite{graves2016act}
introduced variable-iteration neural computation for recurrent
networks, later adapted to transformers by Universal
Transformers~\cite{dehghani2018ut}. The Hierarchical
Reasoning Model~\cite{wang2025hierarchicalreasoningmodel} uses a
dual-loop recurrent architecture, and the Tiny Recursive
Model~\cite{jolicoeurmartineau2025morerecursivereasoningtiny} shows 
that a 2-layer network with recursion can outperform it with a smaller
model. Equilibrium Reasoners~\cite{huang2026equilibrium} train such
recursive models to converge to attractors and scale their inference in depth
and breadth. We build directly on Sotaku~\cite{LouSotaku2026}, a recurrent
transformer specialized for Sudoku.

LDT is a neuro-symbolic method, combining a
neural network with the lattice representation that search-based
constraint solvers use to track partial information.
SATNet~\cite{wang2019satnet} and Learning Modulo
Theories~\cite{yang2023smtlayer} embed MAX-SAT and SMT solvers as
network layers, preserving a hard neural/symbolic boundary. Closer to
our setting, machine learning has been applied throughout the
abstract-interpretation pipeline: Neural Abstract
Interpretation~\cite{gomber2025nai} trains sound differentiable
abstract transformers, LAIT~\cite{he2020lait} learns a neural policy
that prunes redundant constraints, and SAIL~\cite{gu2026sail}
synthesizes sound transformers via LLMs with SMT validation. Earlier
work targets the control loop instead, selecting among sound
transformers via RL~\cite{singh2018rlanalysis} or learning widening
thresholds~\cite{cha2016widening}.

D3PM~\cite{austin2021d3pm}, the discrete-diffusion generalization of
masked language modeling~\cite{devlin2019bert}, iteratively commits
information about partially-masked sequences; we recast this process
as a fixed-point computation over a lattice. D3PM in fact admits an arbitrary token-transition
matrix $Q$, which can be instantiated over tokens that themselves
form a lattice -- closely related in spirit to our setup, though we
do not develop our method as a diffusion procedure. Several recent diffusion variants push in this
direction. For example,
ReMDM~\cite{wang2025remdm} reintroduces backtracking by allowing
already-generated tokens to be remasked at inference time,
Prime~\cite{chao2025prime} relaxes the masked/unmasked dichotomy by
letting tokens occupy intermediate states between $\top$ and a
concrete value, and HDLM~\cite{zhou2025hdlm} organizes the vocabulary
into a refinement hierarchy whose forward process abstracts upward
and reverse process refines downward. Each can be read as moving
along a particular partial-information ordering; in our framing, that
ordering is the lattice.

\section{Methodology}
\label{sec:method}

\subsection{Abstract Interpretation}
\label{subsec:domains}

Following abstract interpretation~\cite{CousotCousot77-1, CousotCousot79}
(Appendix~\ref{app:ai-tutorial}),
LDT represents what is known about a solution as an element of a
\emph{lattice}: a partially ordered set in which every pair of elements $a,
b$ has a greatest lower bound (meet) $a \sqcap b$ and a least upper bound
(join) $a \sqcup b$. The order encodes precision: lower elements are more
informative, the top element $\top$ carries no information, and the bottom
element $\bot$ denotes inconsistency.

A solved Sudoku is a grid $g : \{1,\ldots,9\}^2 \to \{1,\ldots,9\}$. The
sets of possible solutions form the powerset lattice $C = \mathcal{P}(G)$
over the set $G$ of all grids, ordered by inclusion: an element is the set of
grids still considered possible, $\top = G$ is our initial knowledge state,
and $\bot = \emptyset$ means the puzzle has no solution. We call $C$ the \emph{concrete domain}. Since it is
expensive to represent, we work instead in the following \emph{abstract
domain}, which keeps for each cell the digits still considered possible and
drops all correlations between cells: $A = \{1,\ldots,9\}^2 \to
\mathcal{P}(\{1,\ldots,9\})$, ordered pointwise ($a \sqsubseteq b$ iff
$a(c) \subseteq b(c)$ for every cell $c$), with meet and join the pointwise
intersection and union. $\top$ allows every digit in every cell, and a cell
with no candidates means $\bot$. We call $A$ the \emph{grid powerset
lattice}. The two domains are linked by an abstraction function $\alpha : C \to A$,
which maps a set of grids to the digits each cell takes in some grid,
$\alpha(G')(c) = \{g(c) \mid g \in G'\}$, and a concretization function $\gamma : A
\to C$, which maps candidate sets to the grids they allow, $\gamma(a) = \{g
\in G \mid \forall c.\ g(c) \in a(c)\}$. They form a Galois connection
(Figure~\ref{fig:roundtrip}), so any property established on $A$ holds on
$C$.

\begin{figure}[htbp]
  \centering
\tikzset{
  rt/.style={font=\footnotesize},
  rtc/.style={draw, minimum size=0.46cm, inner sep=0pt, font=\footnotesize},
  rta/.style={draw, minimum width=0.92cm, minimum height=0.46cm, inner sep=0pt,
              font=\footnotesize},
  rtb/.style={draw, minimum size=0.46cm, inner sep=0pt, font=\footnotesize,
              fill=red!8},
  rtarr/.style={-{Stealth[length=4pt]}, thick},
}
\begin{tikzpicture}[rt]
\matrix (gA) [matrix of nodes, nodes={rtc}, row sep=-\pgflinewidth,
              column sep=-\pgflinewidth] at (0, 0.58) {1 & 1 \\ 2 & 2 \\};
\matrix (gB) [matrix of nodes, nodes={rtc}, row sep=-\pgflinewidth,
              column sep=-\pgflinewidth] at (0, -0.58) {1 & 2 \\ 2 & 1 \\};
\node at (0, 1.35) {$S$};
\matrix (abs) [matrix of nodes, nodes={rta}, row sep=-\pgflinewidth,
               column sep=-\pgflinewidth] at (2.45, 0)
  {$\{1\}$ & $\{1,2\}$ \\ $\{2\}$ & $\{1,2\}$ \\};
\node at (2.45, 1.35) {$\alpha(S)$};
\matrix (h1) [matrix of nodes, nodes={rtc}, row sep=-\pgflinewidth,
              column sep=-\pgflinewidth] at (4.55, 0.58) {1 & 1 \\ 2 & 1 \\};
\matrix (h2) [matrix of nodes, nodes={rtc}, row sep=-\pgflinewidth,
              column sep=-\pgflinewidth] at (5.5, 0.58) {1 & 1 \\ 2 & 2 \\};
\matrix (h3) [matrix of nodes, nodes={rtc}, row sep=-\pgflinewidth,
              column sep=-\pgflinewidth] at (4.55, -0.58) {1 & 2 \\ 2 & 1 \\};
\matrix (h4) [matrix of nodes, nodes={rtc}, row sep=-\pgflinewidth,
              column sep=-\pgflinewidth] at (5.5, -0.58) {1 & 2 \\ 2 & 2 \\};
\node at (5.025, 1.35) {$\gamma(\alpha(S))$};
\draw[rtarr] (0.65, 0) -- node[above] {$\alpha$} (abs.west);
\draw[rtarr] (abs.east) -- node[above] {$\gamma$} (3.95, 0);
\end{tikzpicture}
\hspace{0.5cm}
\begin{tikzpicture}[rt]
\path (0, -1.05) -- (0, 1.55);   
\matrix (a) [matrix of nodes, nodes={rta}, row sep=-\pgflinewidth,
             column sep=-\pgflinewidth] at (0, 0)
  {$\{1\}$ & $\{1,2\}$ \\ $\{1\}$ & $\{1,2\}$ \\};
\node at (0, 0.75) {$a$};
\node[draw, minimum size=0.92cm] (empty) at (2.2, 0) {$\emptyset$};
\node at (2.2, 0.75) {$\gamma(a)$};
\matrix (bot) [matrix of nodes, nodes={rtb}, row sep=-\pgflinewidth,
               column sep=-\pgflinewidth] at (3.9, 0)
  {$\emptyset$ & $\emptyset$ \\ $\emptyset$ & $\emptyset$ \\};
\node at (3.9, 0.75) {$\alpha(\gamma(a)) = \bot$};
\draw[rtarr] (a.east) -- node[above] {$\gamma$} (empty.west);
\draw[rtarr] (empty.east) -- node[above] {$\alpha$} (bot.west);
\end{tikzpicture}
  \caption{$\alpha$ and $\gamma$ form a Galois connection with
    $\gamma \circ \alpha \supseteq \mathrm{id}_C$ and
    $\alpha \circ \gamma \sqsubseteq \mathrm{id}_A$.}
  \label{fig:roundtrip}
\end{figure}

For a puzzle $p$ with solutions $\|p\| \subseteq G$, the most precise sound
deduction operator on $A$ is
\[
\mathit{ded}_p(a) \;=\; \alpha\bigl(\gamma(a) \cap \|p\|\bigr),
\]
which keeps a candidate only if some solution consistent with $a$ uses it.
Sudoku rules such as naked and hidden singles are sound but weaker deduction
operators: each step removes candidates without ruling out a solution. LDT
trains a transformer to approximate $\mathit{ded}_p$ from solution samples. The learned operator is not guaranteed to be
sound, so soundness becomes an empirical property, and how much to rely on
training versus search becomes a trade-off, which we measure in
Section~\ref{subsec:Sudoku-extreme}.

The construction applies to any problem whose solutions are fixed-length
token strings over a vocabulary $V$: with solution space $S =
\{1,\ldots,k\} \to V$, we take $C = \mathcal{P}(S)$ and $A = \{1,\ldots,k\}
\to \mathcal{P}(V)$, with $\alpha$, $\gamma$ and $\mathit{ded}_p$ as above,
positions in place of cells. We assume only that samples from $\|p\|$ are
available for each training instance.

\subsection{Architecture}
\label{subsec:architecture}

\paragraph{Lattice encoding}
We represent the grid powerset lattice as a multi-hot encoding: $|V|$
binary sigmoids per cell (so $9 \times 9 \times 9 = 729$ sigmoids for
$9 \times 9$ Sudoku; see Figure~\ref{fig:lattice-representation}). Each sigmoid encodes the
confidence that its candidate is still alive. Solutions
$y$ are one-hot (one bit alive per cell), and the abstraction $\alpha$
over a set of solutions reduces to a bit-level OR.

We give $\bot$ a dual representation. The first is implicit in the candidate
encoding: a cell with an empty candidate set is the natural representation of
$\bot$ in the powerset lattice. The second is an explicit binary sigmoid
attached to a distinguished CLS token, which we train to fire on any
unsatisfiable state. Carrying
separate representations lets us train the two heads with different losses,
discussed below.

\begin{figure}[htbp]
\centering
\resizebox{\linewidth}{!}{
\begin{tikzpicture}[
  font=\small,
  cell/.style={draw=black!45, minimum width=0.36cm, minimum height=0.36cm,
               inner sep=0pt, font=\scriptsize},
  dead/.style={cell, draw=black!20, text=black!25},
  alive/.style={cell, fill=blue!22},
  cls/.style={cell, fill=orange!30},
  head/.style={font=\footnotesize, anchor=base},
  rowlabel/.style={font=\scriptsize, anchor=east, text=black!70},
  op/.style={-{Latex[length=2.2mm, width=1.6mm]}, line width=0.9pt, black!60},
  oplabel/.style={font=\footnotesize, align=center},
]
\def\w{0.36}
\definecolor{cm0}{HTML}{0D0887}\definecolor{cm1}{HTML}{4C02A1}
\definecolor{cm2}{HTML}{7E03A8}\definecolor{cm3}{HTML}{A92395}
\definecolor{cm4}{HTML}{CC4778}\definecolor{cm5}{HTML}{E66C5C}
\definecolor{cm6}{HTML}{F89540}\definecolor{cm7}{HTML}{FDC527}
\definecolor{cm8}{HTML}{F0F921}
\newcommand{\cmap}[1]{
  \pgfmathtruncatemacro{\seg}{min(7, int(8*(#1)/100))}%
  \pgfmathtruncatemacro{\mix}{8*(#1) - 100*\seg}%
  \pgfmathtruncatemacro{\up}{\seg + 1}%
  \colorlet{confcolor}{cm\up!\mix!cm\seg}}
\newcommand{\lattice}[3]{
  \foreach \d/\v in {#3} {
    \ifnum\v=1 \node[alive] at ({#1 + (\d-1)*\w}, #2) {\d};
    \else \node[dead] at ({#1 + (\d-1)*\w}, #2) {\d}; \fi
  }}
\newcommand{\confidences}[3]{
  \foreach \d/\p in {#3} {
    \ifnum\p<0 \node[dead] at ({#1 + (\d-1)*\w}, #2) {\d};
    \else \cmap{\p}\ifnum\p<60 \def\ink{white}\else \def\ink{black}\fi
      \node[cell, fill=confcolor, text=\ink] at ({#1 + (\d-1)*\w}, #2) {\d}; \fi
  }}
\def\xa{0}\def\xb{4.95}\def\xc{9.95}
\def\ycls{0}\def\yone{-0.62}\def\ytwo{-1.08}\def\ydots{-1.46}\def\yn{-1.86}
\pgfmathsetmacro{\mid}{4*\w}          
\pgfmathsetmacro{\colw}{8*\w}         

\node[rowlabel] at (-0.3, \ycls) {[CLS]};
\node[rowlabel] at (-0.3, \yone) {cell 1};
\node[rowlabel] at (-0.3, \ytwo) {cell 2};
\node[rowlabel] at (-0.3, \yn) {cell $n$};

\node[head] at ({\xa + \mid}, 0.42) {state $x$};
\node[head] at ({\xb + \mid}, 0.42) {confidences $\sigma(c)$, $\sigma(b)$};
\node[head] at ({\xc + \mid}, 0.42) {next state $x'$};

\node[cls] at (\xa, \ycls) {$\bot$};
\lattice{\xa}{\yone}{1/1, 2/0, 3/1, 4/1, 5/0, 6/1, 7/1, 8/0, 9/1}
\lattice{\xa}{\ytwo}{1/0, 2/1, 3/0, 4/0, 5/1, 6/0, 7/0, 8/1, 9/0}
\node[text=black!40] at ({\xa + \mid}, \ydots) {$\vdots$};
\lattice{\xa}{\yn}{1/1, 2/1, 3/0, 4/1, 5/1, 6/1, 7/0, 8/1, 9/1}

\cmap{2}\node[cell, minimum width=0.62cm, fill=confcolor, text=white, anchor=west]
  at ({\xb - \w/2}, \ycls) {0.02};
\confidences{\xb}{\yone}{1/3, 2/-1, 3/90, 4/60, 5/-1, 6/5, 7/80, 8/-1, 9/2}
\confidences{\xb}{\ytwo}{1/-1, 2/4, 3/-1, 4/-1, 5/97, 6/-1, 7/-1, 8/6, 9/-1}
\node[text=black!40] at ({\xb + \mid}, \ydots) {$\vdots$};
\confidences{\xb}{\yn}{1/1, 2/70, 3/-1, 4/50, 5/4, 6/8, 7/-1, 8/90, 9/3}

\def\ybar{-2.42}
\foreach \i in {0,...,49} {
  \cmap{2*\i + 1}
  \fill[confcolor] ({\xb - \w/2 + \i*(9*\w)/50}, \ybar) rectangle ++({(9*\w)/50 + 0.002}, 0.12);
}
\draw[black!45] ({\xb - \w/2}, \ybar) rectangle ({\xb + 8.5*\w}, \ybar + 0.12);
\node[font=\scriptsize, anchor=east] at ({\xb - \w/2 - 0.04}, \ybar + 0.06) {0};
\node[font=\scriptsize, anchor=west] at ({\xb + 8.5*\w + 0.04}, \ybar + 0.06) {1};
\draw[black!70] ({\xb - \w/2 + 0.1*9*\w}, \ybar - 0.05) -- ++(0, 0.22);
\node[font=\scriptsize, anchor=north] at ({\xb - \w/2 + 0.1*9*\w}, \ybar - 0.04) {$\theta_{\text{elim}}$};

\node[cls, fill=orange!12, text=black!45] at (\xc, \ycls) {$\bot$};
\lattice{\xc}{\yone}{1/0, 2/0, 3/1, 4/1, 5/0, 6/0, 7/1, 8/0, 9/0}
\lattice{\xc}{\ytwo}{1/0, 2/0, 3/0, 4/0, 5/1, 6/0, 7/0, 8/0, 9/0}
\node[text=black!40] at ({\xc + \mid}, \ydots) {$\vdots$};
\lattice{\xc}{\yn}{1/0, 2/1, 3/0, 4/1, 5/0, 6/0, 7/0, 8/1, 9/0}

\def\yop{-0.93}
\draw[op] ({\xa + \colw + 0.38}, \yop) -- ({\xb - 0.38}, \yop)
  node[midway, above=2pt, oplabel] {Transformer}
  node[midway, below=2pt, oplabel, font=\scriptsize, text=black!65] {$16$ iterations};
\draw[op] ({\xb + \colw + 0.38}, \yop) -- ({\xc - 0.38}, \yop)
  node[midway, above=2pt, oplabel] {threshold}
  node[midway, below=2pt, oplabel, font=\scriptsize, text=black!65]
    {$\sigma(b) < \theta_{\text{elim}}$};
\end{tikzpicture}}
\caption{One step on the lattice-tensor encoding. The transformer maps the
state $x$ (alive candidates highlighted, eliminated ones greyed) to a
confidence per alive candidate and a conflict confidence on the CLS token
(color scale below). Thresholding removes candidates with $\sigma(b) < \theta_{\text{elim}}$
and flags a conflict if $\sigma(c) > \theta_{\text{CLS}}$; here cell 2 collapses to a single candidate. If the
resulting state is neither a conflict nor a solution, a branching step then
pins one cell (Algorithm~\ref{alg:step}).}
\label{fig:lattice-representation}
\end{figure}

\paragraph{Recurrent transformer} The model is a recurrent transformer
closely following Sotaku~\cite{LouSotaku2026}: a stack of $4$
attention layers is unrolled for $16$ internal iterations, with the
output of iteration $\ell$ becoming the input to iteration $\ell + 1$.
We re-inject the input lattice encoding at every iteration as a
residual signal. Each iteration emits its own candidate, digit and
conflict logits. At training time every iteration is supervised and contributes
equally to the loss; at inference time we read out only the final
iteration. Position is encoded with a learned
2D positional embedding added to the input; for the larger $30
\times 30$ Maze grids we instead apply 2D
RoPE~\cite{su2024roformer} inside attention.
We find large gains from recurrence (Appendix~\ref{subsec:abl-recurrence}):
with a single iteration instead of $16$, Sudoku-Extreme accuracy drops from
$99.9\%$ to $48\%$, and a single-iteration model still underperforms with
$4\times$ the training compute or the full training split.

\subsection{Solve Loop and Step Operator}
\label{subsec:step-and-solve}

A solve applies the step operator (Algorithm~\ref{alg:step}) repeatedly,
starting from an initial lattice state $x_0$, until a step reaches a conflict
or a solution. We run the same solve procedure at training and inference
time, which ensures that the distribution of lattice states the
model is trained on matches what it encounters at inference. Our setup shares similarities with classical supervised fine-tuning, on-policy
distillation~\cite{agarwal2024onpolicy} and reinforcement
learning~\cite{sutton2018rl}: as in the latter two, the training
distribution is generated by the model's own rollouts and supervision is
state-conditional; as in SFT, the target $\hat{y}$ is computed from data, which we do
in a domain-agnostic manner via the abstraction operator $\alpha$. This
gives strong per-step supervision rather than a sparse trajectory-level
reward or an approximate teacher signal.
The solve loop is a stochastic search that moves down the lattice with
deterministic deduction and random branching. The lattice guarantees termination and monotone
tightening and lets the model abstain by leaving candidates alive; the search
is sound only if every elimination is correct, which depends on the learned
model and which we measure empirically (Section~\ref{sec:eval}). This outer-loop structure contrasts with recurrent
reasoners like TRM~\cite{jolicoeurmartineau2025morerecursivereasoningtiny}
and HRM~\cite{wang2025hierarchicalreasoningmodel}, whose outer loop passes
a latent embedding between steps; ours passes through the lattice itself.
We do not propagate gradients across solve steps;
Appendix~\ref{app:latent-carry} examines carrying continuous state between
them.

A single step is summarized in Algorithm~\ref{alg:step}. We run the recurrent
transformer, which unrolls its $16$ internal iterations and produces per-candidate
confidences at each grid cell and a CLS conflict signal. We then eliminate every candidate
whose confidence falls below a threshold $\theta_{\text{elim}}$. If the
resulting state is neither a complete singleton solution nor flagged
inconsistent, we pick a uniformly random multi-candidate cell (see
Appendix~\ref{app:search-ablations}) and pin it to
a single digit sampled from a softmax of the digit logits $s$ over its
alive candidates at temperature $\tau_{\text{decide}}$ (at inference, the
argmax). The conflict and solved flags always come from the model: its CLS
prediction and the empty-cell test.

\begin{algorithm}[!htbp]
\caption{Step operator $\mathrm{step}(x, Y)$}
\label{alg:step}
\begin{algorithmic}[1]
\Require lattice state $x$;\, ground-truth solutions $Y$ (empty at inference)
\State $\bigl(b^{(1..L)},\, s^{(1..L)},\, c^{(1..L)}\bigr) \gets f_\theta(x)$ \Comment{candidate, digit and CLS logits at every iteration}
\State $b \gets b^{(L)}$;\quad $s \gets s^{(L)}$;\quad $c \gets c^{(L)}$
\State $x' \gets x$;\, for every $(i,j)$ with $\sigma(b_{ij}) < \theta_{\text{elim}}$, set $x'_{ij} \gets 0$ \Comment{threshold elimination}
\State $\text{conflict} \gets [\sigma(c) > \theta_{\text{CLS}}]$ \textbf{or} some cell of $x'$ has zero alive candidates
\State $\text{solved} \gets$ every cell of $x'$ has exactly one alive candidate \textbf{and not} $\text{conflict}$
\If{$Y \neq \emptyset$} \Comment{in training, also use the ground truth}
  \State $\hat{y} \gets x \sqcap \alpha\bigl(\{y \in Y \mid y \text{ consistent with } x\}\bigr)$
  \State $\mathcal{L} \gets \mathrm{compute\_loss}\bigl(\hat{y},\, b^{(1..L)},\, c^{(1..L)}\bigr)$
  \State Update $\theta$ by one optimizer step on $\mathcal{L}$
\EndIf
\If{not $\text{conflict}$ and not $\text{solved}$} \Comment{branch (sample-pin one cell)}
  \State Pick $i^*$ uniformly at random from cells with $\geq 2$ alive candidates
  \State Sample $d^* \sim \mathrm{softmax}(s_{i^*} / \tau_{\text{decide}})$ over alive candidates of $i^*$ \Comment{argmax at inference}
  \State Pin cell $i^*$ to $d^*$ in $x'$ (zero all other candidates of $i^*$)
\EndIf
\State \Return $x',\, \text{conflict},\, \text{solved}$
\end{algorithmic}
\end{algorithm}

\paragraph{Loss design} The candidate-elimination and conflict-detection
heads sit at opposite ends of the soundness/completeness trade-off. Candidate
elimination must be sound (a wrongly eliminated candidate is unrecoverable)
while conflict detection must be complete (a missed conflict can stall search
or lead to returning a wrong answer). We supervise every one of the $16$
internal iterations with three loss terms: $\mathcal{L}_{\text{BCE}}^{(\ell)}$,
an asymmetric BCE between $\sigma(b^{(\ell)})$ and $\hat{y}$ with positive
class weight $w^{+}$ and negative class weight $w^{-} < w^{+}$ to penalize
false eliminations more heavily than false retentions;
$\mathcal{L}_{\text{CLS}}^{(\ell)}$, a symmetric BCE between
$\sigma(c^{(\ell)})$ and $\mathbf{1}[\hat{y} = \bot]$; and
$\mathcal{L}_{\text{CE}}^{(\ell)}$, a per-cell softmax cross-entropy on
digit logits $s^{(\ell)}$ from a separate linear head, at cells where
$\hat{y}$ has a single alive candidate; $s$ also picks the branching digit.
$\mathcal{L}_{\text{CE}}$ substantially improves accuracy and sharpens the
digit choice at branching steps (Appendix~\ref{subsec:abl-ce}). The total loss is
\begin{equation}
\mathcal{L} \;=\; \frac{1}{L} \sum_{\ell=1}^{L} \Big[
  \mathcal{L}_{\text{BCE}}^{(\ell)}
+ \lambda_{\text{cls}} \, \mathcal{L}_{\text{CLS}}^{(\ell)}
+ \lambda_{\text{ce}} \, \mathcal{L}_{\text{CE}}^{(\ell)}
\Big].
\end{equation}
At inference time we eliminate candidates whose probability falls
below a threshold $\theta_{\text{elim}}$ matched to the natural ratio induced
by the asymmetric loss; this outperforms training with a symmetric BCE and
tuning $\theta_{\text{elim}}$ post-hoc. When no $y \in Y$ is still consistent with $x$, the surviving set is empty
and $\hat{y}$ collapses to $\bot$, which would push every candidate sigmoid
toward zero and overload the BCE with conflict-detection pressure. The
lattice contains many representations of an empty solution set; we pick one
that lives further up the lattice by caching the last non-empty
$\hat{y}^{\text{prev}}$ per chain and using $x \sqcap \hat{y}^{\text{prev}}$
as the BCE target once the surviving set becomes empty. This keeps the
conflict localized to the individual cells where $x$ has eliminated every
candidate alive in $\hat{y}^{\text{prev}}$, focuses the BCE on sound
deduction within still-satisfiable cells, and primarily leaves conflict
detection to the CLS head (the candidate head can still flag localized
conflicts when a cell collapses to empty). In our experiments we use $w^{+}/w^{-} = 8$,
$\lambda_{\text{cls}} = 0.1$, $\lambda_{\text{ce}} = 0.2$,
$\theta_{\text{elim}} \approx 0.1$, and decide temperature
$\tau_{\text{decide}} = 1.5$, tuned on Sudoku; we use these settings as the
starting point for the other domains and tune individual settings where
needed (Appendix~\ref{app:hparams}).

\paragraph{Multi-solution supervision via the $\alpha$ operator}
Many problems of interest admit a set of solutions rather than a
unique answer. Standard supervised fine-tuning, written against a single
privileged ground truth, arbitrarily penalizes the model for committing to
any of the equally valid alternatives, echoing the insufficency of using token-level matching (BLEU score) for code generation~\cite{ren2020codebleumethodautomaticevaluation}.
We study this on Maze-Hard, where a maze admits many shortest paths. For
each maze we pre-compute up to $K$ valid solutions $Y = \{y^{(1)}, \ldots,
y^{(K)}\}$ (each a one-hot point on the lattice), sampled uniformly at random
from the all-shortest-paths DAG, and apply the abstraction function
$\alpha$ to the subset of $Y$ still consistent with the current
state, as in Algorithm~\ref{alg:step}. As the state commits against solutions, the
surviving subset shrinks: at branching cells where two or more solutions
still survive, $\alpha$ remains multi-alive, and once the state has
committed to a single solution's path, $\alpha$ collapses to a one-hot.
A candidate is thus a positive target if it appears in at least one
surviving solution. The
special case $K = 1$ recovers standard single-target supervised learning,
which we use for Sudoku and Snowflake Sudoku. Train-time overhead is minimal: the number of
training steps does not depend on $K$, and $K$ adds only minor bookkeeping
per step -- in our Maze experiments, $K = 512$ is roughly $4\%$ slower
per step than $K = 1$.

\paragraph{Parallel solve} A solve follows a single linear chain from the
initial state to a terminal, but in practice we parallelize both training
and inference. For training, we maintain a pool of recent partially-deduced
states, similar to a replay buffer except each instance contributes to a
single gradient step. Each batch samples instances from the pool, runs Step,
removes those the model solves, those it flags as conflicted when they truly
are (checked against $Y$), and those older than $100$ steps, and returns the
rest to the pool. Over training, the pool develops a
natural distribution over lattice depths, giving the model an implicit
curriculum of progressively deeper states. At
inference, we seed the batch with copies of a single puzzle and step each
chain in parallel until one reaches a solution; detected conflicts are
replaced by fresh puzzle instances.

\paragraph{Training and inference details}
The step operator reads the final iteration's logits, with dropout off, both
in training and at inference; only the loss uses every iteration. Averaging
the logits over iterations instead lowers accuracy
(Appendix~\ref{subsec:deduction-readout}).
(i) Each step is wrapped in a random symmetry of the problem
(Appendix~\ref{app:hparams}), which we invert after deduction and
branching. We apply the same per-step augmentation during training
(except for Sudoku, where it did not help).
(ii) At inference we pin the highest-confidence candidate of the chosen
cell instead of sampling it; diversity across chains comes from the
random choice of branching cell, the per-step symmetry and eval-time
dropout (not used for Maze).
(iii) The CLS threshold at inference, $\theta_{\text{CLS}}^{\text{eval}}$,
is set slightly higher than at training (we use $0.6$ for Sudoku and
Snowflake, and $0.53$ for the $30 \times 30$ Maze-Hard runs). The conflict head grows more confident
as the puzzle fills in, so raising the threshold suppresses
false-positive conflicts on early-stage states, reducing the number of
forward passes per solve.
(iv) For Sudoku and Snowflake, a puzzle entering the training pool is the
raw puzzle ($20\%$), has some blanks filled with correct digits ($55\%$), or
has some filled with a share of wrong digits ($25\%$), which gives the
conflict head unsatisfiable states beyond those its own branching reaches
and improves soundness (Appendix~\ref{subsec:abl-inputs}). We skip this for Maze, whose mazes have
multiple solutions.

\section{Evaluation}
\label{sec:eval}

\subsection{Sudoku-Extreme}
\label{subsec:Sudoku-extreme}
We evaluate Lattice Deduction Transformers on Sudoku-Extreme, a benchmark
of hard Sudoku puzzles introduced alongside
HRM~\cite{wang2025hierarchicalreasoningmodel}.
We train an $800$k parameter Lattice Deduction Transformer on
the $1{,}000$ puzzle train set using the procedure described in
Section~\ref{subsec:step-and-solve}. Sudoku-Extreme puzzles have unique
solutions, so each instance is paired with a single ground-truth solution
and the alpha operator reduces to the $K{=}1$ special case. We augment the
training set with the symmetries of Sudoku: any digit permutation and the
dihedral group $D_4$ acting on the grid (rotations and reflections of the
$9 \times 9$ board). We find that tuning the conflict detection threshold
during inference $\theta_{\text{CLS}}^{\text{eval}}$ can improve
solver performance by reducing calls and avoiding timeouts; specifically,
we use $\theta_{\text{CLS}}^{\text{eval}} = 0.6$, tuned on the train set.

We compare to TRM~\cite{jolicoeurmartineau2025morerecursivereasoningtiny},
HRM~\cite{wang2025hierarchicalreasoningmodel} and
Sotaku~v2~\cite{LouSotaku2026v2}, evaluating released checkpoints (for TRM, a
public alphaXiv reproduction), and to Equilibrium Reasoners
(EqR)~\cite{huang2026equilibrium} as reported. Appendix~\ref{app:llm-prompts}
evaluates frontier chain-of-thought LLMs zero-shot on all three benchmarks.

\begin{table}[htbp]
\centering
\footnotesize
\setlength{\tabcolsep}{4pt}
\caption{Results on the $422{,}786$-example Sudoku-Extreme test set.}
\label{tab:Sudoku-extreme}
\begin{tabular}{lrrrrrrrr}
\toprule
& & & & & \textbf{Inference} & \multicolumn{3}{c}{\textbf{Training}} \\
\cmidrule(lr){6-6}\cmidrule(lr){7-9}
\textbf{Model} & \textbf{Params} & \textbf{Train set} & \textbf{Acc.\ \%} & \textbf{Wrong \%} & TFLOPs & PFLOPs & B200 & MFU \\
\midrule
HRM~\cite{wang2025hierarchicalreasoningmodel} & 27M   & 1K$^\ddagger$ & 55.4 & 44.6 & 0.22 & 1{,}466 & 51m$^\dagger$ & 21\% \\
TRM~\cite{jolicoeurmartineau2025morerecursivereasoningtiny} & 5M    & 1K$^\ddagger$ & 79.5 & 20.5 & 0.41 & 2{,}136 & 4.7h$^\dagger$ & 5.6\% \\
\midrule
Sotaku v2~\cite{LouSotaku2026v2} & 800K & 2.7M & 98.96 & 1.04 & 0.14 & 1{,}265 & 4.1h$^\dagger$ & 3.8\% \\
EqR~\cite{huang2026equilibrium} & 5M    & 1K$^\ddagger$ & 99.8 & ---   & 210    & 1{,}640 & 30m$^\dagger$ & 40\% \\
\midrule
LDT, 1K train steps       & 800K  & 1K$^\ddagger$ & 99.36 & 0.06 & 9.6 & 4.7 & 3m & 1.0\% \\
LDT, 2K train steps       & 800K  & 1K$^\ddagger$ & 99.98 & 0.003 & 3.2 & 9.4 & 7m & 1.1\% \\
LDT, 4K train steps       & 800K  & 1K$^\ddagger$ & 99.988 & 0.0003 & 1.5 & 18.7 & 13m & 1.1\% \\
\quad 10{,}000 search rounds & 800K & 1K$^\ddagger$ & \textbf{99.999} & 0.0003 & 1.5 & 18.7 & 13m & 1.1\% \\
\quad portfolio of $5$ models, agree-2 & 4M & 1K$^\ddagger$ & 99.9996 & 0.00008 & 3.1 & 93.5 & 1.1h & 1.1\% \\
\quad portfolio of $15$ models, agree-3 & 12M & 1K$^\ddagger$ & \textbf{100} & 0 & 4.6 & 281 & 3.2h & 1.1\% \\
\quad no training augmentation & 800K & 1K & 96.54 & 3.46 & 0.85 & 18.7 & 11m & 1.3\% \\
\bottomrule
\end{tabular}\\[2pt]
{\scriptsize $^\ddagger$Trained with data augmentation.\quad $^\dagger$Estimated from measured B200 throughput.}

\end{table}

\begin{figure}[htbp]
\centering
\includegraphics[width=\linewidth]{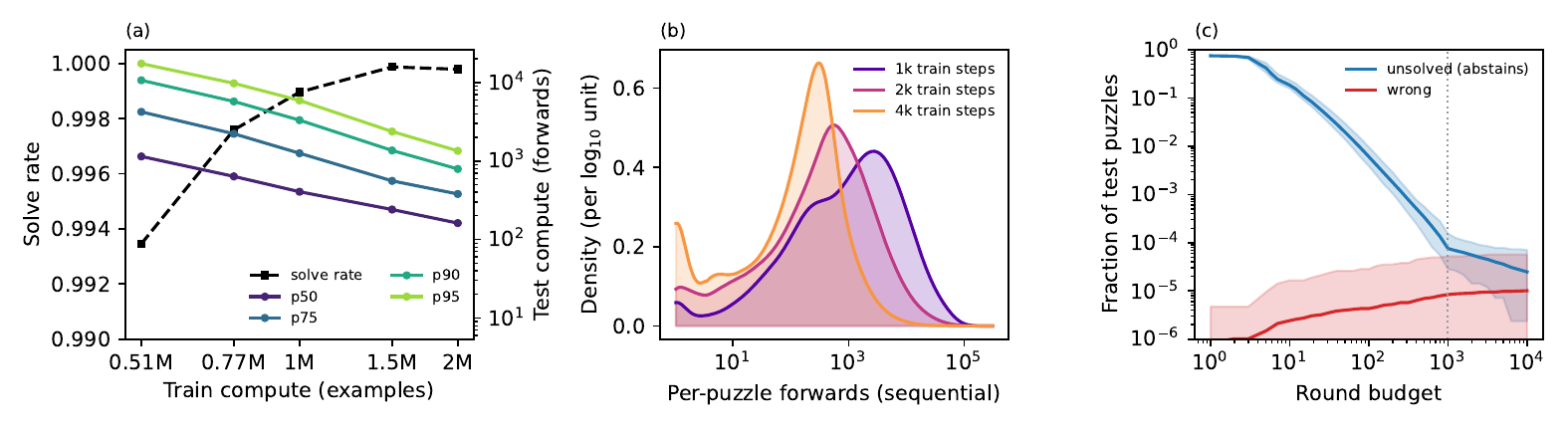}
\caption{Test/train compute trade-off on Sudoku-Extreme. (a) Solve rate and
percentiles of sequential forward passes per solved puzzle against training
examples (training steps $\times$ batch size $512$), mean over $3$ seeds.
(b) Density of per-puzzle sequential forward passes for models trained for
1K, 2K and 4K steps, pooled over seeds. (c) Fraction of the $422{,}786$ test
puzzles left unsolved (the model abstains) and answered wrong within a
round budget, for $15$ models trained for 4K steps on the same $1{,}000$
puzzles (mean, range shaded); the dotted line marks the default budget of
$1{,}000$ rounds.}
\label{fig:sudoku-train-test-compute}
\end{figure}

\paragraph{Results}
Table~\ref{tab:Sudoku-extreme} summarizes the results. LDT rows are means
over $3$ seeds, with at most $1{,}000$ search rounds per puzzle unless noted;
EqR is as reported. To save compute, we evaluate the less accurate models (HRM, TRM,
Sotaku v2, the 1K- and 2K-step rows and the row without training
augmentation) on random subsets of the test set
(Appendix~\ref{app:eval-protocol}). FLOPs and MFU are estimates
(Appendix~\ref{app:compute}). Trained for 4K steps
on the $1{,}000$ training puzzles, LDT solves $99.988\%$ of all $422{,}786$
test puzzles within $1{,}000$ search rounds and $99.999\%$ within
$10{,}000$, with $4$ wrong answers in $1.27$M puzzle evaluations over $3$
seeds; every other unsolved puzzle is a timeout, where the model abstains.
The released HRM and TRM checkpoints solve $55.4\%$ and $79.5\%$, and their
training uses $78$ to $114$ times as many FLOPs
(Figure~\ref{fig:ldt-overview}(b)). Training longer
does not help on the $1{,}000$ training puzzles: at 8K steps accuracy drops slightly and wrong answers rise to
$282$. The row marked ``no training augmentation'' trains without dataset or
per-step augmentation but still applies per-step augmentation at
evaluation. With only $1{,}000$
training puzzles, training augmentation matters mostly for soundness: without
it the model still solves $96.5\%$ of the test set, but $3.5\%$ of its
answers are wrong, against $0.0003\%$ with it (and it solves only $31.6\%$
if evaluation is not augmented either, as it memorizes the training
puzzles). Trained on the full $3.8$M-puzzle training split without
augmentation, the same model solves $99.994\%$. Portfolios of
independently trained models raise accuracy further
(Appendix~\ref{app:portfolio}): on the test set, $5$ models sometimes solve
every puzzle (one of three portfolios), and requiring three of all $15$
models to agree solves every test puzzle (Table~\ref{tab:Sudoku-extreme}). Search remains
stochastic, so a small chance of error always remains.

Figure~\ref{fig:sudoku-train-test-compute}(a) shows the underlying
train/test compute trade-off: more training shortens search, with the
median and tail percentiles (p50, p75, p90, p95) of per-puzzle forward
passes all falling, the median by a factor of $7$ between 1K and 4K
training steps (from $1{,}142$ to $162$, mean over
seeds). Figure~\ref{fig:sudoku-train-test-compute}(b)
shows the distribution: as training increases, the fraction of puzzles
solved by the first deduction step grows from $2.5\%$ to $10.9\%$, and the
peak of the search mode moves to fewer forward passes; better deductions
translate to less guessing. Figure~\ref{fig:sudoku-train-test-compute}(c)
shows the trade-off between completeness and soundness as the search budget
grows: from $1$ to $10{,}000$ rounds, the fraction of unsolved puzzles falls
from $75\%$ to $0.0025\%$, while wrong answers rise from about $1$ to $10$
per million puzzles.

\subsection{Snowflake Sudoku}
\label{subsec:snowflake}

\begin{figure}[htbp]
\centering
\footnotesize
\begin{minipage}[b]{0.2\linewidth}
\centering
\begin{tikzpicture}[x=0.69cm, y=0.69cm, font=\footnotesize]
\fill[blue!25, fill opacity=0.6] (1.500,-0.866)--(1.000,0.000)--(0.000,0.000)--(-0.500,-0.866)--(-0.000,-1.732)--(1.000,-1.732)--cycle;
\draw[blue!60, dashed] (1.500,-0.866)--(1.000,0.000)--(0.000,0.000)--(-0.500,-0.866)--(-0.000,-1.732)--(1.000,-1.732)--cycle;
\draw[gray!60] (0.000,-0.000)--(1.000,0.000);
\draw[gray!60] (0.000,-0.000)--(0.500,0.866);
\draw[gray!60] (0.000,-0.000)--(-0.500,0.866);
\draw[gray!60] (0.000,-0.000)--(-1.000,0.000);
\draw[gray!60] (0.000,-0.000)--(-0.500,-0.866);
\draw[gray!60] (0.000,-0.000)--(0.500,-0.866);
\draw[thick] (1.000,0.000)--(0.500,0.866)--(-0.500,0.866)--(-1.000,0.000)--(-0.500,-0.866)--(0.500,-0.866)--cycle;
\node at (0.000,-0.577) {2};
\node at (0.500,0.289) {3};
\node at (-0.500,0.289) {1};
\node at (-0.500,-0.289) {5};
\draw[gray!60] (1.500,-0.866)--(2.500,-0.866);
\draw[gray!60] (1.500,-0.866)--(2.000,0.000);
\draw[gray!60] (1.500,-0.866)--(1.000,0.000);
\draw[gray!60] (1.500,-0.866)--(0.500,-0.866);
\draw[gray!60] (1.500,-0.866)--(1.000,-1.732);
\draw[gray!60] (1.500,-0.866)--(2.000,-1.732);
\draw[thick] (2.500,-0.866)--(2.000,0.000)--(1.000,0.000)--(0.500,-0.866)--(1.000,-1.732)--(2.000,-1.732)--cycle;
\node at (1.500,-1.443) {6};
\node at (2.000,-1.155) {3};
\node at (1.500,-0.289) {2};
\node at (1.000,-1.155) {5};
\draw[gray!60] (0.000,-1.732)--(1.000,-1.732);
\draw[gray!60] (0.000,-1.732)--(0.500,-0.866);
\draw[gray!60] (0.000,-1.732)--(-0.500,-0.866);
\draw[gray!60] (0.000,-1.732)--(-1.000,-1.732);
\draw[gray!60] (0.000,-1.732)--(-0.500,-2.598);
\draw[gray!60] (0.000,-1.732)--(0.500,-2.598);
\draw[thick] (1.000,-1.732)--(0.500,-0.866)--(-0.500,-0.866)--(-1.000,-1.732)--(-0.500,-2.598)--(0.500,-2.598)--cycle;
\node at (0.000,-2.309) {1};
\node at (0.500,-2.021) {5};
\node at (0.500,-1.443) {3};
\node at (-0.500,-1.443) {6};
\end{tikzpicture}\\[2pt]
Order $3$
\end{minipage}\hfill
\begin{minipage}[b]{0.2\linewidth}
\centering
\begin{tikzpicture}[x=0.69cm, y=0.69cm, font=\footnotesize]
\fill[blue!25, fill opacity=0.6] (0.500,0.866)--(0.000,1.732)--(-1.000,1.732)--(-1.500,0.866)--(-1.000,-0.000)--(0.000,-0.000)--cycle;
\draw[blue!60, dashed] (0.500,0.866)--(0.000,1.732)--(-1.000,1.732)--(-1.500,0.866)--(-1.000,-0.000)--(0.000,-0.000)--cycle;
\fill[blue!25, fill opacity=0.6] (0.000,0.000)--(-0.500,0.866)--(-1.500,0.866)--(-2.000,0.000)--(-1.500,-0.866)--(-0.500,-0.866)--cycle;
\draw[blue!60, dashed] (0.000,0.000)--(-0.500,0.866)--(-1.500,0.866)--(-2.000,0.000)--(-1.500,-0.866)--(-0.500,-0.866)--cycle;
\draw[gray!60] (0.000,-0.000)--(1.000,0.000);
\draw[gray!60] (0.000,-0.000)--(0.500,0.866);
\draw[gray!60] (0.000,-0.000)--(-0.500,0.866);
\draw[gray!60] (0.000,-0.000)--(-1.000,0.000);
\draw[gray!60] (0.000,-0.000)--(-0.500,-0.866);
\draw[gray!60] (0.000,-0.000)--(0.500,-0.866);
\draw[thick] (1.000,0.000)--(0.500,0.866)--(-0.500,0.866)--(-1.000,0.000)--(-0.500,-0.866)--(0.500,-0.866)--cycle;
\node at (0.500,-0.289) {3};
\node at (0.000,-0.577) {4};
\draw[gray!60] (-1.500,0.866)--(-0.500,0.866);
\draw[gray!60] (-1.500,0.866)--(-1.000,1.732);
\draw[gray!60] (-1.500,0.866)--(-2.000,1.732);
\draw[gray!60] (-1.500,0.866)--(-2.500,0.866);
\draw[gray!60] (-1.500,0.866)--(-2.000,0.000);
\draw[gray!60] (-1.500,0.866)--(-1.000,0.000);
\draw[thick] (-0.500,0.866)--(-1.000,1.732)--(-2.000,1.732)--(-2.500,0.866)--(-2.000,0.000)--(-1.000,0.000)--cycle;
\node at (-2.000,0.577) {1};
\node at (-2.000,1.155) {2};
\draw[gray!60] (-1.500,-0.866)--(-0.500,-0.866);
\draw[gray!60] (-1.500,-0.866)--(-1.000,0.000);
\draw[gray!60] (-1.500,-0.866)--(-2.000,0.000);
\draw[gray!60] (-1.500,-0.866)--(-2.500,-0.866);
\draw[gray!60] (-1.500,-0.866)--(-2.000,-1.732);
\draw[gray!60] (-1.500,-0.866)--(-1.000,-1.732);
\draw[thick] (-0.500,-0.866)--(-1.000,0.000)--(-2.000,0.000)--(-2.500,-0.866)--(-2.000,-1.732)--(-1.000,-1.732)--cycle;
\node at (-1.500,-1.443) {3};
\node at (-2.000,-1.155) {4};
\node at (-2.000,-0.577) {6};
\node at (-1.500,-0.289) {1};
\draw[gray!60] (0.000,1.732)--(1.000,1.732);
\draw[gray!60] (0.000,1.732)--(0.500,2.598);
\draw[gray!60] (0.000,1.732)--(-0.500,2.598);
\draw[gray!60] (0.000,1.732)--(-1.000,1.732);
\draw[gray!60] (0.000,1.732)--(-0.500,0.866);
\draw[gray!60] (0.000,1.732)--(0.500,0.866);
\draw[thick] (1.000,1.732)--(0.500,2.598)--(-0.500,2.598)--(-1.000,1.732)--(-0.500,0.866)--(0.500,0.866)--cycle;
\node at (0.500,1.443) {6};
\node at (-0.500,1.443) {4};
\node at (-0.500,2.021) {1};
\node at (0.000,2.309) {2};
\end{tikzpicture}\\[2pt]
Order $4$
\end{minipage}\hfill
\begin{minipage}[b]{0.24\linewidth}
\centering
\begin{tikzpicture}[x=0.69cm, y=0.69cm, font=\footnotesize]
\fill[blue!25, fill opacity=0.6] (0.500,-0.866)--(0.000,0.000)--(-1.000,0.000)--(-1.500,-0.866)--(-1.000,-1.732)--(0.000,-1.732)--cycle;
\draw[blue!60, dashed] (0.500,-0.866)--(0.000,0.000)--(-1.000,0.000)--(-1.500,-0.866)--(-1.000,-1.732)--(0.000,-1.732)--cycle;
\fill[blue!25, fill opacity=0.6] (0.000,0.000)--(-0.500,0.866)--(-1.500,0.866)--(-2.000,0.000)--(-1.500,-0.866)--(-0.500,-0.866)--cycle;
\draw[blue!60, dashed] (0.000,0.000)--(-0.500,0.866)--(-1.500,0.866)--(-2.000,0.000)--(-1.500,-0.866)--(-0.500,-0.866)--cycle;
\fill[blue!25, fill opacity=0.6] (0.500,0.866)--(0.000,1.732)--(-1.000,1.732)--(-1.500,0.866)--(-1.000,-0.000)--(0.000,-0.000)--cycle;
\draw[blue!60, dashed] (0.500,0.866)--(0.000,1.732)--(-1.000,1.732)--(-1.500,0.866)--(-1.000,-0.000)--(0.000,-0.000)--cycle;
\draw[gray!60] (-1.500,-0.866)--(-0.500,-0.866);
\draw[gray!60] (-1.500,-0.866)--(-1.000,0.000);
\draw[gray!60] (-1.500,-0.866)--(-2.000,0.000);
\draw[gray!60] (-1.500,-0.866)--(-2.500,-0.866);
\draw[gray!60] (-1.500,-0.866)--(-2.000,-1.732);
\draw[gray!60] (-1.500,-0.866)--(-1.000,-1.732);
\draw[thick] (-0.500,-0.866)--(-1.000,0.000)--(-2.000,0.000)--(-2.500,-0.866)--(-2.000,-1.732)--(-1.000,-1.732)--cycle;
\node at (-2.000,-1.155) {2};
\node at (-2.000,-0.577) {3};
\node at (-1.500,-0.289) {6};
\draw[gray!60] (0.000,-0.000)--(1.000,0.000);
\draw[gray!60] (0.000,-0.000)--(0.500,0.866);
\draw[gray!60] (0.000,-0.000)--(-0.500,0.866);
\draw[gray!60] (0.000,-0.000)--(-1.000,0.000);
\draw[gray!60] (0.000,-0.000)--(-0.500,-0.866);
\draw[gray!60] (0.000,-0.000)--(0.500,-0.866);
\draw[thick] (1.000,0.000)--(0.500,0.866)--(-0.500,0.866)--(-1.000,0.000)--(-0.500,-0.866)--(0.500,-0.866)--cycle;
\node at (0.500,-0.289) {4};
\draw[gray!60] (-1.500,0.866)--(-0.500,0.866);
\draw[gray!60] (-1.500,0.866)--(-1.000,1.732);
\draw[gray!60] (-1.500,0.866)--(-2.000,1.732);
\draw[gray!60] (-1.500,0.866)--(-2.500,0.866);
\draw[gray!60] (-1.500,0.866)--(-2.000,0.000);
\draw[gray!60] (-1.500,0.866)--(-1.000,0.000);
\draw[thick] (-0.500,0.866)--(-1.000,1.732)--(-2.000,1.732)--(-2.500,0.866)--(-2.000,0.000)--(-1.000,0.000)--cycle;
\node at (-2.000,0.577) {1};
\node at (-2.000,1.155) {2};
\draw[gray!60] (0.000,-1.732)--(1.000,-1.732);
\draw[gray!60] (0.000,-1.732)--(0.500,-0.866);
\draw[gray!60] (0.000,-1.732)--(-0.500,-0.866);
\draw[gray!60] (0.000,-1.732)--(-1.000,-1.732);
\draw[gray!60] (0.000,-1.732)--(-0.500,-2.598);
\draw[gray!60] (0.000,-1.732)--(0.500,-2.598);
\draw[thick] (1.000,-1.732)--(0.500,-0.866)--(-0.500,-0.866)--(-1.000,-1.732)--(-0.500,-2.598)--(0.500,-2.598)--cycle;
\node at (0.000,-2.309) {6};
\node at (-0.500,-2.021) {1};
\node at (-0.500,-1.443) {5};
\node at (0.500,-1.443) {3};
\draw[gray!60] (0.000,1.732)--(1.000,1.732);
\draw[gray!60] (0.000,1.732)--(0.500,2.598);
\draw[gray!60] (0.000,1.732)--(-0.500,2.598);
\draw[gray!60] (0.000,1.732)--(-1.000,1.732);
\draw[gray!60] (0.000,1.732)--(-0.500,0.866);
\draw[gray!60] (0.000,1.732)--(0.500,0.866);
\draw[thick] (1.000,1.732)--(0.500,2.598)--(-0.500,2.598)--(-1.000,1.732)--(-0.500,0.866)--(0.500,0.866)--cycle;
\node at (0.500,1.443) {1};
\node at (0.000,1.155) {6};
\node at (0.000,2.309) {3};
\node at (0.500,2.021) {4};
\end{tikzpicture}\\[2pt]
Order $5$
\end{minipage}\hfill
\begin{minipage}[b]{0.28\linewidth}
\centering
\resizebox{!}{3cm}{\begin{tikzpicture}
\path[draw=gray!70] (-2.000,5.196)--(-2.500,6.062)--(-3.500,6.062)--(-4.000,5.196)--(-3.500,4.330)--(-2.500,4.330)--cycle;
\path[draw=gray!70] (-2.000,3.464)--(-2.500,4.330)--(-3.500,4.330)--(-4.000,3.464)--(-3.500,2.598)--(-2.500,2.598)--cycle;
\path[draw=gray!70] (-2.000,1.732)--(-2.500,2.598)--(-3.500,2.598)--(-4.000,1.732)--(-3.500,0.866)--(-2.500,0.866)--cycle;
\path[draw=gray!70] (-2.000,0.000)--(-2.500,0.866)--(-3.500,0.866)--(-4.000,0.000)--(-3.500,-0.866)--(-2.500,-0.866)--cycle;
\path[draw=gray!70] (-2.000,-1.732)--(-2.500,-0.866)--(-3.500,-0.866)--(-4.000,-1.732)--(-3.500,-2.598)--(-2.500,-2.598)--cycle;
\path[draw=gray!70] (-0.500,4.330)--(-1.000,5.196)--(-2.000,5.196)--(-2.500,4.330)--(-2.000,3.464)--(-1.000,3.464)--cycle;
\path[draw=gray!70] (-0.500,2.598)--(-1.000,3.464)--(-2.000,3.464)--(-2.500,2.598)--(-2.000,1.732)--(-1.000,1.732)--cycle;
\path[fill=gray!30, draw=black, thick] (-0.500,0.866)--(-1.000,1.732)--(-2.000,1.732)--(-2.500,0.866)--(-2.000,0.000)--(-1.000,0.000)--cycle;
\path[fill=gray!30, draw=black, thick] (-0.500,-0.866)--(-1.000,0.000)--(-2.000,0.000)--(-2.500,-0.866)--(-2.000,-1.732)--(-1.000,-1.732)--cycle;
\path[draw=gray!70] (-0.500,-2.598)--(-1.000,-1.732)--(-2.000,-1.732)--(-2.500,-2.598)--(-2.000,-3.464)--(-1.000,-3.464)--cycle;
\path[draw=gray!70] (1.000,3.464)--(0.500,4.330)--(-0.500,4.330)--(-1.000,3.464)--(-0.500,2.598)--(0.500,2.598)--cycle;
\path[fill=gray!30, draw=black, thick] (1.000,1.732)--(0.500,2.598)--(-0.500,2.598)--(-1.000,1.732)--(-0.500,0.866)--(0.500,0.866)--cycle;
\path[fill=gray!30, draw=black, thick] (1.000,0.000)--(0.500,0.866)--(-0.500,0.866)--(-1.000,0.000)--(-0.500,-0.866)--(0.500,-0.866)--cycle;
\path[fill=gray!30, draw=black, thick] (1.000,-1.732)--(0.500,-0.866)--(-0.500,-0.866)--(-1.000,-1.732)--(-0.500,-2.598)--(0.500,-2.598)--cycle;
\path[draw=gray!70] (1.000,-3.464)--(0.500,-2.598)--(-0.500,-2.598)--(-1.000,-3.464)--(-0.500,-4.330)--(0.500,-4.330)--cycle;
\path[draw=gray!70] (2.500,2.598)--(2.000,3.464)--(1.000,3.464)--(0.500,2.598)--(1.000,1.732)--(2.000,1.732)--cycle;
\path[draw=gray!70] (2.500,0.866)--(2.000,1.732)--(1.000,1.732)--(0.500,0.866)--(1.000,0.000)--(2.000,0.000)--cycle;
\path[draw=gray!70] (2.500,-0.866)--(2.000,0.000)--(1.000,0.000)--(0.500,-0.866)--(1.000,-1.732)--(2.000,-1.732)--cycle;
\path[draw=gray!70] (2.500,-2.598)--(2.000,-1.732)--(1.000,-1.732)--(0.500,-2.598)--(1.000,-3.464)--(2.000,-3.464)--cycle;
\path[draw=gray!70] (2.500,-4.330)--(2.000,-3.464)--(1.000,-3.464)--(0.500,-4.330)--(1.000,-5.196)--(2.000,-5.196)--cycle;
\path[draw=gray!70] (4.000,1.732)--(3.500,2.598)--(2.500,2.598)--(2.000,1.732)--(2.500,0.866)--(3.500,0.866)--cycle;
\path[draw=gray!70] (4.000,0.000)--(3.500,0.866)--(2.500,0.866)--(2.000,0.000)--(2.500,-0.866)--(3.500,-0.866)--cycle;
\path[draw=gray!70] (4.000,-1.732)--(3.500,-0.866)--(2.500,-0.866)--(2.000,-1.732)--(2.500,-2.598)--(3.500,-2.598)--cycle;
\path[draw=gray!70] (4.000,-3.464)--(3.500,-2.598)--(2.500,-2.598)--(2.000,-3.464)--(2.500,-4.330)--(3.500,-4.330)--cycle;
\path[draw=gray!70] (4.000,-5.196)--(3.500,-4.330)--(2.500,-4.330)--(2.000,-5.196)--(2.500,-6.062)--(3.500,-6.062)--cycle;
\end{tikzpicture}}\\[2pt]
Order $5$ in the covering grid
\end{minipage}
\caption{Snowflake Sudoku puzzles of orders $3$, $4$ and $5$, each with a
unique solution. The six cells of each hexagon must hold distinct values, as
must the six shaded cells around each point where three hexagons meet. Every
puzzle is placed in a fixed grid of $25$ hexagon positions.}
\label{fig:snowflake}
\end{figure}

We use snowflake Sudoku to study whether LDT carries over to puzzles of
variable size, and how it generalizes to sizes not seen in training. A
snowflake Sudoku of order $n$ consists of $n$ hexagons joined edge to edge,
each divided into six triangular cells, so it has $6n$ cells
(Figure~\ref{fig:snowflake}). Each cell takes a value in $\{1, \dots, 6\}$.
The six cells of each hexagon must be distinct, and so must the six cells
around every point where three hexagons meet (two from each hexagon). Every
puzzle has its own randomly grown layout, so the number and position of these
cross-hexagon groups vary between puzzles. We place puzzles of every order in
a fixed covering grid, with a read-only mask per cell marking which positions
belong to the puzzle; Appendix~\ref{app:snowflake-data} describes the data and
training setup.

We train and test on minimal puzzles (no given can be removed without losing
a unique solution) that naked and hidden singles alone cannot solve. These
are still much easier than Sudoku-Extreme: current frontier LLMs solve
$96$--$100\%$ of them zero-shot, against $44$--$68\%$ of Sudoku-Extreme
puzzles (Appendix~\ref{app:llm-prompts}). We compare LDT, the same LDT model in one shot (one model call, the
highest-scoring digit in every cell), and direct prediction (the same model
trained only with cross-entropy against
the solution and evaluated in one shot). Table~\ref{tab:snowflake}
summarizes the results. Trained on $1{,}000$ puzzles of orders $4$ to $8$
for under an hour on a single B200, LDT solves every test puzzle in every
seed, against $95.5\%$ in one shot and $88.9\%$ for direct prediction; with
$2{,}000$ training puzzles the gap to direct prediction narrows to $2.4$
points. On puzzle sizes not seen in training the gap is larger
(Appendix~\ref{app:ood-snowflake}). Trained on orders $6$ to $8$ and tested
on orders $9$ and $10$, LDT solves $99.4\%$ with $0.2\%$ wrong answers,
while one shot ($78.9\%$) and direct prediction ($78.4\%$) are tied, so the
gain comes from search. Trained only on orders $4$ to $6$, every method
degrades: LDT solves $77.2\%$ of orders $7$ to $10$ but gives $15.8\%$ wrong
answers, against $67.4\%$ for direct prediction. When training data is
scarce or the test puzzles are far from the training sizes, LDT training
appears to help twice: the LDT model's one-shot predictions are better than
direct prediction's ($95.5\%$ against $88.9\%$ with $1{,}000$ puzzles,
$54.0\%$ against $32.6\%$ trained on orders $4$ to $6$), and search improves
them further. Training on states from the model's own solving may thus also help
prediction without search.

\begin{table}[htbp]
\centering
\footnotesize
\setlength{\tabcolsep}{4pt}
\caption{Snowflake Sudoku, minimal puzzles that singles cannot solve: accuracy / wrong answers (\%), mean over $3$
seeds. In distribution: orders $4$ to $8$, $1$K or $2$K training puzzles.
Order transfer: training orders $\to$ test orders.}
\label{tab:snowflake}
\begin{tabular}{lcccc}
\toprule
& \multicolumn{2}{c}{\textbf{In distribution}} & \multicolumn{2}{c}{\textbf{Order transfer}} \\
\cmidrule(lr){2-3} \cmidrule(lr){4-5}
& \textbf{1K} & \textbf{2K} & \textbf{6--8 $\to$ 9--10} & \textbf{4--6 $\to$ 7--10} \\
\midrule
LDT & \textbf{100.00 / 0.00} & \textbf{100.00 / 0.00} & \textbf{99.40 / 0.22} & \textbf{77.21 / 15.80} \\
LDT, one shot & 95.52 / 4.48 & 96.77 / 3.23 & 78.88 / 21.12 & 53.99 / 46.01 \\
Direct prediction & 88.87 / 11.13 & 97.63 / 2.37 & 78.43 / 21.57 & 32.60 / 67.40 \\
\bottomrule
\end{tabular}
\end{table}

\subsection{Maze-Hard}
\label{subsec:maze}

\begin{figure}[htbp]
\centering
\begin{subfigure}[b]{0.48\linewidth}
\centering
\raisebox{0.6cm}{\includegraphics[height=3.8cm]{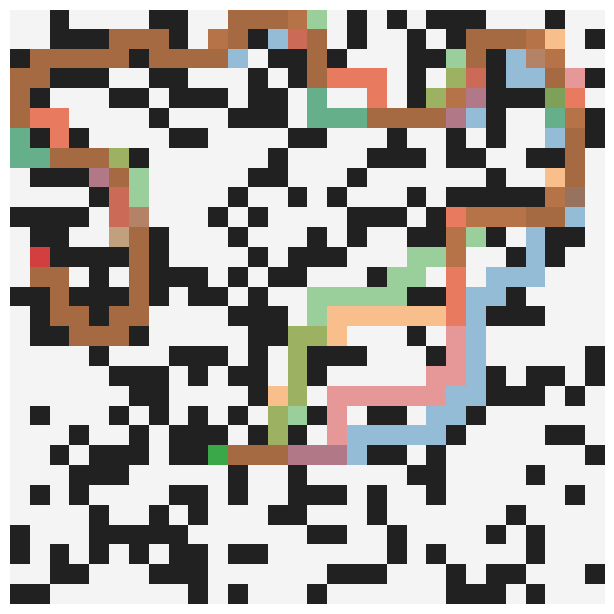}}
\caption{Maze-Hard example ($30 \times 30$).}
\label{fig:maze-example}
\end{subfigure}\hfill
\begin{subfigure}[b]{0.48\linewidth}
\centering
\includegraphics[height=4.6cm]{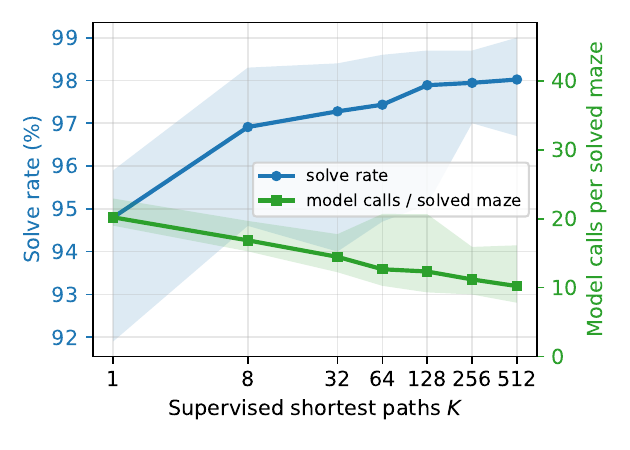}
\caption{Solve rate and model calls vs.\ $K$.}
\label{fig:maze-k-sweep-loss}
\end{subfigure}
\caption{Maze-Hard task and multi-solution supervision via $\alpha$.
(a) one $30 \times 30$ Maze-Hard instance with start, goal, and
several of the many shortest paths of equal length. (b) solve rate
and model calls per solved maze as a function of the sample budget $K$:
mean over $9$ seeds (seed range shaded) on $1{,}000$ test mazes of size $15 \times 15$,
trained for $4{,}000$ steps.}
\label{fig:maze-k-sweep}
\end{figure}

We evaluate Lattice Deduction Transformers on Maze-Hard, a benchmark
introduced alongside HRM~\cite{wang2025hierarchicalreasoningmodel}. The puzzles are
$30 \times 30$ mazes whose shortest path between start and goal cells
has length at least $110$. Unlike Sudoku, a single instance can admit
millions of distinct shortest paths (Figure~\ref{fig:maze-example}), so the
alpha operator can run at $K > 1$ by sampling uniformly from the
all-shortest-paths DAG, which is computationally inexpensive. Each cell
takes one of two values (on or off the path), so eliminating one value
selects the other; here, the lattice's role is to let the model leave a cell
undecided.

We follow the same experimental setup as Sudoku-Extreme, with
adjustments for the larger grid: embedding dimension $d = 192$, 2D
RoPE in place of the learned positional embedding, $20{,}000$ training
steps at batch size $192$ and a conflict threshold of $0.53$ at
evaluation (Appendix~\ref{app:maze-hparams}). We report two settings, both trained on the $1{,}000$-puzzle
Maze-Hard split (matching the TRM setup). The first uses $K = 1$,
pairing each instance with a single ground-truth solution; the
second uses $K = 512$, sampling $512$ shortest paths per instance
from the all-shortest-paths DAG.

\begin{table}[htbp]
\centering
\footnotesize
\setlength{\tabcolsep}{4pt}
\caption{Maze-Hard test results. A path is correct if it is \emph{any} shortest
path from start to goal. Every row reports the best checkpoint over
training, selected on the test set. LDT:
mean over $3$ seeds. Training covers each method up to its reported checkpoint;
FLOPs are estimates.}
\label{tab:maze-hard}
\begin{tabular}{lrrrrrrr}
\toprule
& & & & & \multicolumn{3}{c}{\textbf{Training}} \\
\cmidrule(lr){6-8}
\textbf{Model} & \textbf{Params} & \textbf{Train set} & \textbf{Acc.\ \%} & \textbf{Wrong \%} & PFLOPs & B200 & MFU \\
\midrule
HRM~\cite{wang2025hierarchicalreasoningmodel} & 27M  & 1K$^\ddagger$ & 84.6 & 15.4 & 46{,}944 & 20.7h$^\dagger$ & 28\% \\
TRM~\cite{jolicoeurmartineau2025morerecursivereasoningtiny} & 7M   & 1K & 91.9 & 8.1 & 5{,}971 & 7.9h$^\dagger$ & 9.3\% \\
\midrule
LDT, $K=1$       & 1.8M & 1K$^\ddagger$ & 99.0 & 0.9 & 1{,}024 & 6.9h & 1.8\% \\
LDT, $K=512$     & 1.8M & 1K$^\ddagger$ & \textbf{99.8} & 0.2 & 1{,}303 & 9.1h & 1.8\% \\
\bottomrule
\end{tabular}\\[2pt]
{\scriptsize $^\ddagger$Trained with data augmentation.\quad $^\dagger$Estimated from measured B200 throughput.}

\end{table}

\begin{figure}[htbp]
\centering
\begin{subfigure}[b]{0.57\linewidth}
\centering
\includegraphics[width=\linewidth]{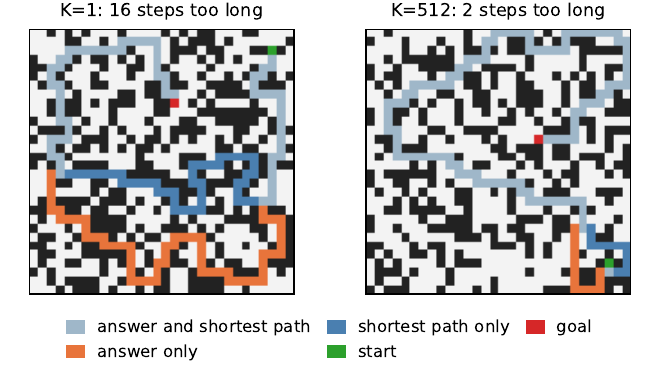}
\caption{One wrong answer at each $K$.}
\label{fig:maze-wrong-examples}
\end{subfigure}\hfill
\begin{subfigure}[b]{0.41\linewidth}
\centering
\includegraphics[width=\linewidth]{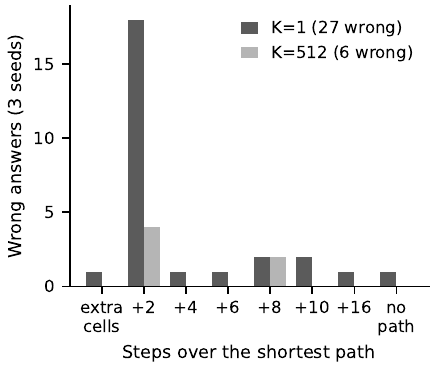}
\caption{Excess length of all wrong answers.}
\label{fig:maze-wrong-excess}
\end{subfigure}
\caption{LDT wrong answers on Maze-Hard (Table~\ref{tab:maze-hard} checkpoints).
(a) each answer against the shortest path that shares the most cells with it.
(b) wrong answers over $3$ seeds by the number of steps their path is longer
than the shortest path; ``extra cells'' marks a shortest path with additional
cells off the path, ``no path'' an answer that does not reach the goal.}
\label{fig:maze-wrong}
\end{figure}

Training is unstable for every method, so Table~\ref{tab:maze-hard} reports
the best checkpoint over training for each, selected on the test set
(Appendix~\ref{app:eval-protocol}); FLOPs are estimated as in
Appendix~\ref{app:compute}. HRM and TRM count a path as correct only if it
matches the dataset's reference path; since every shortest path solves the
maze, we accept any of them for all methods. We train HRM and TRM with and
without $8\times$ dihedral augmentation and report each one's better run:
augmentation lowers TRM's best accuracy from $91.9\%$ to $21.3\%$ and raises
HRM's from $82.2\%$ to $84.6\%$, reached only after about $122$K steps
instead of $39$K. Under exact match, the reported HRM and TRM checkpoints
score $16.7\%$ and $84.5\%$. Table~\ref{tab:maze-hard} reports two LDT settings, both trained on the
$1{,}000$-puzzle Maze-Hard split. With $K = 1$, LDT solves $99.0\%$ of the
test mazes; with $K = 512$, $99.8\%$, against $91.9\%$ for the best TRM
checkpoint. Neither setting is empirically sound: almost all unsolved mazes
are wrong answers rather than abstentions ($27$ wrong answers and $3$
timeouts over $3$ seeds at $K = 1$, $6$ wrong answers at $K = 512$). Nearly
all wrong answers are paths from start to goal that are slightly longer than
the shortest path, most often by $2$ steps (Figure~\ref{fig:maze-wrong};
Appendix~\ref{app:maze-failures}).

\paragraph{Multi-solution supervision via $\alpha$}
A typical $30 \times 30$ maze admits millions of distinct shortest
paths, so $K = 512$ samples only a small fraction of the valid
solution set. The solve loop's
per-step target is $\hat{y} = x \sqcap \alpha(\{y \in Y \mid y
\text{ consistent with } x\})$, so as $K$ grows we feed the model an
increasingly informative intersection of valid solutions and the
supervision signal sharpens without changing the architecture or
loss. Figure~\ref{fig:maze-k-sweep-loss} plots solve rate and
model calls per solved maze as a function of $K$; larger $K$
yields a strictly tighter target, a higher solve rate, and shorter
searches at inference. The gain is concentrated at small $K$ and
saturates quickly: most of the benefit is recovered with a handful
of sampled solutions, and further increases yield diminishing
returns. Tasks with different solution-set geometry may show
different curves.

\section{Conclusion}
\label{sec:conclusion}

We introduced the Lattice Deduction Transformer (LDT), an architecture
that iteratively refines a solution through
forward passes capturing sound step-by-step deduction, rather than
opaque latent updates. The model is trained via the
solve procedure on a batch of partially-deduced lattice states that
evolves under its own forward pass and supervised against the alpha
operator. At inference, a parallel solve runs many chains per puzzle through the same
deduction operator and exploits its soundness to either return a correct
answer or abstain. Together, this allows an
$800{,}000$-parameter model to exceed prior small recurrent reasoners on hard reasoning benchmarks while rarely returning a
wrong answer. While we instantiate LDT on combinatorial
puzzles, the architecture is agnostic to the underlying lattice. In principle, 
any abstract domain admitting a deduction operator could be substituted.

\paragraph{Limitations} LDT is well-suited to tasks with clear
logical structure, where a fixed set of rules induces a sound
deduction operator that the model learns to apply. Benchmarks like
ARC-AGI~\cite{chollet2019measure, chollet2026arcagi2newchallengefrontier, foundation2026arcagi3newchallengefrontier}
test programming-by-example, which demands something stronger: each task carries its own rules, which
the model must \emph{infer} from a handful of demonstrations before
it can deduce anything about the test input. A naive port of LDT
plateaus at roughly $36\%$ pass rate, with no gains from test-time
search. This is because the conflict head becomes unreliable, so the search cannot
tell good branches from bad. Closing this gap will likely require additional machinery from
TRM~\cite{jolicoeurmartineau2025morerecursivereasoningtiny}, in
particular deep supervision that pushes a useful gradient through
every intermediate lattice state and stronger recursive structure
across deduction steps in place of simple iteration. More speculatively, this setting may require a different abstract
domain entirely: one over the \emph{programs} that produce solutions,
rather than over the solution states themselves.

\bibliographystyle{unsrt}
\bibliography{bibli}

@article{davis1960dp,
  author  = {Davis, Martin and Putnam, Hilary},
  title   = {A Computing Procedure for Quantification Theory},
  journal = {Journal of the ACM},
  volume  = {7},
  number  = {3},
  pages   = {201--215},
  year    = {1960},
  doi     = {10.1145/321033.321034}
}

@article{davis1962dpll,
  author  = {Davis, Martin and Logemann, George and Loveland, Donald},
  title   = {A Machine Program for Theorem-Proving},
  journal = {Communications of the ACM},
  volume  = {5},
  number  = {7},
  pages   = {394--397},
  year    = {1962},
  doi     = {10.1145/368273.368557}
}

@incollection{marquessilva2009cdcl,
  author    = {Jo{\~{a}}o P. Marques-Silva and In{\^{e}}s Lynce and
               Sharad Malik},
  title     = {Conflict-Driven Clause Learning {SAT} Solvers},
  booktitle = {Handbook of Satisfiability},
  editor    = {Armin Biere and Marijn Heule and Hans van Maaren and
               Toby Walsh},
  series    = {Frontiers in Artificial Intelligence and Applications},
  volume    = {185},
  pages     = {131--153},
  publisher = {IOS Press},
  year      = {2009}
}

@inproceedings{dsilva2013acdcl,
  author    = {Vijay D'Silva and Leopold Haller and Daniel Kroening},
  title     = {Abstract Conflict Driven Learning},
  booktitle = {Proceedings of the 40th Annual ACM SIGPLAN-SIGACT Symposium
               on Principles of Programming Languages (POPL)},
  pages     = {143--154},
  publisher = {ACM},
  year      = {2013},
  doi       = {10.1145/2429069.2429087}
}

@inproceedings{thakurReps2012stalmarck,
  author    = {Aditya V. Thakur and Thomas W. Reps},
  title     = {A Generalization of St{\aa}lmarck's Method},
  booktitle = {Static Analysis -- 19th International Symposium (SAS)},
  series    = {Lecture Notes in Computer Science},
  volume    = {7460},
  pages     = {334--351},
  publisher = {Springer},
  year      = {2012}
}

@inproceedings{CousotCousot77-1,
   author =    {Cousot, P{.} and Cousot, R{.}},
   title =     {Abstract interpretation: a unified lattice model for static
                analysis of programs by construction or approximation of
                fixpoints},
   pages =     {238--252},
   booktitle = {Conference Record of the Fourth Annual ACM SIGPLAN-SIGACT
                Symposium on Principles of Programming Languages},
   address =   {Los Angeles, California},
   publisher = {ACM Press, New York, NY},
   year =      1977,
}

@inproceedings{CousotCousot79,
   author =    {Cousot, P{.} and Cousot, R{.}},
   title =     {Systematic Design of Program Analysis Frameworks},
   pages =     {269--282},
   booktitle = {Conference Record of the Sixth Annual ACM Symposium on
                Principles of Programming Languages},
   address =   {San Antonio, Texas},
   publisher = {ACM Press, New York, NY},
   year =      1979,
}

@misc{jolicoeurmartineau2025morerecursivereasoningtiny,
      title={Less is More: Recursive Reasoning with Tiny Networks}, 
      author={Alexia Jolicoeur-Martineau},
      year={2025},
      eprint={2510.04871},
      archivePrefix={arXiv},
      primaryClass={cs.LG},
      url={https://arxiv.org/abs/2510.04871}, 
}

@misc{wang2025hierarchicalreasoningmodel,
      title={Hierarchical Reasoning Model}, 
      author={Guan Wang and Jin Li and Yuhao Sun and Xing Chen and Changling Liu and Yue Wu and Meng Lu and Sen Song and Yasin Abbasi Yadkori},
      year={2025},
      eprint={2506.21734},
      archivePrefix={arXiv},
      primaryClass={cs.AI},
      url={https://arxiv.org/abs/2506.21734}, 
}

@book{sutton2018rl,
  author    = {Sutton, Richard S. and Barto, Andrew G.},
  title     = {Reinforcement Learning: An Introduction},
  edition   = {2},
  publisher = {MIT Press},
  year      = {2018}
}

@inproceedings{agarwal2024onpolicy,
  author        = {Rishabh Agarwal and Nino Vieillard and Yongchao Zhou and Piotr Stanczyk and Sabela Ramos and Matthieu Geist and Olivier Bachem},
  title         = {On-Policy Distillation of Language Models: Learning from Self-Generated Mistakes},
  booktitle     = {International Conference on Learning Representations (ICLR)},
  year          = {2024},
  eprint        = {2306.13649},
  archivePrefix = {arXiv},
  primaryClass  = {cs.CL},
  url           = {https://arxiv.org/abs/2306.13649}
}

@misc{DeepSeek2026V4Release,
  author       = {DeepSeek-AI},
  title        = {DeepSeek-V4 Preview Release: Ushering in the Era of 1M Context Length},
  howpublished = {DeepSeek Official Blog},
  year         = {2026},
  month        = {April},
  url          = {https://api-docs.deepseek.com/news/news260424},
  note         = {Accessed: 2026-04-29}
}

@misc{Anthropic2026Claude46,
  author       = {Anthropic},
  title        = {Introducing Claude Opus 4.6},
  howpublished = {Anthropic News},
  year         = {2026},
  month        = {February},
  url          = {https://www.anthropic.com/news/claude-opus-4-6},
  note         = {Accessed: 2026-04-29}
}

@techreport{OpenAI2026GPT54,
  author      = {OpenAI},
  title       = {GPT-5.4 Thinking System Card},
  institution = {OpenAI},
  year        = {2026},
  month       = {March},
  url         = {https://openai.com/index/gpt-5-4-thinking-system-card/}
}

@misc{LouSotaku2026,
  author       = {Lou, Cheng},
  title        = {Sotaku: From-scratch experiments on iterative neural {Sudoku} solvers},
  howpublished = {Software, GitHub repository \url{https://github.com/chenglou/sotaku}, commit \texttt{9e13341}},
  year         = {2026},
  note         = {Accessed: 2026-05-07}
}

@inproceedings{DBLP:conf/tacas/BarbosaBBKLMMMN22,
  author       = {Haniel Barbosa and
                  Clark W. Barrett and
                  Martin Brain and
                  Gereon Kremer and
                  Hanna Lachnitt and
                  Makai Mann and
                  Abdalrhman Mohamed and
                  Mudathir Mohamed and
                  Aina Niemetz and
                  Andres N{\"{o}}tzli and
                  Alex Ozdemir and
                  Mathias Preiner and
                  Andrew Reynolds and
                  Ying Sheng and
                  Cesare Tinelli and
                  Yoni Zohar},
  editor       = {Dana Fisman and
                  Grigore Rosu},
  title        = {cvc5: {A} Versatile and Industrial-Strength {SMT} Solver},
  booktitle    = {Tools and Algorithms for the Construction and Analysis of Systems
                  - 28th International Conference, {TACAS} 2022, Held as Part of the
                  European Joint Conferences on Theory and Practice of Software, {ETAPS}
                  2022, Munich, Germany, April 2-7, 2022, Proceedings, Part {I}},
  series       = {Lecture Notes in Computer Science},
  pages        = {415--442},
  publisher    = {Springer},
  year         = {2022},
  url          = {https://doi.org/10.1007/978-3-030-99524-9\_24},
  doi          = {10.1007/978-3-030-99524-9\_24},
  bibsource    = {dblp computer science bibliography, https://dblp.org}
}

@inproceedings{wei2022chain,
  author    = {Jason Wei and Xuezhi Wang and Dale Schuurmans and Maarten Bosma and Brian Ichter and Fei Xia and Ed Chi and Quoc Le and Denny Zhou},
  title     = {Chain-of-Thought Prompting Elicits Reasoning in Large Language Models},
  booktitle = {Advances in Neural Information Processing Systems},
  volume    = {35},
  year      = {2022}
}

@article{chollet2019measure,
  author    = {Fran{\c{c}}ois Chollet},
  title     = {On the Measure of Intelligence},
  journal   = {arXiv preprint arXiv:1911.01547},
  year      = {2019}
}

@inproceedings{wang2019satnet,
  author    = {Po-Wei Wang and Priya L. Donti and Bryan Wilder and J. Zico Kolter},
  title     = {{SATN}et: Bridging Deep Learning and Logical Reasoning Using a Differentiable Satisfiability Solver},
  booktitle = {Proceedings of the 36th International Conference on Machine Learning (ICML)},
  series    = {Proceedings of Machine Learning Research},
  volume    = {97},
  pages     = {6545--6554},
  year      = {2019},
  publisher = {PMLR}
}

@article{yang2023smtlayer,
  author    = {Matt Fredrikson and Kaiji Lu and Saranya Vijayakumar and
               Somesh Jha and Vijay Ganesh and Zifan Wang},
  title     = {Learning Modulo Theories},
  journal   = {arXiv preprint arXiv:2301.11435},
  year      = {2023}
}

@inproceedings{wang2023selfconsistency,
  author    = {Xuezhi Wang and Jason Wei and Dale Schuurmans and Quoc Le and Ed H. Chi and Sharan Narang and Aakanksha Chowdhery and Denny Zhou},
  title     = {Self-Consistency Improves Chain of Thought Reasoning in Language Models},
  booktitle = {International Conference on Learning Representations (ICLR)},
  year      = {2023},
  url       = {https://arxiv.org/abs/2203.11171}
}

@inproceedings{snell2025ttc,
  author    = {Snell, Charlie and Lee, Jaehoon and Xu, Kelvin and Kumar, Aviral},
  title     = {Scaling {LLM} Test-Time Compute Optimally Can be More Effective than Scaling Parameters for Reasoning},
  booktitle = {International Conference on Learning Representations (ICLR)},
  year      = {2025}
}

@misc{chollet2026arcagi2newchallengefrontier,
      title={ARC-AGI-2: A New Challenge for Frontier AI Reasoning Systems},
      author={Francois Chollet and Mike Knoop and Gregory Kamradt and Bryan Landers and Henry Pinkard},
      year={2025},
      eprint={2505.11831},
      archivePrefix={arXiv},
      primaryClass={cs.AI},
      url={https://arxiv.org/abs/2505.11831},
}

@misc{foundation2026arcagi3newchallengefrontier,
      title={ARC-AGI-3: A New Challenge for Frontier Agentic Intelligence}, 
      author={ARC Prize Foundation},
      year={2026},
      eprint={2603.24621},
      archivePrefix={arXiv},
      primaryClass={cs.AI},
      url={https://arxiv.org/abs/2603.24621}, 
}

@InProceedings{10.1007/3-540-49519-3_7,
    author="Sheeran, Mary
    and St{\aa}lmarck, Gunnar",
    editor="Gopalakrishnan, Ganesh
    and Windley, Phillip",
    title="A Tutorial on St{\aa}lmarck's Proof Procedure for Propositional Logic",
    booktitle="Formal Methods in Computer-Aided Design",
    year="1998",
    publisher="Springer Berlin Heidelberg",
    address="Berlin, Heidelberg",
    pages="82--99",
    isbn="978-3-540-49519-2"
}

@misc{leonov2025optimizationsstaalmarckprocedure,
      title={Two Optimizations on the St\aa lmarck Procedure},
      author={Sergei Leonov and Liam Davis},
      year={2025},
      eprint={2509.16172},
      archivePrefix={arXiv},
      primaryClass={cs.LO},
      url={https://arxiv.org/abs/2509.16172},
}

@misc{graves2016act,
  title         = {Adaptive Computation Time for Recurrent Neural Networks},
  author        = {Alex Graves},
  year          = {2016},
  eprint        = {1603.08983},
  archivePrefix = {arXiv},
  primaryClass  = {cs.NE},
  url           = {https://arxiv.org/abs/1603.08983}
}

@inproceedings{dehghani2018ut,
  author    = {Mostafa Dehghani and Stephan Gouws and Oriol Vinyals and Jakob Uszkoreit and {\L}ukasz Kaiser},
  title     = {Universal Transformers},
  booktitle = {International Conference on Learning Representations (ICLR)},
  year      = {2019},
  url       = {https://arxiv.org/abs/1807.03819}
}

@inproceedings{devlin2019bert,
  author    = {Jacob Devlin and Ming-Wei Chang and Kenton Lee and Kristina Toutanova},
  title     = {{BERT}: Pre-training of Deep Bidirectional Transformers for Language Understanding},
  booktitle = {Proceedings of the 2019 Conference of the North American Chapter of the Association for Computational Linguistics: Human Language Technologies (NAACL-HLT)},
  pages     = {4171--4186},
  year      = {2019},
  url       = {https://arxiv.org/abs/1810.04805}
}

@inproceedings{austin2021d3pm,
  author    = {Jacob Austin and Daniel D. Johnson and Jonathan Ho and Daniel Tarlow and Rianne van den Berg},
  title     = {Structured Denoising Diffusion Models in Discrete State-Spaces},
  booktitle = {Advances in Neural Information Processing Systems (NeurIPS)},
  year      = {2021},
  url       = {https://arxiv.org/abs/2107.03006}
}

@inproceedings{wang2025remdm,
  title     = {Remasking Discrete Diffusion Models with Inference-Time Scaling},
  author    = {Guanghan Wang and Yair Schiff and Subham Sekhar Sahoo and Volodymyr Kuleshov},
  booktitle = {Advances in Neural Information Processing Systems (NeurIPS)},
  year      = {2025}
}

@inproceedings{chao2025prime,
  title     = {Beyond Masked and Unmasked: Discrete Diffusion Models via Partial Masking},
  author    = {Chen-Hao Chao and Wei-Fang Sun and Hanwen Liang and Chun-Yi Lee and Rahul G. Krishnan},
  booktitle = {Advances in Neural Information Processing Systems (NeurIPS)},
  year      = {2025}
}

@inproceedings{gomber2025nai,
  author    = {Shaurya Gomber and Gagandeep Singh},
  title     = {Neural Abstract Interpretation},
  booktitle = {ICLR 2025 Workshop: VerifAI: AI Verification in the Wild},
  year      = {2025},
  url       = {https://openreview.net/forum?id=WTyyhWhp4m}
}

@inproceedings{he2020lait,
  author    = {Jingxuan He and Gagandeep Singh and Markus P{\"u}schel and Martin Vechev},
  title     = {Learning Fast and Precise Numerical Analysis},
  booktitle = {Proceedings of the 41st ACM SIGPLAN Conference on Programming Language Design and Implementation (PLDI)},
  pages     = {1112--1127},
  year      = {2020},
  doi       = {10.1145/3385412.3386016}
}

@article{gu2026sail,
  author    = {Qiuhan Gu and Avaljot Singh and Gagandeep Singh},
  title     = {{SAIL}: Sound Abstract Interpreters with {LLMs}},
  journal   = {Proceedings of the ACM on Programming Languages},
  volume    = {10},
  number    = {PLDI},
  articleno = {230},
  year      = {2026},
  doi       = {10.1145/3808308},
  note      = {To appear}
}

@inproceedings{singh2018rlanalysis,
  author    = {Gagandeep Singh and Markus P{\"u}schel and Martin Vechev},
  title     = {Fast Numerical Program Analysis with Reinforcement Learning},
  booktitle = {Computer Aided Verification (CAV)},
  series    = {Lecture Notes in Computer Science},
  volume    = {10981},
  pages     = {211--229},
  publisher = {Springer},
  year      = {2018},
  doi       = {10.1007/978-3-319-96145-3\_12}
}

@inproceedings{cha2016widening,
  author    = {Sooyoung Cha and Sehun Jeong and Hakjoo Oh},
  title     = {Learning a Strategy for Choosing Widening Thresholds from a Large Codebase},
  booktitle = {Programming Languages and Systems (APLAS)},
  series    = {Lecture Notes in Computer Science},
  volume    = {10017},
  pages     = {25--41},
  publisher = {Springer},
  year      = {2016},
  doi       = {10.1007/978-3-319-47958-3\_2}
}

@inproceedings{zhou2025hdlm,
  title     = {Next Semantic Scale Prediction via Hierarchical Diffusion Language Models},
  author    = {Cai Zhou and Chenyu Wang and Dinghuai Zhang and Shangyuan Tong and Yifei Wang and Stephen Bates and Tommi Jaakkola},
  booktitle = {Advances in Neural Information Processing Systems (NeurIPS)},
  year      = {2025}
}

@misc{ren2020codebleumethodautomaticevaluation,
      title={CodeBLEU: a Method for Automatic Evaluation of Code Synthesis}, 
      author={Shuo Ren and Daya Guo and Shuai Lu and Long Zhou and Shujie Liu and Duyu Tang and Neel Sundaresan and Ming Zhou and Ambrosio Blanco and Shuai Ma},
      year={2020},
      eprint={2009.10297},
      archivePrefix={arXiv},
      primaryClass={cs.SE},
      url={https://arxiv.org/abs/2009.10297}, 
}

@article{su2024roformer,
  title={{RoFormer}: Enhanced transformer with rotary position embedding},
  author={Su, Jianlin and Ahmed, Murtadha and Lu, Yu and Pan, Shengfeng and Bo, Wen and Liu, Yunfeng},
  journal={Neurocomputing},
  volume={568},
  pages={127063},
  year={2024},
  publisher={Elsevier}
}

@inproceedings{mikolov2013linguistic,
  title     = {Linguistic Regularities in Continuous Space Word Representations},
  author    = {Mikolov, Tomas and Yih, Wen-tau and Zweig, Geoffrey},
  booktitle = {Proceedings of the 2013 Conference of the North American Chapter of the Association for Computational Linguistics: Human Language Technologies},
  pages     = {746--751},
  year      = {2013}
}

@inproceedings{huang2026equilibrium,
  title     = {Equilibrium Reasoners: Learning Attractors Enables Scalable Reasoning},
  author    = {Huang, Benhao and Geng, Zhengyang and Kolter, Zico},
  booktitle = {Proceedings of the 43rd International Conference on Machine Learning},
  year      = {2026},
  note      = {arXiv:2605.21488}
}

@misc{LouSotaku2026v2,
  author       = {Lou, Cheng},
  title        = {Sotaku v2: From-scratch experiments on iterative neural {Sudoku} solvers},
  howpublished = {Software, GitHub repository \url{https://github.com/chenglou/sotaku}, commit \texttt{d0ab125}},
  year         = {2026},
  note         = {Accessed: 2026-10-23}
}

\appendix

\section{Abstract Interpretation for Logical Problems}
\label{app:ai-tutorial}

This appendix gives a formal introduction to abstract interpretation. The
framework is most often presented in the setting of program analysis, where
the concrete domain is the (typically infinite) set of program states; here
we sketch the simpler instance that suffices for our purposes, in which the
concrete domain captures candidate solutions to a finite-domain logical
problem such as Sudoku.

\subsection{Lattices and Galois Connections}
\label{subsec:lattices}
A \emph{lattice} $(L, \sqsubseteq, \sqcup, \sqcap)$ is a set $L$ equipped
with a partial order $\sqsubseteq$ in which every pair $a, b \in L$ has a
least upper bound $a \sqcup b$ (the \emph{join}) and a greatest lower bound
$a \sqcap b$ (the \emph{meet}). A lattice is \emph{bounded} if it has a top
element $\top$ above every element and a bottom element $\bot$ below every
element. Following the main text, we read $\sqsubseteq$ as a precision
ordering: $a \sqsubseteq b$ means $a$ is at least as informative as $b$,
with $\bot$ inconsistent and $\top$ uninformative.

Recall that a Sudoku solution is a function $g : \{1,\ldots,9\}^2 \to
\{1,\ldots,9\}$, and write $G$ for the set of all such grids. We construct
two lattices over $G$.

Our \emph{concrete domain} is the \emph{powerset lattice} $C = (2^G,
\subseteq, \cup, \cap)$ over $G$, with $\bot = \emptyset$ and $\top = G$. An
element of $C$ is the set of complete grids still considered viable;
smaller sets carry more information about which grid is correct and
therefore sit lower in the lattice.

The \emph{grid powerset lattice} is the per-cell pointwise version, $A =
(\{1,\ldots,9\}^2 \to \mathcal{P}(\{1,\ldots,9\}))$, ordered by $a \sqsubseteq b$ iff
$a(c) \subseteq b(c)$ for every cell $c$. Meet is pointwise intersection,
join is pointwise union, $\top$ is the constant map sending every cell to
$\{1,\ldots,9\}$, and $\bot$ is identified with any map having an empty
candidate set at some cell. An element of $A$ records, for each cell, which
digits are still considered viable for that cell. By construction, the grid
powerset lattice loses all relational information across cells: any
correlation between distinct cell values present in a concrete state is
forgotten when only per-cell candidate sets are tracked.

An abstract element $a \in A$ can be read as the set of complete grids that
respect its candidate sets. We make this precise with the
\emph{concretization} map $\gamma : A \to C$,
\[
  \gamma(a) = \{\, g \in G : g(c) \in a(c) \text{ for every cell } c \,\}.
\]
Concretization is monotone: $a \sqsubseteq b$ implies $\gamma(a) \subseteq
\gamma(b)$.

Conversely, given $S \in C$ we look for the most informative abstract
element whose concretization still covers $S$ — that is, $S \subseteq
\gamma(a)$, so that abstracting away does not discard any candidate in $S$. The
\emph{abstraction} map $\alpha : C \to A$ returns this best sound
approximation, given elementwise by
\[
  \alpha(S)(c) = \{\, g(c) : g \in S \,\},
\]
the set of digits assigned to cell $c$ by some grid in $S$. The pair
$(\alpha, \gamma)$ forms a \emph{Galois connection}, written in the
standard notation
\[
  C \galois{\alpha}{\gamma} A.
\]
Here $\alpha$ is the \emph{lower adjoint} (the lower, rightward arrow) and
$\gamma$ is the \emph{upper adjoint} (the upper, leftward arrow). The two
arrows encode two round-trip conditions, each obtained by tracing a path
from one side back to itself:
\[
  S \subseteq \gamma(\alpha(S))
\]
\[
  \alpha(\gamma(a)) \sqsubseteq a.
\]
We use $\subseteq$ for the concrete order on $C$ and $\sqsubseteq$ for the
abstract order on $A$. The first condition says that going to the abstract
domain and back can only \emph{add} candidates to the original solution
set: abstraction is sound because the round trip overapproximates. The
second is more intuitive: going to the concrete and back lets us
\emph{refine} the abstract representation by stripping out structure that
no concrete element witnesses. Figures~\ref{fig:gamma-alpha}
and~\ref{fig:alpha-gamma} illustrate the two directions on $2 \times 2$
Sudoku puzzles.

\begin{figure}[htbp]
\centering
\begin{tikzpicture}[
  font=\footnotesize,
  c/.style={draw, minimum size=0.5cm, inner sep=1pt, font=\scriptsize},
  ac/.style={draw, minimum width=0.85cm, minimum height=0.5cm, inner sep=1pt,
             font=\scriptsize},
  arr/.style={-{Stealth[length=5pt]}, thick},
  lbl/.style={font=\scriptsize},
]
\matrix (gA) [matrix of nodes, nodes={c}, row sep=-\pgflinewidth,
              column sep=-\pgflinewidth] at (0, 0.65) {
  1 & 1 \\
  2 & 2 \\
};
\matrix (gB) [matrix of nodes, nodes={c}, row sep=-\pgflinewidth,
              column sep=-\pgflinewidth] at (0, -0.65) {
  1 & 2 \\
  2 & 1 \\
};
\node[lbl] at (0, 1.5) {$S \in C$};

\matrix (abs) [matrix of nodes, nodes={ac}, row sep=-\pgflinewidth,
               column sep=-\pgflinewidth] at (3.5, 0) {
  $\{1\}$ & $\{1,2\}$ \\
  $\{2\}$ & $\{1,2\}$ \\
};
\node[lbl] at (3.5, 0.85) {$\alpha(S) \in A$};

\matrix (h1) [matrix of nodes, nodes={c}, row sep=-\pgflinewidth,
              column sep=-\pgflinewidth] at (7, 1.95) {
  1 & 1 \\
  2 & 1 \\
};
\matrix (h2) [matrix of nodes, nodes={c}, row sep=-\pgflinewidth,
              column sep=-\pgflinewidth] at (7, 0.65) {
  1 & 1 \\
  2 & 2 \\
};
\matrix (h3) [matrix of nodes, nodes={c}, row sep=-\pgflinewidth,
              column sep=-\pgflinewidth] at (7, -0.65) {
  1 & 2 \\
  2 & 1 \\
};
\matrix (h4) [matrix of nodes, nodes={c}, row sep=-\pgflinewidth,
              column sep=-\pgflinewidth] at (7, -1.95) {
  1 & 2 \\
  2 & 2 \\
};
\node[lbl] at (7, 2.8) {$\gamma(\alpha(S)) \in C$};

\draw[arr] (gA.east) -- (abs.north west);
\draw[arr] (gB.east) -- (abs.south west);
\node[lbl] at (1.75, 1.2) {$\alpha$};

\draw[arr] (abs.east) -- (h1.west);
\draw[arr] (abs.east) -- (h2.west);
\draw[arr] (abs.east) -- (h3.west);
\draw[arr] (abs.east) -- (h4.west);
\node[lbl] at (5.25, 1.55) {$\gamma$};

\end{tikzpicture}
\caption{Round-trip $\gamma \circ \alpha$ on $2 \times 2$ Sudoku. The set
$S$ contains two grids that differ in the right-hand column. Abstraction
drops the cell-to-cell correlation, so concretizing the result yields four
grids -- the original two plus two additional grids that satisfy the
per-cell candidate sets but were not in $S$.}
\label{fig:gamma-alpha}
\end{figure}

\begin{figure}[htbp]
\centering
\begin{tikzpicture}[
  font=\footnotesize,
  ac/.style={draw, minimum width=0.85cm, minimum height=0.5cm, inner sep=1pt,
             font=\scriptsize},
  bc/.style={draw, minimum width=0.85cm, minimum height=0.5cm, inner sep=1pt,
             font=\scriptsize, fill=red!8},
  arr/.style={-{Stealth[length=5pt]}, thick},
  lbl/.style={font=\scriptsize},
]
\matrix (a) [matrix of nodes, nodes={ac}, row sep=-\pgflinewidth,
             column sep=-\pgflinewidth] at (0, 0) {
  $\{1\}$ & $\{1,2\}$ \\
  $\{1\}$ & $\{1,2\}$ \\
};
\node[lbl] at (0, 0.85) {$a \in A$};

\node[draw, minimum width=1cm, minimum height=1cm] (empty) at (3.5, 0)
  {$\emptyset$};
\node[lbl] at (3.5, 0.85) {$\gamma(a) \in C$};

\matrix (bot) [matrix of nodes, nodes={bc}, row sep=-\pgflinewidth,
               column sep=-\pgflinewidth] at (7, 0) {
  $\emptyset$ & $\emptyset$ \\
  $\emptyset$ & $\emptyset$ \\
};
\node[lbl] at (7, 0.85) {$\alpha(\gamma(a)) = \bot$};

\draw[arr] (a.east) -- node[above, lbl] {$\gamma$} (empty.west);
\draw[arr] (empty.east) -- node[above, lbl] {$\alpha$} (bot.west);

\end{tikzpicture}
\caption{Round-trip $\alpha \circ \gamma$ on an impossible abstract
element. Column~1 of $a$ forces both rows to take the value $1$,
contradicting the column constraint, so $\gamma(a) = \emptyset$.
Re-abstracting then collapses every cell to the empty candidate set,
refining $a$ all the way to $\bot$.}
\label{fig:alpha-gamma}
\end{figure}

\subsection{Sound Function Approximation and Deduction}
\label{subsec:fn-approx}

Abstract interpretation is usually concerned not with single states but
with \emph{operations} on states -- for instance, the function that takes
a program point to its successor state. We are therefore interested in
approximating monotone functions $f : L \to L$ on a lattice. In the
abstract interpretation literature these are called \emph{transformers};
we avoid that term to prevent confusion with the transformer neural
architecture, and simply refer to monotone functions on $L$.

The intuition for monotonicity is that the function is applied uniformly
to every member of a state of knowledge: if $a$ is at least as informative
as $b$, then $f(a)$ is at least as informative as $f(b)$, because
strengthening the input can only produce a more committed output.

Given lattices $L_1$ and $L_2$, the monotone functions $L_1 \to L_2$
themselves form a lattice under the pointwise order: $f \sqsubseteq g$
iff $f(x) \sqsubseteq g(x)$ for every $x$. A Galois connection
$C \galois{\alpha}{\gamma} A$ on point lattices lifts to a Galois
connection between the function lattices $[C \to C]$ and $[A \to A]$, with
abstraction $\alpha^\sharp(f) = \alpha \circ f \circ \gamma$ and
concretization $\gamma^\sharp(f^\sharp) = \gamma \circ f^\sharp \circ
\alpha$. Figure~\ref{fig:func-galois} sketches the construction.

\begin{figure}[htbp]
\centering
\begin{tikzpicture}[
  font=\footnotesize,
  box/.style={draw, rounded corners=2pt, minimum width=1.8cm,
              minimum height=0.85cm, font=\normalsize, align=center},
  darr/.style={-{Stealth[length=4pt]}, thick, dashed},
  lbl/.style={font=\scriptsize},
]
\node[box] (c)  at (0,    1.4) {$C$};
\node[box] (a)  at (4.2,  1.4) {$A$};
\node             at (2.1,  1.4)
  {$\displaystyle \galois{\alpha}{\gamma}$};

\node[box] (cc) at (0,   -1.4) {$[C \to C]$};
\node[box] (aa) at (4.2, -1.4) {$[A \to A]$};
\node             at (2.1, -1.4)
  {$\displaystyle \galois{\alpha^\sharp}{\gamma^\sharp}$};

\draw[darr] (c.south)  -- (cc.north);
\draw[darr] (a.south)  -- (aa.north);
\node[lbl, anchor=east] at (-0.2, 0) {lifts to};
\node[lbl, anchor=west] at (4.4,  0) {lifts to};

\end{tikzpicture}
\caption{Lifting a Galois connection to monotone-function lattices. The
Galois connection between the point lattices $C$ and $A$ induces a
Galois connection between the corresponding lattices of monotone
functions, with $\alpha^\sharp(f) = \alpha \circ f \circ \gamma$ and
$\gamma^\sharp(f^\sharp) = \gamma \circ f^\sharp \circ \alpha$.}
\label{fig:func-galois}
\end{figure}

For a concrete monotone function $f$, the lifted abstraction
$\alpha^\sharp(f) = \alpha \circ f \circ \gamma$ is the most precise sound
approximation of $f$ on $A$ (traditionally called the \emph{best abstract
transformer}). It is typically intractable to compute directly, so in
practice one works with a weaker abstract function $f^\sharp$ that is
still sound, meaning $\alpha \circ f \circ \gamma \sqsubseteq f^\sharp$.

The result that justifies all of this is that fixed points
transfer~\cite{CousotCousot77-1, CousotCousot79}: if $f$ is monotone on $C$
and $f^\sharp$ is any sound abstraction, then
\[
  \alpha(\mathrm{lfp}\, f) \sqsubseteq \mathrm{lfp}\, f^\sharp
  \qquad\text{and}\qquad
  \alpha(\mathrm{gfp}\, f) \sqsubseteq \mathrm{gfp}\, f^\sharp.
\]
Computing a fixed point in the abstract domain therefore yields a sound
overapproximation of the corresponding fixed point in the concrete.

\paragraph{Logical problems and deduction} In a logical problem the goal
is solution-finding rather than the analysis of a dynamic system over
time. The function of interest is the \emph{deduction function} of the
puzzle $\phi$,
\[
  \mathrm{ded}_\phi : C \to C, \qquad
  \mathrm{ded}_\phi(S) = \{\, g \in S : g \text{ satisfies } \phi \,\},
\]
which filters a candidate set down to those grids that respect the
constraints imposed by $\phi$. (The function $\mathrm{ded}_\phi$ is a
\emph{lower closure operator}: monotone, reductive
$\mathrm{ded}_\phi(S) \subseteq S$, and idempotent.) The same pattern
applies to any logic $L$ whose formulas are interpreted over a set of
models: $\mathrm{ded}_\phi$ keeps only the models of $\phi$.

The greatest fixed point of $\mathrm{ded}_\phi$ from $\top = G$ is the set
of all solutions to $\phi$, and a single application of the best abstract
transformer $\alpha \circ \mathrm{ded}_\phi \circ \gamma$ would solve the
puzzle in one shot. Practical deduction rules -- naked singles, hidden
singles, naked pairs, X-wings, and so on -- are sound but weaker abstract
approximations of $\mathrm{ded}_\phi$: each removes only some of the
candidates the full filter would. Solving a Sudoku then amounts to
computing the greatest fixed point of an abstract deduction function in
the grid powerset lattice, starting from $\top$ and descending until no
further candidates can be removed. Depending on the strength of the
chosen rules, the fixed point is either a fully determined grid (the
puzzle is solved) or one that leaves multiple candidates open in some
cells, in which case search must be invoked to finish the job.

A dual function of interest is the \emph{abduction function}
$\mathrm{abd}_\phi : C \to C$, which extends a candidate set by adding all
grids that fail to satisfy $\phi$:
\[
  \mathrm{abd}_\phi(S) = S \cup \{\, g \in G : g \not\models \phi \,\}.
\]
Where deduction lives in the over-approximation regime we have discussed
so far -- abstract states bound the true answer from above -- abduction
lives in the dual under-approximation regime: an underapproximation of
$\mathrm{abd}_\phi$ identifies a region of the search space whose elements
are all guaranteed failures. This is the engine of conflict analysis in
modern SAT solvers~\cite{marquessilva2009cdcl}: on a failed search
branch, the solver generalizes the witnessed failure into a learned clause
that excludes a much larger region than just the assignment that failed,
and this adaptive search-space compression (clause learning) is the core
reason for the extraordinary performance of modern SAT solvers. The
clause-learning step can itself be cast as an abstract
interpretation~\cite{dsilva2013acdcl}. One motivation for this work is to
begin closing the gap between diffusion-style sampling procedures and the
advanced search techniques developed in the automated reasoning
community.

\paragraph{A note for readers familiar with program analysis}
The shape of this setup may look unfamiliar to readers fluent in
abstract interpretation as it is usually presented. Classically one
computes a \emph{least} fixed point over a monotone function encoding
program semantics -- typically the strongest-postcondition transformer
forward, or the weakest-precondition transformer backward. Here we
instead compute a \emph{greatest} fixed point over an approximation of
a lower closure operator. The difference is purely conventional and
historical: abstract interpretation is fundamentally a theory of sound
approximation and representation in inferential computation, and the
program-semantics framing is one application among several. Logical
deduction, constraint satisfaction, and the puzzle setting we use
here all fall comfortably within the same theory.

Program analyzers already compute greatest fixed points of this kind when
they evaluate a branch condition. To refine the abstract state that enters
the branch, they apply the guard's conjuncts repeatedly until the state stops
changing. Consider the guard
\begin{center}
\small\texttt{if x < y and y < z and z < 3:}
\end{center}
with integer variables in $[0, 100]$, analyzed in the interval domain. One
pass through the three conjuncts gives $z \in [2, 2]$ but leaves $y \in [1,
99]$ and $x \in [0, 99]$; the second pass gives $y \in [1, 1]$; only the
third gives $x \in [0, 0]$, and a fourth pass confirms that the state no
longer changes. Each pass is a sound deduction step in our sense, and the
refined state is the greatest fixed point of these steps below the incoming
state.

\section{Portfolios on Sudoku-Extreme}
\label{app:portfolio}

A single 4K-step model leaves about $1$ in $8{,}000$ test puzzles unsolved
within $1{,}000$ rounds and answers a few wrong (Table~\ref{tab:Sudoku-extreme}).
Independently trained models fail on largely different puzzles, so a
portfolio of models can resolve most of these failures. We train $15$
models for 4K steps on the same $1{,}000$ training puzzles, differing only
in the training seed (seeds $0$--$2$ are the 4K models of
Table~\ref{tab:Sudoku-extreme}).

\paragraph{Rules} A portfolio runs its models in a fixed order and combines
their answers with one of four rules: \emph{first} returns the first answer;
\emph{agree-$k$} runs models until $k$ of them return the same answer and
abstains if none does; \emph{majority} returns the most frequent answer of
all models and abstains on a tie. A model's answer is its answer within
$1{,}000$ rounds; on the test split, a model that times out is given
$10{,}000$ rounds instead.

\paragraph{Choosing a rule on held-out puzzles} To choose the rule and
portfolio size without the test set, we evaluate all $15$ models on
$100{,}000$ training-split puzzles that none of them was trained on, and
simulate every portfolio size $m$ over $200$ random model orders.
Table~\ref{tab:portfolio-sim} scales the resulting failure rates to the
$422{,}786$ test puzzles, against the portfolio's training compute ($m$
times the $12.7$ B200 minutes of one model).

\begin{table}[htbp]
\centering
\footnotesize
\caption{Expected wrong answers / unresolved puzzles on the $422{,}786$ test
puzzles for same-data portfolios of $m$ models, extrapolated from
simulations on $100{,}000$ held-out puzzles at $1{,}000$ rounds (mean over
$200$ random model orders). Training is the portfolio's total B200 time.}
\label{tab:portfolio-sim}
\begin{tabular}{rrccc}
\toprule
$m$ & \textbf{Training} & \textbf{first} & \textbf{agree-2} & \textbf{majority} \\
\midrule
1  & 0.2h & 3.0 / 59  & ---       & 3.0 / 59 \\
2  & 0.4h & 3.3 / 20  & 0 / 102   & 0.7 / 25 \\
3  & 0.6h & 3.4 / 11  & 0 / 38    & 0.4 / 12 \\
5  & 1.1h & 3.8 / 4.8 & 0 / 17    & 0.4 / 5.8 \\
7  & 1.5h & 3.8 / 2.6 & 0 / 8.9   & 0.1 / 3.4 \\
10 & 2.1h & 3.8 / 1.6 & 0 / 5.1   & 0 / 1.6 \\
13 & 2.8h & 3.8 / 0.5 & 0 / 4.4   & 0 / 0.5 \\
15 & 3.2h & 3.8 / 0   & 0 / 4.2   & 0 / 0 \\
\bottomrule
\end{tabular}
\end{table}

Taking the first answer quickly removes unresolved puzzles but keeps every
wrong answer. Requiring two agreeing answers removes the wrong answers
entirely in simulation: no two models returned the same wrong grid on any
held-out puzzle, so agree-2 never accepted one. Majority voting lies in
between and reaches zero failures with all $15$ models.

\paragraph{Abstention keeps portfolios cheap} A single model abstains on
$0.014\%$ of the held-out puzzles and answers wrongly on $0.0007\%$. Because
a model that has not found a solution abstains rather than guessing, a
portfolio can accept an answer as soon as it appears and run further models
only on the puzzles that the earlier ones left unresolved. With \emph{first},
a $15$-model portfolio uses $1.01$ times the model calls of a single model;
with agree-2, which waits for a second matching answer, $2.0$ times. A model
that always answered would give no such signal, so every puzzle would need
votes from several models, at $m$ times the cost of one model for majority
voting over $m$ models.

\begin{figure}[htbp]
\centering
\includegraphics[width=0.55\linewidth]{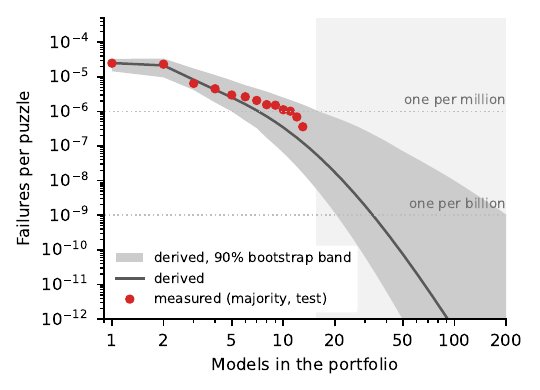}
\caption{Failure rate (wrong answers and unresolved puzzles per puzzle) of
majority voting against portfolio size on the test split, measured over
subsets of the $15$ models and derived from per-puzzle failure
probabilities, with a 90\% bootstrap band. The shaded region lies beyond the
$15$ trained models.}
\label{fig:portfolio-size}
\end{figure}

\paragraph{Failure rate beyond the test set} To estimate failure rates below
what a test set can measure, we model each puzzle $i$ as failing a model with
probability $p_i$: the model returns no correct answer within $10{,}000$
rounds. A failure is a wrong answer with probability $r = 0.41$, the share of
the test failures that are wrong answers, and an abstention otherwise. We fit
a $\mathrm{Beta}(a, b)$ prior on $p$ by maximum likelihood to the number of
the $15$ models that fail each test puzzle ($a = 1.2 \times 10^{-4}$,
$b = 4.6$), and update it with each puzzle's own count to a posterior
$\mathrm{Beta}(a + f_i, b + 15 - f_i)$. A new model is drawn from this
posterior, independently of the others. Majority voting over $m$ new models
fails a puzzle when none of them answers correctly, or when exactly one does
and another returns a wrong grid (a tie, assuming distinct wrong grids). We
compute this probability in closed form, which equals the limit of
simulating many drawn models, and average it over the test puzzles.
Figure~\ref{fig:portfolio-size} compares the derived rate with majority
voting measured over subsets of the $15$ models. The derived rate is within
$30\%$ of the measured one up to $5$ models and up to $3.6$ times lower from $8$ to
$13$ models. It falls below one failure per million puzzles at $8$ models
and below one per billion at $34$. A bootstrap over puzzles and models
(resampled with replacement, prior refit each time) puts these at $6$ to
$17$ models and at $21$ to more than $200$ models (90\% intervals): the
estimate is precise only within the range of measured portfolio sizes. As an
out-of-sample check, the same model fit on the $100{,}000$ held-out puzzles
at $1{,}000$ rounds predicts majority voting measured on the test split at
$1{,}000$ rounds within a factor of $0.7$ to $1.1$ for $1$ to $13$ models.

\paragraph{Test results} We then evaluate three disjoint $5$-model
portfolios and all $15$ models on the full test split
(Table~\ref{tab:portfolio-test}). With agree-2, one $5$-model portfolio and
the $15$-model portfolio solve every test puzzle with no wrong answer.
Agreement is not a guarantee of correctness, however: on $3$ of the
$422{,}786$ test puzzles, two or three models returned the same wrong grid,
and one of these is the wrong answer of portfolio seeds $10$--$14$ under
agree-2. Agree-$k$ also depends on the order in which the models run: over
random orders of the $15$ models, agree-2 solves every test puzzle in $60\%$
of orders and agree-3 in $82\%$, while majority voting over $14$ or $15$
models solves every test puzzle for every choice of models.

\begin{table}[htbp]
\centering
\footnotesize
\caption{Same-data portfolios on all $422{,}786$ Sudoku-Extreme test
puzzles: wrong answers / unresolved puzzles (accuracy \%).}
\label{tab:portfolio-test}
\begin{tabular}{lcccc}
\toprule
\textbf{Portfolio} & \textbf{first} & \textbf{agree-2} & \textbf{agree-3} & \textbf{majority} \\
\midrule
seeds 0--4   & 2 / 0 (99.9995) & 0 / 0 (\textbf{100})   & 0 / 2 (99.9995) & 0 / 0 (\textbf{100}) \\
seeds 5--9   & 4 / 0 (99.9991) & 0 / 2 (99.9995)        & 0 / 4 (99.9991) & 0 / 1 (99.9998) \\
seeds 10--14 & 3 / 0 (99.9993) & 1 / 2 (99.9993)        & 0 / 6 (99.9986) & 1 / 1 (99.9995) \\
all 15       & 2 / 0 (99.9995) & 0 / 0 (\textbf{100})   & 0 / 0 (\textbf{100}) & 0 / 0 (\textbf{100}) \\
\bottomrule
\end{tabular}
\end{table}

\section{Ablation Study on Sudoku-Extreme}
\label{app:ablations}

The Sudoku-Extreme LDT configuration combines several design choices: a
weight-tied backbone of $4$ layers unrolled for $L = 16$ internal
iterations, supervision of every internal iteration, an auxiliary
per-cell cross-entropy $\mathcal{L}_{\text{CE}}$, an asymmetric
candidate-elimination loss, and the CLS conflict head. This appendix
changes one of these choices at a time and measures the effect.

\paragraph{Protocol} Every arm uses the Sudoku-Extreme configuration of
Appendix~\ref{app:sudoku-hparams} ($d = 128$, $4$ layers, $L = 16$,
batch size $512$, the same $1{,}000$ training puzzles with dataset-level
augmentation) with two changes: training stops at $2{,}000$ steps
instead of $4{,}000$, and training also applies per-step augmentation
(Appendix~\ref{app:hparams}). We call this configuration the reference.
At $2{,}000$ steps the reference fails on about $1$ in $1{,}700$ test
puzzles, which leaves enough failures to compare arms at half the
training cost of the $4{,}000$-step model. Each arm changes the
settings listed in its table row and nothing else. Arms are trained
with seeds $0$, $1$ and $2$; arms with $4\times$ the reference's
training compute or the full training split use seeds $0$ and $1$.
Training compute is counted in FLOPs as in Appendix~\ref{app:compute};
the reference uses $9.35$ PFLOPs. All arms use the inference settings
of Table~\ref{tab:Sudoku-extreme} ($64$ chains, $1{,}000$ rounds,
$\theta_{\text{elim}} = 0.1$) unless a row changes them.

Each seed is evaluated with the sequential rule of
Appendix~\ref{app:eval-protocol} on the same fixed ordering of the
test split: evaluation stops at $100$ failures (wrong answers plus
timeouts) or once the $95\%$ accuracy interval has half-width at most
$0.1$ percentage points, and in any case after $4{,}096$ puzzles. For
the reference (except in Table~\ref{tab:abl-supervision}), the
loop-count arms, both $\lambda_{\text{ce}}$ arms and the two
$32\times$-asymmetry arms, the cap is $10{,}240$ puzzles instead. Arms
with low accuracy stop early at $100$ failures per seed, so their
accuracies have wider intervals.

Each table reports, per arm: $n$, the number of test puzzles
evaluated, summed over seeds; accuracy, the percentage of these
puzzles solved (mean $\pm$ standard deviation over seeds); $\Delta$, the accuracy
difference from the reference in percentage points, paired by puzzle
within each seed over the puzzles both runs evaluated, pooled over
seeds, with a $95\%$ interval in brackets; calls per solve
(``Calls''), the number of batched
model calls spent on all evaluated puzzles divided by the number
solved; and the numbers of wrong answers and timeouts, summed over
seeds. A wrong answer is a returned grid that differs from the
solution; a timeout is a puzzle for which no chain returns a grid
within the round budget.

\subsection{Is recurrence necessary?}
\label{subsec:abl-recurrence}

We vary the number of internal iterations $L$ and keep the
weight-tied $4$-layer, $d = 128$ backbone ($\approx 800$K parameters)
and all other settings fixed (Table~\ref{tab:abl-loops}). Every row
has the same parameter count and the same $2{,}000$ training steps;
the computation per model call, and therefore the training compute,
is proportional to $L$.

\begin{table}[htbp]
\centering
\footnotesize
\setlength{\tabcolsep}{4pt}
\caption{Number of internal iterations $L$ at a fixed parameter budget
($\approx 800$K), $2{,}000$ training steps, $3$ seeds. Training compute
is relative to the reference ($L = 16$).}
\label{tab:abl-loops}
\begin{tabular}{crrrrrrr}
\toprule
\textbf{$L$} & \textbf{Compute} & \textbf{$n$} & \textbf{Acc.\ \%} & \textbf{$\Delta$ (pp)} & \textbf{Calls} & \textbf{Wrong} & \textbf{Timeout} \\
\midrule
1             & $0.06\times$ & $578$       & $47.86 \pm 4.30$ & $-51.90$ {\scriptsize[$-55.98$, $-47.83$]} & $169$  & $0$ & $300$ \\
2             & $0.13\times$ & $4{,}781$   & $93.35 \pm 1.80$ & $-6.21$ {\scriptsize[$-6.90$, $-5.52$]}    & $31.6$ & $0$ & $300$ \\
4             & $0.25\times$ & $30{,}720$  & $99.68 \pm 0.09$ & $-0.26$ {\scriptsize[$-0.32$, $-0.19$]}    & $7.7$  & $0$ & $97$ \\
8             & $0.5\times$  & $23{,}200$  & $\mathbf{99.98 \pm 0.01}$ & $+0.04$ {\scriptsize[$+0.00$, $+0.07$]}    & $3.5$  & $0$ & $4$ \\
16 (reference) & $1\times$   & $30{,}720$  & $99.94 \pm 0.02$ & ---                           & $3.9$  & $0$ & $18$ \\
32            & $2\times$    & $30{,}720$  & $99.91 \pm 0.04$ & $-0.03$ {\scriptsize[$-0.07$, $+0.01$]}    & $3.7$  & $0$ & $27$ \\
\bottomrule
\end{tabular}
\end{table}

Accuracy rises from $47.9\%$ at $L = 1$ to $93.4\%$ at $L = 2$ and
$99.7\%$ at $L = 4$. From $L = 8$ on it stays between $99.9\%$ and
$100\%$: $L = 8$ is $0.04$ points above the reference, and $L = 32$
is not distinguishable from it. Calls per solve fall from $169$ at
$L = 1$ to between $3.5$ and $3.9$ for $L \geq 8$, so more iterations
per call means fewer guesses per solved puzzle. No loop count returns a
wrong answer. Rows with fewer iterations also use less training
compute; Section~\ref{subsec:abl-compute} removes this difference.

\subsection{Can more training match the performance of recurrence?}
\label{subsec:abl-compute}

We retrain models with $L = 1$ and $L = 2$ for more optimizer steps,
so that their total training compute matches the reference ($1\times$)
or is four times larger ($4\times$). One $4\times$ arm also replaces
the $1{,}000$ training puzzles with the full training split
($\approx 3.8$M puzzles). Table~\ref{tab:abl-escalation} lists the
arms; the $2{,}000$-step rows for $L = 1$ and $L = 2$ are repeated from
Table~\ref{tab:abl-loops}.

\begin{table}[htbp]
\centering
\footnotesize
\setlength{\tabcolsep}{3pt}
\caption{Models with $L = 1$ or $L = 2$ trained with more optimizer
steps. Training compute is relative to the reference. ``Full split''
trains on the full training split instead of $1{,}000$ puzzles. $3$ seeds; rows marked $^{\ast}$ use $2$.}
\label{tab:abl-escalation}
\begin{tabular}{lrrrrrrr}
\toprule
\textbf{Model, steps} & \textbf{Compute} & \textbf{$n$} & \textbf{Acc.\ \%} & \textbf{$\Delta$ (pp)} & \textbf{Calls} & \textbf{Wrong} & \textbf{Timeout} \\
\midrule
$L = 2$, $2$K                & $0.13\times$ & $4{,}781$  & $93.35 \pm 1.80$ & $-6.21$ {\scriptsize[$-6.90$, $-5.52$]}    & $31.6$ & $0$ & $300$ \\
$L = 2$, $16$K               & $1\times$    & $12{,}288$ & $99.81 \pm 0.09$ & $-0.11$ {\scriptsize[$-0.20$, $-0.03$]}    & $6.7$  & $1$ & $22$ \\
\midrule
$L = 1$, $2$K                & $0.06\times$ & $578$      & $47.86 \pm 4.30$ & $-51.90$ {\scriptsize[$-55.98$, $-47.83$]} & $169$  & $0$ & $300$ \\
$L = 1$, $32$K               & $1\times$    & $2{,}902$  & $89.15 \pm 2.69$ & $-10.27$ {\scriptsize[$-11.38$, $-9.16$]}  & $40.3$ & $0$ & $300$ \\
$L = 1$, $128$K$^{\ast}$ & $4\times$    & $6{,}017$  & $96.59 \pm 0.78$ & $-3.26$ {\scriptsize[$-3.71$, $-2.80$]}    & $21.0$ & $1$ & $199$ \\
$L = 1$, $128$K, full split$^{\ast}$ & $4\times$    & $6{,}900$  & $97.62 \pm 1.68$ & $-2.07$ {\scriptsize[$-2.42$, $-1.72$]}    & $18.8$ & $0$ & $149$ \\
\midrule
Reference ($L = 16$, $2$K) & $1\times$    & $30{,}720$ & $\mathbf{99.94 \pm 0.02}$ & ---                           & $3.9$  & $0$ & $18$ \\
\bottomrule
\end{tabular}
\end{table}

At matched training compute, $L = 2$ reaches $99.81\%$, $0.11$ points
below the reference, with $1.7\times$ the calls per solve and one wrong
answer. $L = 1$ reaches $89.1\%$ at matched compute and $96.6\%$ at
four times the compute. Training on the full split instead of $1{,}000$
puzzles raises this to $97.6\%$, still $2.1$ points below the
reference, with $4.8\times$ its calls per solve. In every
configuration we tested, a model with a single iteration stays below
the recurrent reference.

\subsection{Can a different non-recurrent architecture substitute?}
\label{subsec:abl-architecture}

A non-recurrent transformer cannot apply the same weights several times
per call, so its computation per call is fixed by its parameter count.
We test two ways of compensating. First, we keep $\approx 800$K
parameters and redistribute them across depth and width, with one
iteration ($L = 1$) and $2{,}000$ training steps
(Table~\ref{tab:abl-shapes}). The $4 \times 128$ row is the $L = 1$ row
of Table~\ref{tab:abl-loops}. Each of these models uses about $0.07\times$
the reference's training compute.

\begin{table}[htbp]
\centering
\footnotesize
\setlength{\tabcolsep}{3pt}
\caption{Non-recurrent models ($L = 1$) at the reference's parameter
budget, varying depth and width. $2{,}000$ training steps, $3$ seeds.}
\label{tab:abl-shapes}
\begin{tabular}{lrrrrrrr}
\toprule
\textbf{Layers $\times$ width} & \textbf{Params} & \textbf{$n$} & \textbf{Acc.\ \%} & \textbf{$\Delta$ (pp)} & \textbf{Calls} & \textbf{Wrong} & \textbf{Timeout} \\
\midrule
$4 \times 128$   & 0.80M & $578$      & $47.86 \pm 4.30$  & $-51.90$ {\scriptsize[$-55.98$, $-47.83$]} & $169$  & $0$  & $300$ \\
$8 \times 92$    & 0.83M & $1{,}229$  & $73.23 \pm 8.78$  & $-24.41$ {\scriptsize[$-26.81$, $-22.01$]} & $80.7$ & $9$  & $291$ \\
$16 \times 64$   & 0.80M & $525$      & $41.42 \pm 10.85$ & $-57.14$ {\scriptsize[$-61.38$, $-52.91$]} & $189$  & $39$ & $261$ \\
$32 \times 44$   & 0.76M & $453$      & $30.53 \pm 17.80$ & $-66.23$ {\scriptsize[$-70.58$, $-61.87$]} & $492$  & $35$ & $265$ \\
\midrule
Reference ($L = 16$) & 0.80M & $30{,}720$ & $\mathbf{99.94 \pm 0.02}$ & --- & $3.9$ & $0$ & $18$ \\
\bottomrule
\end{tabular}
\end{table}

Doubling the depth to $8$ layers raises accuracy from $47.9\%$ to
$73.2\%$. Deeper and narrower models ($16 \times 64$, $32 \times 44$)
are less accurate than $8 \times 92$, vary widely across seeds, and
return wrong answers ($74$ in total over the two arms); the recurrent
reference returns none.

Second, we give non-recurrent models more parameters and scale the
optimizer steps so that each arm uses about the reference's training
compute; one arm uses $4\times$ the compute and the full training split
(Table~\ref{tab:abl-params}).

\begin{table}[htbp]
\centering
\footnotesize
\setlength{\tabcolsep}{2pt}
\caption{Non-recurrent models ($L = 1$) with larger parameter budgets.
Training compute is relative to the reference ($4 \times 128$, $L = 16$,
$2$K steps). ``Full'' trains on the full training split instead of
$1{,}000$ puzzles. $3$ seeds; rows marked $^{\ast}$ use $2$.}
\label{tab:abl-params}
\begin{tabular}{lrrrrrrrr}
\toprule
\textbf{Layers $\times$ width, steps} & \textbf{Params} & \textbf{Compute} & \textbf{$n$} & \textbf{Acc.\ \%} & \textbf{$\Delta$ (pp)} & \textbf{Calls} & \textbf{Wrong} & \textbf{Timeout} \\
\midrule
$8 \times 128$, $16$K   & 1.6M & $1\times$    & $12{,}288$ & $99.40 \pm 0.41$ & $-0.53$ {\scriptsize[$-0.67$, $-0.39$]}    & $11.1$ & $25$ & $49$ \\
$16 \times 128$, $8$K   & 3.2M & $1\times$    & $12{,}288$ & $99.69 \pm 0.28$ & $-0.24$ {\scriptsize[$-0.35$, $-0.13$]}    & $7.0$  & $17$ & $21$ \\
$4 \times 256$, $8$K    & 3.2M & $0.95\times$ & $781$      & $61.45 \pm 2.85$ & $-38.41$ {\scriptsize[$-41.82$, $-35.00$]} & $111$  & $0$  & $300$ \\
$16 \times 128$, $32$K, full$^{\ast}$ & 3.2M & $4\times$ & $8{,}192$ & $99.78 \pm 0.03$ & $-0.15$ {\scriptsize[$-0.26$, $-0.03$]} & $2.6$ & $8$ & $10$ \\
\midrule
Reference ($L = 16$) & 0.80M & $1\times$ & $30{,}720$ & $\mathbf{99.94 \pm 0.02}$ & --- & $3.9$ & $0$ & $18$ \\
\bottomrule
\end{tabular}
\end{table}

At matched training compute, deep non-recurrent models come within
$0.2$--$0.5$ points of the reference: $99.40\%$ with $8$ layers
($1.6$M parameters) and $99.69\%$ with $16$ layers ($3.2$M). The
$16$-layer model with $4\times$ the compute and the full training split
reaches $99.78\%$, $0.15$ points below the reference, and needs fewer
calls per solve ($2.6$ vs.\ $3.9$). All three deep models return wrong
answers ($25$, $17$ and $8$), whereas the recurrent reference returns
none in $30{,}720$ puzzles. Parameter count alone does not determine
accuracy: the $4 \times 256$ model has as many parameters as the
$16$-layer model and solves $61.5\%$. Depth of computation does, and
recurrence provides it with $0.8$M parameters, $1\times$ training
compute and $1{,}000$ training puzzles.

\subsection{Does supervising every internal iteration matter?}
\label{subsec:abl-supervision}

The reference applies the loss to the output of each of the $16$
internal iterations and averages over iterations; at inference only
the final iteration is read out. The ablated arm applies the loss to
the final iteration only and is otherwise identical
(Table~\ref{tab:abl-supervision}). Both arms are evaluated with the
$4{,}096$-puzzle cap, so the reference row covers $4{,}096$ puzzles
per seed.

\begin{table}[htbp]
\centering
\footnotesize
\setlength{\tabcolsep}{4pt}
\caption{Supervising all internal iterations vs.\ only the final one.
$2{,}000$ training steps, $3$ seeds, $4{,}096$ test puzzles per seed.}
\label{tab:abl-supervision}
\begin{tabular}{lrrrrrr}
\toprule
\textbf{Supervision} & \textbf{$n$} & \textbf{Acc.\ \%} & \textbf{$\Delta$ (pp)} & \textbf{Calls} & \textbf{Wrong} & \textbf{Timeout} \\
\midrule
All 16 iterations (reference) & $12{,}288$ & $\mathbf{99.93 \pm 0.00}$ & ---                        & $3.9$ & $0$  & $9$ \\
Final iteration only          & $12{,}288$ & $99.72 \pm 0.04$ & $-0.20$ {\scriptsize[$-0.31$, $-0.10$]} & $6.2$ & $16$ & $18$ \\
\bottomrule
\end{tabular}
\end{table}

Supervising only the final iteration lowers accuracy by $0.20$ points,
increases calls per solve from $3.9$ to $6.2$, and produces $16$ wrong
answers in $12{,}288$ puzzles, against none for the reference.

\subsection{Does the auxiliary cross-entropy matter?}
\label{subsec:abl-ce}

We set the weight $\lambda_{\text{ce}}$ of the auxiliary per-cell
cross-entropy to $0$ or $1$ instead of $0.2$
(Table~\ref{tab:abl-ce}).

\begin{table}[htbp]
\centering
\footnotesize
\setlength{\tabcolsep}{4pt}
\caption{Weight $\lambda_{\text{ce}}$ of the auxiliary cross-entropy.
$2{,}000$ training steps, $3$ seeds.}
\label{tab:abl-ce}
\begin{tabular}{crrrrrr}
\toprule
$\boldsymbol{\lambda}_{\textbf{ce}}$ & \textbf{$n$} & \textbf{Acc.\ \%} & \textbf{$\Delta$ (pp)} & \textbf{Calls} & \textbf{Wrong} & \textbf{Timeout} \\
\midrule
0               & $395$      & $24.02 \pm 1.88$ & $-75.95$ {\scriptsize[$-80.16$, $-71.73$]} & $375$ & $0$ & $300$ \\
0.2 (reference) & $30{,}720$ & $99.94 \pm 0.02$ & ---                           & $3.9$ & $0$ & $18$ \\
1               & $23{,}100$ & $\mathbf{99.98 \pm 0.01}$ & $+0.04$ {\scriptsize[$+0.00$, $+0.07$]}    & $1.6$ & $2$ & $2$ \\
\bottomrule
\end{tabular}
\end{table}

Without the auxiliary cross-entropy the model solves $24.0\%$ of the
puzzles, and all of its failures are timeouts. With
$\lambda_{\text{ce}} = 1$ accuracy is $0.04$ points above the reference
and calls per solve drop from $3.9$ to $1.6$, but the model returns $2$
wrong answers in $23{,}100$ puzzles, against none for the reference.
At the main $4$K-step budget the advantage disappears: on $12{,}000$
validation puzzles ($3$ seeds), $\lambda_{\text{ce}} = 0.2$ solves
$99.992\%$ with no wrong answers and $\lambda_{\text{ce}} = 1$ solves
$99.967\%$ with $2$ wrong answers, so we keep $\lambda_{\text{ce}} = 0.2$.

\subsection{What enforces soundness?}
\label{subsec:abl-soundness}

Two mechanisms are meant to keep the model from returning a wrong
answer: the asymmetric elimination loss, which weights false
eliminations $w^{+}/w^{-} = 8\times$ more than false retentions
($w^{+} = 4$, $w^{-} = 0.5$), and the CLS conflict head. We change the
ratio by setting $w^{+}$ to $0.5$, $1$ or $16$ with $w^{-} = 0.5$ fixed
($1\times$, $2\times$, $32\times$). Separately, we remove the conflict
head: the model has no CLS token and $\lambda_{\text{cls}} = 0$, so at
inference a conflict is detected only when a cell has no candidates
left. The reference evaluates every ratio at $\theta_{\text{elim}} =
0.1$, which is close to $1/(1 + w^{+}/w^{-}) = 0.11$ for $8\times$. The
last three rows re-evaluate the $1\times$, $2\times$ and $32\times$
checkpoints with $\theta_{\text{elim}} = 1/(1 + w^{+}/w^{-})$, the
threshold that matches each ratio (Table~\ref{tab:abl-soundness}).
``Unsound'' is the percentage of the model's candidate eliminations
that remove the reference digit, averaged over seeds.

\begin{table}[htbp]
\centering
\footnotesize
\setlength{\tabcolsep}{2.5pt}
\caption{Soundness mechanisms: elimination-loss asymmetry $w^{+}/w^{-}$,
the CLS conflict head, and the elimination threshold
$\theta_{\text{elim}}$. $2{,}000$ training steps, $3$ seeds.}
\label{tab:abl-soundness}
\vspace{3pt}
\resizebox{\linewidth}{!}{%
\begin{tabular}{lrrrrrrr}
\toprule
\textbf{Model} & \textbf{$n$} & \textbf{Acc.\ \%} & \textbf{$\Delta$ (pp)} & \textbf{Calls} & \textbf{Wrong} & \textbf{Timeout} & \textbf{Unsound \%} \\
\midrule
$1\times$ (symmetric)                & $403$      & $25.55 \pm 0.84$ & $-74.44$ {\scriptsize[$-78.70$, $-70.18$]} & $320$  & $2$   & $298$ & $1.39$ \\
$2\times$                            & $2{,}242$  & $85.66 \pm 4.81$ & $-13.29$ {\scriptsize[$-14.70$, $-11.88$]} & $37.4$ & $0$   & $300$ & $1.35$ \\
$8\times$, conflict head (reference) & $30{,}720$ & $99.94 \pm 0.02$ & ---                           & $3.9$  & $0$   & $18$  & $1.76$ \\
$32\times$                           & $27{,}140$ & $\mathbf{99.97 \pm 0.02}$ & $+0.02$ {\scriptsize[$-0.02$, $+0.05$]}    & $3.5$  & $0$   & $10$  & $0.14$ \\
$8\times$, no conflict head          & $409$      & $26.63 \pm 1.35$ & $-73.35$ {\scriptsize[$-77.63$, $-69.06$]} & $19.0$ & $300$ & $0$   & $0.12$ \\
\midrule
$1\times$, $\theta_{\text{elim}} = 0.5$   & $360$      & $16.64 \pm 1.82$ & $-83.33$ {\scriptsize[$-87.18$, $-79.48$]} & $562$ & $11$ & $289$ & $8.89$ \\
$2\times$, $\theta_{\text{elim}} = 0.33$  & $399$      & $24.80 \pm 0.99$ & $-75.19$ {\scriptsize[$-79.43$, $-70.95$]} & $331$ & $0$  & $300$ & $4.98$ \\
$32\times$, $\theta_{\text{elim}} = 0.03$ & $25{,}740$ & $99.94 \pm 0.07$ & $-0.02$ {\scriptsize[$-0.06$, $+0.02$]}    & $3.7$ & $12$ & $5$   & $0.01$ \\
\bottomrule
\end{tabular}}
\end{table}

The conflict head determines whether failures are abstentions or wrong
answers. Without it, every failure is a wrong answer: the model returns
an incorrect grid on $300$ of $409$ puzzles and never times out. The
loss asymmetry determines accuracy. At $2\times$ accuracy drops to
$85.7\%$ and at $1\times$ to $25.6\%$, with almost all failures being
timeouts. Raising the ratio from $8\times$ to $32\times$ does not change
accuracy measurably ($+0.02$ points, interval includes $0$), returns no
wrong answers, and lowers the unsound-elimination rate from $1.76\%$ to
$0.14\%$. Evaluating each checkpoint at the threshold matched to its
ratio lowers accuracy for $1\times$ and $2\times$ (to $16.6\%$ and
$24.8\%$) and adds $12$ wrong answers for $32\times$. The low accuracy
of the $1\times$ and $2\times$ models therefore comes from the trained
weights, not from a mismatched inference threshold.

\subsection{Do hinted and corrupted training puzzles help?}
\label{subsec:abl-inputs}

A puzzle entering the training pool is the raw puzzle ($20\%$), the puzzle
with a random share of its blanks filled correctly ($55\%$), or the puzzle
with a random share filled and some of those digits wrong ($25\%$). Building
these inputs is domain independent: it uses only a solution and the set of
values a cell can take, not the rules of the puzzle. We train
the 4K-step recipe of Table~\ref{tab:Sudoku-extreme} ($1{,}000$ training
puzzles, at most $1{,}000$ search rounds) with two other mixes, $3$ seeds
each: raw puzzles only, so that every unsatisfiable state in training comes
from the model's own branching, and raw puzzles with correct hints only
($20\%$ / $80\%$). The arms are evaluated with the sequential stop rule
($5{,}900$ to $37{,}300$ test puzzles per seed); Table~\ref{tab:abl-inputs}
compares each with the Table~\ref{tab:Sudoku-extreme} models on the same
puzzles and seeds.

\begin{table}[htbp]
\centering
\footnotesize
\caption{Training inputs on Sudoku-Extreme (4K steps, $3$ seeds), each arm
against the Table~\ref{tab:Sudoku-extreme} models (trained with hinted and
corrupted puzzles) on the same test puzzles and seeds: paired accuracy
difference with a $95\%$ interval, and wrong answers and timeouts on those
puzzles, summed over seeds, as arm / reference.}
\label{tab:abl-inputs}
\begin{tabular}{lrrrr}
\toprule
\textbf{Training inputs} & \textbf{Puzzles} & \textbf{$\Delta$ (pp)} & \textbf{Wrong} & \textbf{Timeouts} \\
\midrule
Raw and correct hints & $43{,}900$ & $-0.08$ {\scriptsize$[-0.11, -0.05]$} & $33$ / $0$ & $7$ / $5$ \\
Raw only & $84{,}500$ & $-0.13$ {\scriptsize$[-0.15, -0.10]$} & $59$ / $0$ & $59$ / $11$ \\
\bottomrule
\end{tabular}
\end{table}

Without corrupted inputs the model returns more wrong answers: $33$ in
$43{,}900$ puzzles with correct hints and $59$ in $84{,}500$ with raw
puzzles only, against none for the reference on the same puzzles. The
states the model reaches through its own branching do not give the conflict
head enough unsatisfiable examples at this budget. Removing the correct
hints as well also raises the number of timeouts.

\section{Search-Policy Ablations on Sudoku-Extreme}
\label{app:search-ablations}

When deduction stops removing candidates, the solver branches: it picks
one cell with two or more alive candidates and pins it to one digit. The
default policy picks the cell uniformly at random and, at inference,
pins the alive candidate of that cell with the highest digit logit (argmax). During
training, the pool states are advanced with the same uniform cell choice
but a digit sampled from the softmax at $\tau_{\text{decide}} = 1.5$.
When a chain detects a conflict, it restarts from the puzzle. This
appendix varies three parts of this policy: the cell and digit rules at
inference, the digit rule during training, and where a chain restarts
after a conflict.

\paragraph{Protocol} We use the $2{,}000$-step ablation baseline of
Appendix~\ref{app:ablations} (the Sudoku-Extreme configuration with
per-step training augmentation, stopped at $2{,}000$ steps) and the same
configuration stopped at $1{,}000$ steps. The $1{,}000$-step model
deduces less, so it branches more often and differences between search
policies are larger. Each model is trained with $3$ seeds. All arms in
Tables~\ref{tab:search-decision} and~\ref{tab:search-backtrack} only
change the solver and reuse these checkpoints; the two arms in
Table~\ref{tab:search-train} that train with argmax digits are separate
$2{,}000$-step runs with the same $3$ seeds. The solver uses $64$ chains
per puzzle and at most $1{,}000$ rounds. Evaluation walks the fixed
ordering of the test split (Appendix~\ref{app:eval-protocol}) for at
most $4{,}096$ puzzles per seed and stops early at the $100$th failure,
so low-accuracy arms are evaluated on fewer puzzles (at least $140$ per
seed). The default policy on the $2{,}000$-step model and the
backtracking arms with exclusion are evaluated on up to $10{,}240$
puzzles per seed, with the same failure rule and the confidence-sequence
rule of Appendix~\ref{app:eval-protocol}. Accuracy is the mean $\pm$
standard deviation over seeds. $\Delta$ is the accuracy difference from
the default policy on the same model, paired by puzzle and pooled over
seeds, with a $95\%$ interval. \emph{Calls} is the number of batched
model calls divided by the number of puzzles solved; one call processes
$512$ chain states ($8$ puzzles with $64$ chains each).

\subsection{Cell and digit choice at inference}

The cell rules are: \emph{uniform}, a uniformly random cell;
\emph{fewest candidates}, a cell with the fewest alive candidates;
\emph{lowest entropy} and \emph{highest entropy}, a cell whose softmax
over alive candidates has the lowest or highest entropy. Ties break
uniformly at random. The digit rules are: \emph{sample}, a softmax
sample at $\tau_{\text{decide}} = 1.5$; \emph{argmax}, the
highest-scoring alive candidate; \emph{ranked}, where chain $k$ of a
puzzle ($k = 1, \dots, 64$) takes the $k$-th highest-scoring alive
candidate, or the lowest-scoring one if fewer than $k$ remain. Under
the ranked rule, most of the $64$ chains therefore take the
lowest-scoring candidate. Table~\ref{tab:search-decision} reports all
$12$ combinations on both models.

\begin{table}[htbp]
\centering
\footnotesize
\caption{Cell and digit rules at inference on Sudoku-Extreme, for models
trained for $2{,}000$ and $1{,}000$ steps ($3$ seeds each). Uniform cell
with argmax digit is the default policy. Calls are batched model calls
per solved puzzle.}
\label{tab:search-decision}
\begin{tabular}{llrrrr}
\toprule
& & \multicolumn{2}{c}{\textbf{2{,}000 training steps}}
& \multicolumn{2}{c}{\textbf{1{,}000 training steps}} \\
\cmidrule(lr){3-4}\cmidrule(lr){5-6}
\textbf{Cell} & \textbf{Digit} & \textbf{Accuracy \%} & \textbf{Calls}
& \textbf{Accuracy \%} & \textbf{Calls} \\
\midrule
Uniform           & Argmax (default) & $\mathbf{99.9 \pm 0.0}$ & $3.9$   & $\mathbf{99.2 \pm 0.2}$ & $9.8$   \\
Uniform           & Sample           & $98.6 \pm 0.8$ & $13.1$  & $82.6 \pm 3.3$ & $50.5$  \\
Uniform           & Ranked           & $72.6 \pm 3.3$ & $80.2$  & $48.0 \pm 0.8$ & $166.6$ \\
\midrule
Fewest candidates & Argmax           & $96.0 \pm 0.4$ & $13.9$  & $93.6 \pm 0.6$ & $21.1$  \\
Fewest candidates & Sample           & $99.7 \pm 0.1$ & $7.4$   & $96.5 \pm 1.3$ & $20.3$  \\
Fewest candidates & Ranked           & $60.1 \pm 3.0$ & $110.5$ & $45.4 \pm 6.1$ & $168.1$ \\
\midrule
Lowest entropy    & Argmax           & $80.1 \pm 1.8$ & $43.2$  & $80.7 \pm 1.4$ & $45.5$  \\
Lowest entropy    & Sample           & $99.2 \pm 0.3$ & $10.9$  & $96.9 \pm 0.9$ & $20.2$  \\
Lowest entropy    & Ranked           & $51.8 \pm 1.1$ & $140.5$ & $44.9 \pm 1.9$ & $176.4$ \\
\midrule
Highest entropy   & Argmax           & $90.9 \pm 0.2$ & $27.0$  & $70.1 \pm 1.4$ & $73.4$  \\
Highest entropy   & Sample           & $88.9 \pm 2.2$ & $38.1$  & $44.7 \pm 3.7$ & $173.7$ \\
Highest entropy   & Ranked           & $44.7 \pm 2.5$ & $162.7$ & $29.4 \pm 1.4$ & $303.4$ \\
\bottomrule
\end{tabular}
\end{table}

The default policy has the highest accuracy and the fewest calls on
both models. Every other combination is less accurate, and each paired
difference excludes zero. With a uniform cell, sampling the digit
instead of taking the argmax lowers accuracy by $1.3$ points
($[1.1, 1.6]$) on the $2{,}000$-step model and by $16.6$ points
($[14.9, 18.4]$) on the $1{,}000$-step model, and it uses $3$ to $5$
times more calls. The fewest-candidates and lowest-entropy cell rules
work better with a sampled digit than with argmax. With sampling they
come close to the default on the $2{,}000$-step model ($-0.2$ and
$-0.7$ points) and lose $2.5$ and $2.2$ points on the $1{,}000$-step
model. With argmax they lose between $3.9$ and $19.7$ points. The
ranked digit rule is the worst digit rule for every cell rule, and the
highest-entropy cell rule is the worst cell rule with sampled or ranked
digits. Almost all failures in this table are timeouts: no arm returns
more than $6$ wrong answers over its $3$ seeds.

\subsection{Digit choice during training}

The training pool advances its states with the same step operator as
the solver, so the digit rule also changes which states the model is
trained on. Table~\ref{tab:search-train} crosses the training digit rule
(sample or argmax) with the inference digit rule, on $2{,}000$-step
models with a uniform cell rule throughout.

\begin{table}[htbp]
\centering
\footnotesize
\caption{Digit rule during training and at inference on Sudoku-Extreme
($2{,}000$ training steps, uniform cell, $3$ seeds). Training with a
sampled digit and inferring with argmax is the default. $\Delta$ is
paired against the default; wrong answers and timeouts are summed over
seeds; $n$ is the number of puzzles evaluated, summed over seeds.}
\label{tab:search-train}
\begin{tabular}{llrrrrrr}
\toprule
\textbf{Training} & \textbf{Inference} & \textbf{Accuracy \%}
& \textbf{$\Delta$ (pp)} & \textbf{Calls} & \textbf{Wrong}
& \textbf{Timeouts} & \textbf{$n$} \\
\midrule
Sample & Argmax (default) & $\mathbf{99.9 \pm 0.0}$ & ---                    & $3.9$  & $0$  & $18$  & $30{,}720$ \\
Sample & Sample           & $98.6 \pm 0.8$ & $-1.34$ [$-1.55$, $-1.13$] & $13.1$ & $0$  & $174$ & $12{,}288$ \\
Argmax & Argmax           & $99.7 \pm 0.5$ & $-0.19$ [$-0.29$, $-0.09$] & $2.4$  & $32$ & $0$   & $12{,}288$ \\
Argmax & Sample           & $99.3 \pm 0.8$ & $-0.60$ [$-0.75$, $-0.45$] & $5.4$  & $51$ & $32$  & $12{,}288$ \\
\bottomrule
\end{tabular}
\end{table}

Argmax at inference is better under both training rules. For the model
trained with argmax, sampling at inference lowers accuracy by $0.42$
points ($[0.25, 0.58]$). Training with argmax lowers the calls per
solve from $3.9$ to $2.4$, but it lowers accuracy by $0.19$ points and
produces wrong answers: $32$ with argmax inference and $51$ with
sampled inference. All of these wrong answers come from one of the
three seeds; the other two seeds return none. The default trains with
a sampled digit and uses argmax only at inference.

\subsection{Backtracking after a conflict}

The default restarts a conflicting chain from the puzzle. The
alternatives keep a stack of the chain's last $64$ decisions, each with
the post-deduction state it was made from, and restore one of these
states: \emph{last}, the state before the most recent decision;
\emph{geometric}, the state $j$ decisions back with
$j \sim \mathrm{Geometric}(0.5)$, $j \geq 1$; \emph{uniform depth}, the
state before a uniformly random earlier decision. A chain with no
recorded decision restarts from the puzzle. With \emph{exclusion}, the
digit chosen at the restored decision is also removed from its cell; if
this leaves the cell empty, the chain restarts from the puzzle. We
evaluate exclusion on the $2{,}000$-step model only.
Table~\ref{tab:search-backtrack} also reports the \emph{unsound rate}:
the percentage of the model's candidate eliminations that remove the
solution's digit from a cell, over all chains. Digits removed by
exclusion are not counted. Chains that already contain a wrong decision
also count, so this rate depends on how much time the search spends
below wrong decisions as well as on the model.

\begin{table}[htbp]
\centering
\footnotesize
\caption{Restart target after a conflict on Sudoku-Extreme (uniform
cell, argmax digit, $3$ seeds). Restarting from the puzzle is the
default. $\Delta$ is paired against the default on the same model; wrong
answers and timeouts are summed over seeds; $n$ is the number of
puzzles evaluated, summed over seeds; Unsound is the percentage of
model eliminations that remove the solution's digit.}
\label{tab:search-backtrack}
\vspace{3pt}
\resizebox{\linewidth}{!}{%
\begin{tabular}{lrrrrrrr}
\toprule
\textbf{Restore} & \textbf{Accuracy \%} & \textbf{$\Delta$ (pp)}
& \textbf{Calls} & \textbf{Wrong} & \textbf{Timeouts} & \textbf{$n$}
& \textbf{Unsound \%} \\
\midrule
\multicolumn{8}{l}{\emph{$2{,}000$ training steps}} \\
Puzzle (default) & $\mathbf{99.9 \pm 0.0}$ & ---                          & $3.9$  & $\mathbf{0}$  & $18$  & $30{,}720$ & $1.76$ \\
Last             & $79.5 \pm 5.3$ & $-19.56$ [$-21.54$, $-17.57$] & $36.9$ & $11$ & $289$ & $1{,}529$  & $16.67$ \\
Geometric        & $98.3 \pm 1.5$ & $-1.49$ [$-1.72$, $-1.26$]    & $9.0$  & $6$  & $168$ & $11{,}102$ & $11.10$ \\
Uniform depth    & $99.8 \pm 0.0$ & $-0.11$ [$-0.20$, $-0.02$]    & $4.0$  & $1$  & $22$  & $12{,}288$ & $5.47$ \\
\midrule
\multicolumn{8}{l}{\emph{$2{,}000$ training steps, with exclusion}} \\
Last             & $99.0 \pm 1.2$ & $-0.57$ [$-0.67$, $-0.47$]    & $7.3$  & $3$  & $150$ & $24{,}700$ & $7.56$ \\
Geometric        & $99.9 \pm 0.1$ & $-0.06$ [$-0.11$, $-0.02$]    & $3.9$  & $4$  & $30$  & $28{,}440$ & $5.26$ \\
Uniform depth    & $99.9 \pm 0.0$ & $0.00$ [$-0.04$, $0.04$]      & $3.1$  & $1$  & $14$  & $28{,}440$ & $3.87$ \\
\midrule
\multicolumn{8}{l}{\emph{$1{,}000$ training steps}} \\
Puzzle (default) & $99.2 \pm 0.2$ & ---                          & $9.8$  & $5$  & $94$  & $12{,}288$ & $3.90$ \\
Last             & $72.4 \pm 4.6$ & $-26.78$ [$-29.40$, $-24.16$] & $57.9$ & $31$ & $269$ & $1{,}109$  & $21.92$ \\
Geometric        & $97.1 \pm 1.0$ & $-1.97$ [$-2.31$, $-1.63$]    & $14.5$ & $40$ & $231$ & $9{,}849$  & $18.13$ \\
Uniform depth    & $\mathbf{99.4 \pm 0.3}$ & $+0.20$ [$+0.01$, $+0.39$]    & $8.2$  & $10$ & $64$  & $12{,}288$ & $13.30$ \\
\bottomrule
\end{tabular}}
\end{table}

Restoring the last decision without exclusion is the worst choice on
both models ($-19.6$ and $-26.8$ points). Without exclusion, a chain
restored to an earlier state can repeat the same decision. Restoring an
earlier state at random works better, and uniform depth is the best of
the three targets. On the $2{,}000$-step model no backtracking arm is
more accurate than the default. Uniform depth with exclusion has the
same accuracy ($0.00$ points, $[-0.04, 0.04]$) and uses $3.1$ instead of
$3.9$ calls per solve. On the $1{,}000$-step model, uniform depth gains
$0.20$ points ($[0.01, 0.39]$) and uses $8.2$ instead of $9.8$ calls,
but it returns $10$ wrong answers instead of $5$. Every backtracking arm
has a higher unsound rate than the default: $3.87\%$ to $16.67\%$
against $1.76\%$ on the $2{,}000$-step model, and $13.30\%$ to $21.92\%$
against $3.90\%$ on the $1{,}000$-step model. Because the gains in
accuracy and calls are small and the unsound rate and wrong answers
increase, the default policy restarts from the puzzle.

Search that records information from conflicts, such as stack-based
replay or tracking dead ends so that they are not revisited, is left to
future work.

\section{Inference-Time Deduction Ablations}
\label{app:deduction-ablations}

This section varies the step operator at evaluation time only. The
trained weights are fixed. We vary four settings of the operator: the
number of internal iterations of the recurrent transformer, the number
of deduction passes before each decision, the elimination threshold
$\theta_{\text{elim}}$, and the conflict threshold
$\theta_{\text{CLS}}^{\text{eval}}$. Each arm changes exactly one of
these settings and keeps the others at their defaults.

\paragraph{Checkpoints} All arms evaluate the same three checkpoints.
They are trained with the Sudoku configuration of
Section~\ref{sec:eval} for $2{,}000$ steps instead of $4{,}000$, with
per-step augmentation in the training rollouts, on the same $1{,}000$
training puzzles. The three checkpoints differ only in the training
seed ($0$, $1$, $2$). Every arm uses one evaluation seed per
checkpoint.

\paragraph{Default operator} The reference arm uses the inference
settings of the main experiments: $16$ internal iterations (the
trained value), one deduction pass per step, $\theta_{\text{elim}} =
0.1$, $\theta_{\text{CLS}}^{\text{eval}} = 0.6$, a random symmetry per
step, a uniformly random branching cell, the highest-confidence
candidate at that cell, $64$ parallel chains per puzzle, restart from
the original puzzle after a conflict, and at most $1{,}000$ steps per
puzzle.

\paragraph{Test puzzles} Every arm walks the Sudoku-Extreme test split in
the same fixed random order, so all arms see the same puzzles in the
same positions. A run stops at the first of three conditions: $100$
failures (wrong answers plus timeouts), a $95\%$ confidence-sequence
half-width of at most $0.1$ percentage points (checked every $100$
puzzles from $1{,}000$ on), or a maximum number of puzzles. The
maximum is $4{,}096$ for the deduction-pass and elimination-threshold
arms and $10{,}240$ for the default, internal-iteration and
conflict-threshold arms. The $n$ column of each table gives the number
of puzzles per seed.

\paragraph{Metrics} Accuracy is the mean $\pm$ standard deviation over
the three seeds. $\Delta$ is the accuracy difference from the default
operator, paired by puzzle on the same seed, pooled over seeds, with a
$95\%$ interval. It uses only the puzzles both runs evaluated, so for a
run stopped by the failure count it covers that run's first $n$
puzzles. Calls per solve is the number of batched model calls (one call
advances all $64$ chains of a puzzle by one forward pass) divided by
the number of solved puzzles. A pass of the operator is one call, so
an arm with more passes per step spends more calls per step. Wrong
counts accepted solutions that differ from the ground truth; timeouts
count puzzles not solved within $1{,}000$ steps. Both are summed over
seeds. Unsound is the fraction of eliminated candidates that belong to
the puzzle's solution, pooled over all chains and steps of a run and
averaged over seeds.

\subsection{Internal iterations at evaluation}

The recurrent transformer applies the same weights $16$ times per call
during training. Because the weights are shared, the number of
iterations can be set to a different value at evaluation. We evaluate
$1$, $2$, $4$, $8$ and $32$ iterations.
Table~\ref{tab:deduction-loops} gives the results.

\begin{table}[htbp]
\centering
\footnotesize
\caption{Number of internal iterations at evaluation, for checkpoints
trained with $16$. Three seeds; $n$ is the number of test puzzles per
seed. $\Delta$ is in percentage points against the $16$-iteration
default, with a $95\%$ interval.}
\label{tab:deduction-loops}
\vspace{3pt}
\resizebox{\linewidth}{!}{%
\begin{tabular}{rrrrrrrr}
\toprule
\textbf{Iterations} & $n$ & \textbf{Acc.\ \%} & $\Delta$ \textbf{(pp)}
& \textbf{Calls/solve} & \textbf{Wrong} & \textbf{Timeouts}
& \textbf{Unsound \%} \\
\midrule
1  & $104$ & $3.8 \pm 0.0$ & $-96.15$ [$-98.29$, $-94.02$] & $5{,}115$ & $0$ & $300$ & $0.33$ \\
2  & $124$--$131$ & $21.4 \pm 2.2$ & $-78.53$ [$-82.65$, $-74.42$] & $469$ & $0$ & $300$ & $0.81$ \\
4  & $749$--$1{,}474$ & $90.3 \pm 3.4$ & $-8.91$ [$-9.88$, $-7.94$] & $38.0$ & $1$ & $299$ & $1.79$ \\
8  & $10{,}240$ & $99.5 \pm 0.2$ & $-0.48$ [$-0.56$, $-0.39$] & $8.8$ & $0$ & $165$ & $1.87$ \\
16 (default) & $10{,}240$ & $\mathbf{99.9 \pm 0.0}$ & ref. & $3.9$ & $\mathbf{0}$ & $18$ & $1.76$ \\
32 & $10{,}240$ & $99.9 \pm 0.0$ & $-0.05$ [$-0.09$, $-0.01$] & $3.0$ & $1$ & $32$ & $1.97$ \\
\bottomrule
\end{tabular}}
\end{table}

Accuracy increases with the number of iterations up to $16$. With $1$
or $2$ iterations the solver times out on most puzzles, and the runs
stop after about $100$ to $130$ puzzles. With $4$ iterations accuracy
is $90.3\%$ and a solve takes $38.0$ calls. With $8$ iterations
accuracy is $0.48$ points below the default and a solve takes $8.8$
calls instead of $3.9$. With $32$ iterations a solve takes $3.0$
calls, but each call runs twice as many iterations as the default, and
accuracy is $0.05$ points lower (interval $[-0.09, -0.01]$). The
unsound rate is between $1.76\%$ and $1.97\%$ from $4$ to $32$
iterations.

\subsection{Deduction passes before each decision}

By default, a step runs the model once, eliminates candidates, and
then branches. A multi-pass step runs the model again on the reduced
state before it branches, with a new random symmetry for each pass.
Each pass is one model call. A chain stops making passes once it is
solved or in conflict. We test $2$ and $4$ passes, and repeating
passes until a pass eliminates nothing (at most $16$ passes).
Table~\ref{tab:deduction-passes} gives the results.

\begin{table}[htbp]
\centering
\footnotesize
\caption{Deduction passes before each decision. Three seeds; $n$ is
the number of test puzzles per seed. $\Delta$ is in percentage points
against the one-pass default, paired on the first $4{,}096$ puzzles,
with a $95\%$ interval.}
\label{tab:deduction-passes}
\vspace{3pt}
\resizebox{\linewidth}{!}{%
\begin{tabular}{lrrrrrrr}
\toprule
\textbf{Passes} & $n$ & \textbf{Acc.\ \%} & $\Delta$ \textbf{(pp)}
& \textbf{Calls/solve} & \textbf{Wrong} & \textbf{Timeouts}
& \textbf{Unsound \%} \\
\midrule
1 (default) & $10{,}240$ & $\mathbf{99.9 \pm 0.0}$ & ref. & $\mathbf{3.9}$ & $0$ & $18$ & $1.76$ \\
2 & $4{,}096$ & $99.9 \pm 0.1$ & $-0.03$ [$-0.10$, $+0.04$] & $7.7$ & $0$ & $13$ & $2.05$ \\
4 & $4{,}096$ & $99.9 \pm 0.1$ & $-0.05$ [$-0.12$, $+0.03$] & $13.3$ & $0$ & $15$ & $2.28$ \\
Until no change ($\le 16$) & $4{,}096$ & $99.7 \pm 0.1$ & $-0.24$ [$-0.35$, $-0.14$] & $54.6$ & $0$ & $39$ & $2.87$ \\
\bottomrule
\end{tabular}}
\end{table}

Two and four passes do not change accuracy measurably: both intervals
contain zero. They raise calls per solve from $3.9$ to $7.7$ and
$13.3$. Repeating until no change lowers accuracy by $0.24$ points and
takes $54.6$ calls per solve. The unsound rate rises with the number
of passes, from $1.76\%$ with one pass to $2.87\%$ when repeating until
no change, because each pass makes further eliminations on the output
of the previous pass. One pass per step has the lowest cost and the
lowest unsound rate, with no measurable loss in accuracy.

\subsection{Elimination and conflict thresholds}

The elimination threshold $\theta_{\text{elim}}$ removes every
candidate whose predicted probability is below it. The default is
$0.1$; we test $0.05$, $0.2$, $0.3$ and $0.5$. The conflict threshold
$\theta_{\text{CLS}}^{\text{eval}}$ declares a state inconsistent when
the conflict head's probability is above it. The default at
evaluation is $0.6$; we test $0.5$ (the value used in training
rollouts), $0.7$ and $0.8$. Table~\ref{tab:deduction-thresholds} gives
the results.

\begin{table}[htbp]
\centering
\footnotesize
\caption{Elimination threshold $\theta_{\text{elim}}$ and conflict
threshold $\theta_{\text{CLS}}^{\text{eval}}$ at evaluation. Three
seeds; $n$ is the number of test puzzles per seed. $\Delta$ is in
percentage points against the default ($\theta_{\text{elim}} = 0.1$,
$\theta_{\text{CLS}}^{\text{eval}} = 0.6$), with a $95\%$ interval.}
\label{tab:deduction-thresholds}
\vspace{3pt}
\resizebox{\linewidth}{!}{%
\begin{tabular}{lrrrrrrr}
\toprule
\textbf{Threshold} & $n$ & \textbf{Acc.\ \%} & $\Delta$ \textbf{(pp)}
& \textbf{Calls/solve} & \textbf{Wrong} & \textbf{Timeouts}
& \textbf{Unsound \%} \\
\midrule
Default & $10{,}240$ & $\mathbf{99.9 \pm 0.0}$ & ref. & $\mathbf{3.9}$ & $0$ & $18$ & $1.76$ \\
\midrule
$\theta_{\text{elim}} = 0.05$ & $4{,}096$ & $99.9 \pm 0.1$ & $-0.05$ [$-0.13$, $+0.03$] & $4.8$ & $0$ & $15$ & $0.67$ \\
$\theta_{\text{elim}} = 0.2$ & $4{,}096$ & $99.7 \pm 0.1$ & $-0.27$ [$-0.38$, $-0.16$] & $4.1$ & $3$ & $39$ & $3.53$ \\
$\theta_{\text{elim}} = 0.3$ & $4{,}096$ & $98.4 \pm 0.3$ & $-1.57$ [$-1.80$, $-1.35$] & $6.7$ & $9$ & $193$ & $4.89$ \\
$\theta_{\text{elim}} = 0.5$ & $617$--$762$ & $85.6 \pm 1.6$ & $-14.18$ [$-15.68$, $-12.69$] & $26.0$ & $44$ & $256$ & $6.21$ \\
\midrule
$\theta_{\text{CLS}}^{\text{eval}} = 0.5$ & $320$--$447$ & $72.7 \pm 4.5$ & $-26.81$ [$-29.41$, $-24.21$] & $75.7$ & $0$ & $300$ & $0.09$ \\
$\theta_{\text{CLS}}^{\text{eval}} = 0.7$ & $10{,}240$ & $99.9 \pm 0.1$ & $-0.04$ [$-0.08$, $+0.01$] & $3.7$ & $5$ & $23$ & $2.21$ \\
$\theta_{\text{CLS}}^{\text{eval}} = 0.8$ & $8{,}600$--$10{,}240$ & $99.8 \pm 0.3$ & $-0.15$ [$-0.20$, $-0.09$] & $3.9$ & $36$ & $24$ & $2.54$ \\
\bottomrule
\end{tabular}}
\end{table}

A lower elimination threshold of $0.05$ removes fewer candidates. It
lowers the unsound rate from $1.76\%$ to $0.67\%$ and raises calls per
solve from $3.9$ to $4.8$, with no measurable change in accuracy.
Higher thresholds remove more candidates, including correct ones. The
unsound rate rises to $3.53\%$, $4.89\%$ and $6.21\%$ at $0.2$, $0.3$
and $0.5$, and accuracy falls by $0.27$, $1.57$ and $14.18$ points.
These are also the only elimination thresholds that produce wrong
answers ($3$, $9$ and $44$ over three seeds).

At $\theta_{\text{CLS}}^{\text{eval}} = 0.5$ the conflict head stops
too many chains that have not reached a contradiction. Accuracy falls
to $72.7\%$, all $300$ failures are timeouts, and a solve takes $75.7$
calls. At $0.7$, calls per solve fall from $3.9$ to $3.7$ and accuracy
does not change measurably, but the solver returns $5$ wrong answers,
all on the seed-$0$ checkpoint. At $0.8$, it returns $36$ wrong answers,
also all on the seed-$0$ checkpoint, and accuracy falls by $0.15$
points. A higher threshold flags fewer states as inconsistent, so
fewer chains are stopped before they reach a complete grid. The default
of $0.6$ is the highest tested value that returns no wrong answers.

\subsection{Final-iteration and averaged logits}
\label{subsec:deduction-readout}

The step operator reads the logits of the last internal iteration. We also
evaluate the same checkpoints with the logits of each head averaged over the
$16$ iterations ($3$ seeds, $10{,}240$ test puzzles per seed).

\begin{table}[htbp]
\centering
\footnotesize
\caption{Logits read by the step operator. $\Delta$ is in percentage points
against the default, paired by puzzle, with a $95\%$ interval.}
\label{tab:deduction-readout}
\vspace{3pt}
\resizebox{\linewidth}{!}{%
\begin{tabular}{lrrrrrr}
\toprule
\textbf{Logits} & \textbf{Acc.\ \%} & $\Delta$ \textbf{(pp)} & \textbf{Calls/solve} & \textbf{Wrong} & \textbf{Timeouts} & \textbf{Unsound \%} \\
\midrule
Final iteration (default) & $\mathbf{99.94 \pm 0.01}$ & ref. & $3.9$ & $0$ & $18$ & $1.76$ \\
Mean over iterations      & $99.79 \pm 0.07$ & $-0.15$ {\scriptsize$[-0.20, -0.09]$} & $6.2$ & $0$ & $64$ & $1.18$ \\
\bottomrule
\end{tabular}}
\end{table}

Averaging lowers accuracy by $0.15$ points, needs $6.2$ instead of $3.9$
calls per solve and times out on $64$ instead of $18$ puzzles; neither
returns a wrong answer. Its share of eliminations that remove a correct
candidate is lower ($1.18\%$ against $1.76\%$).

\section{Carrying Continuous State Between Solve Steps}
\label{app:latent-carry}

LDT passes only the lattice state from one solve step to the next. The
transformer's hidden activations are discarded after every step.
Recurrent reasoners such as TRM instead pass a continuous latent state
between steps. This appendix tests whether adding such a state to LDT
changes its accuracy.

\paragraph{Models} We compare four models. All of them apply the
$4$-layer backbone $16$ times per model call.
\begin{itemize}
\item \emph{Lattice only}: the standard one-stream LDT. Nothing except
  the lattice is passed between steps.
\item \emph{Carry hidden state}: the one-stream model, with the final
  hidden state of the previous step passed through a learned linear map
  and added to the input embedding of the next step.
\item \emph{Carry scratchpad}: a two-stream model. Each model call runs
  $8$ cycles. In each cycle, one backbone application updates a
  scratchpad stream from the embedding of the original puzzle, the
  answer stream and the scratchpad, and a second backbone application
  updates the answer stream from the answer stream and the scratchpad.
  Only the answer stream is read out. The final scratchpad of a step is
  passed through a learned linear map and becomes the initial scratchpad
  of the next step.
\item \emph{Zeroed scratchpad}: the same two-stream model, with the
  scratchpad set to zero at the start of every step. This control
  separates the effect of carrying the scratchpad from the effect of
  the two-stream architecture.
\end{itemize}
The linear maps are initialized to zero. The carried state is reset to
zero when a puzzle enters the training pool and when a solve chain
restarts from the original puzzle after a conflict. During training,
the carried state is computed when the pool is advanced, without
gradients, so no gradient crosses a solve-step boundary. The two-stream
models have $8$ supervised readouts per model call; the one-stream
models have $16$.

\paragraph{Setup} All four models use the ablation protocol of
Appendix~\ref{app:ablations}: the hyperparameters of
Appendix~\ref{app:sudoku-hparams}, $2{,}000$ training steps on the
$1{,}000$ training puzzles, and seeds $0$, $1$ and $2$. A carried vector
is only meaningful if consecutive steps use the same digit labels and
grid orientation. We therefore turn off per-step augmentation, in
training and at inference, for all four models, including the lattice-only
model. The augmentation applied to each puzzle when it enters the
training pool is unchanged. Inference uses the defaults of
Appendix~\ref{app:sudoku-hparams} ($64$ chains per puzzle, $1{,}000$
rounds) without per-step augmentation.

\paragraph{Evaluation} Every model and seed walks the same fixed
ordering of the Sudoku-Extreme test split
(Appendix~\ref{app:eval-protocol}). Evaluation stops after $4{,}096$
puzzles or at the $100$th failure, whichever comes first. The
lattice-only model reaches $4{,}096$ puzzles for every seed. The other
three models reach $100$ failures first, so their evaluations cover
fewer puzzles; Table~\ref{tab:latent-carry} lists the range per seed.
Because the models see different numbers of puzzles, we also report the
difference in accuracy from the lattice-only model on the same puzzles:
for each seed we compare the two models on the puzzles that both
evaluated, and we pool these paired outcomes over seeds.

\begin{table}[htbp]
\centering
\footnotesize
\setlength{\tabcolsep}{3.5pt}
\caption{Carrying continuous state between Sudoku-Extreme solve steps
($2{,}000$ training steps, no per-step augmentation, $3$ seeds).
$n$/seed is the range over seeds of the number of test puzzles
evaluated. Accuracy is the mean $\pm$ standard deviation over seeds.
$\Delta$ is the difference in accuracy from the lattice-only model on the
same puzzles, pooled over seeds, with a $95\%$ interval. ``Calls/solve''
is the number of batched model calls per solved puzzle, averaged over
seeds. ``Wrong'' is the total number of wrong answers over seeds; all
other failures are timeouts. ``Unsound'' is
the percentage of eliminated candidates that are the solution digit,
averaged over seeds.}
\label{tab:latent-carry}
\begin{tabular}{lrrrrrr}
\toprule
\textbf{State carried} & \textbf{$n$/seed} & \textbf{Acc.\ \%}
& \textbf{$\Delta$ (pp)} & \textbf{Calls/solve} & \textbf{Wrong}
& \textbf{Unsound \%} \\
\midrule
Lattice only       & $4{,}096$        & $\mathbf{98.4 \pm 0.1}$ & ---                        & $8.3$   & $1$   & $1.32$ \\
Hidden state       & $262$--$547$     & $71.6 \pm 9.9$ & $-24.7$ [$-27.3$, $-22.2$] & $74.9$  & $0$   & $0.23$ \\
Scratchpad         & $157$--$192$     & $43.3 \pm 6.2$ & $-55.2$ [$-59.4$, $-50.9$] & $182.5$ & $1$   & $0.14$ \\
Zeroed scratchpad  & $425$--$1{,}198$ & $85.8 \pm 8.2$ & $-10.4$ [$-11.7$, $-9.1$]  & $14.3$  & $231$ & $3.93$ \\
\bottomrule
\end{tabular}
\end{table}

\paragraph{Results} Both models that carry continuous state are less
accurate than the lattice-only model. Carrying the hidden state lowers
accuracy by $24.7$ points on the same puzzles, and carrying the
scratchpad lowers it by $55.2$ points. Both models also use more model
calls per solved puzzle: $9\times$ and $22\times$ as many as the
lattice-only model. Almost all of their failures are timeouts ($599$ of
$600$), not wrong answers. Their unsound-elimination rates are lower
than the lattice-only model's.

The zeroed-scratchpad control shows that the two-stream architecture
alone already reduces accuracy, by $10.4$ points, and that it returns
wrong answers: $231$ of the $2{,}552$ puzzles it evaluated, against $1$
of $12{,}288$ for the lattice-only model. Carrying the scratchpad
reduces accuracy further: on the puzzles both evaluated, the carry
scratchpad model is $42.6$ points less accurate than the
zeroed-scratchpad model ($95\%$ interval $[-47.0, -38.2]$).

In this training and search setup, the lattice is a better state to
pass between solve steps than the final hidden activations or a
dedicated scratchpad. The result applies to the two designs tested. It
does not test a scratchpad with the same number of supervised readouts
as the baseline, training with gradients through several solve steps,
or a continuous state that is freely revised as in TRM. LDT's lattice
can only lose candidates from step to step, and a continuous state
would need to be designed to work with that constraint.

\section{Maze-Hard Failure Analysis}
\label{app:maze-failures}

We examine every unsolved maze of the checkpoints reported in
Table~\ref{tab:maze-hard}: each seed's best checkpoint, $3$ seeds each at
$K = 1$ and $K = 512$, $1{,}000$ test mazes per run. Of the $36$ failures,
$3$ are timeouts and $33$ are wrong answers, on $20$ distinct mazes. The
same mazes recur across runs: maze $299$ is answered wrongly by $4$ of the
$6$ runs, and mazes $640$, $706$ and $892$ by $3$.

Almost every wrong answer is a path from start to goal that is longer than
the shortest path (counts summed over $3$ seeds per setting):

\begin{center}
\footnotesize
\begin{tabular}{lrrr}
\toprule
\textbf{Failure} & $K = 1$ & $K = 512$ & \textbf{Total} \\
\midrule
$2$ steps too long                    & $18$ & $4$ & $22$ \\
$4$--$10$ steps too long              & $6$  & $2$ & $8$ \\
$16$ steps too long                   & $1$  & $0$ & $1$ \\
shortest path plus extra marked cells & $1$  & $0$ & $1$ \\
does not reach the goal               & $1$  & $0$ & $1$ \\
\midrule
wrong answers                         & $27$ & $6$ & $33$ \\
timeouts                              & $3$  & $0$ & $3$ \\
\bottomrule
\end{tabular}
\end{center}

\begin{figure}[htbp]
\centering
\includegraphics[width=0.62\linewidth]{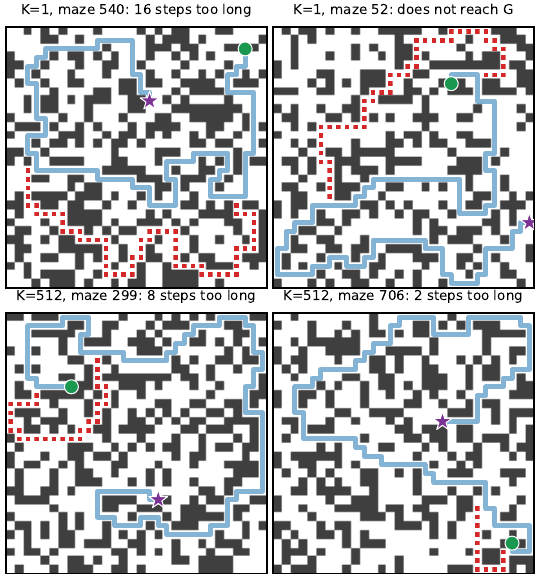}
\caption{Wrong Maze-Hard answers, two each at $K = 1$ and $K = 512$. Blue:
the true shortest path sharing the most cells with the model's answer; red:
the answer's cells off that path. Green circle: start; purple star: goal.}
\label{fig:maze-failures}
\end{figure}

Most of each wrong path lies on some shortest path ($65$ to $110$ of its
cells, typically about $90$). Figure~\ref{fig:maze-failures} shows two examples from each
setting, with the shortest path that shares the most cells with the model's
answer. Whether a path is shortest depends on the whole maze, not on any
single cell, and the conflict head does not flag these answers.

\section{Snowflake Sudoku: Order Transfer and Training Set Size}
\label{app:ood-snowflake}

A single model handles every Snowflake order through the shared covering
grid, so we can test whether it solves larger puzzles than it was trained
on. We train on hard puzzles of orders $4$ to $6$ (all $1{,}308$ training
puzzles of these orders) and evaluate on orders $7$ to $10$ ($768$ test
puzzles), and train on orders $6$ to $8$ ($2{,}000$ puzzles) and evaluate on
orders $9$ and $10$ ($445$ test puzzles). All other settings follow the main
Snowflake recipe (Section~\ref{subsec:snowflake}). We compare LDT with
search, the same LDT model in one shot (one model call, the highest-scoring
digit in every cell), and direct prediction (the same model trained only
with cross-entropy against the solution, evaluated in one shot).

\begin{table}[htbp]
\centering
\footnotesize
\setlength{\tabcolsep}{4pt}
\caption{Snowflake Sudoku order transfer (hard puzzles): accuracy / wrong answers (\%) on
test puzzles of orders not seen in training, mean over $3$ seeds. The
remainder are timeouts; one shot and direct prediction always answer.}
\label{tab:abl-ood}
\begin{tabular}{llccccc}
\toprule
\textbf{Train} & \textbf{Method} & \textbf{7} & \textbf{8} & \textbf{9} & \textbf{10} & \textbf{All} \\
\midrule
4--6 & LDT & 99.27 / 0.73 & 93.91 / 6.09 & 78.36 / 17.38 & 50.00 / 30.91 & 77.21 / 15.80 \\
     & LDT, one shot & 91.24 / 8.76 & 75.63 / 24.37 & 46.60 / 53.40 & 21.65 / 78.35 & 53.99 / 46.01 \\
     & Direct prediction & 81.27 / 18.73 & 51.79 / 48.21 & 16.59 / 83.41 & 3.28 / 96.72 & 32.60 / 67.40 \\
\midrule
6--8 & LDT & --- & --- & 100.00 / 0.00 & 98.86 / 0.43 & 99.40 / 0.22 \\
     & LDT, one shot & --- & --- & 85.94 / 14.06 & 72.51 / 27.49 & 78.88 / 21.12 \\
     & Direct prediction & --- & --- & 85.78 / 14.22 & 71.79 / 28.21 & 78.43 / 21.57 \\
\bottomrule
\end{tabular}
\end{table}

Table~\ref{tab:abl-ood} gives the results by order. Trained on orders $6$
to $8$, LDT transfers to orders $9$ and $10$ almost without loss: it solves
$99.4\%$ with $0.2\%$ wrong answers, and $98.5\%$ of the puzzles that need
at least $8$ guesses. One shot and direct prediction solve $78.9\%$ and
$78.4\%$, and $63.2\%$ and $67.4\%$ of the puzzles that need at least $8$
guesses, so the training objective alone does not explain the difference;
search does. Trained on orders $4$ to $6$, whose puzzles have at most $36$
cells against $60$ at order $10$, every method degrades with distance from
the training orders. LDT stays close to sound one order above the training
range ($0.7\%$ wrong at order $7$), but at order $10$ it solves $50.0\%$ and
gives $30.9\%$ wrong answers; search then finds assignments that the model
accepts but that are wrong. Direct prediction solves $3.3\%$ of order-$10$
puzzles.

\begin{table}[htbp]
\centering
\footnotesize
\setlength{\tabcolsep}{4pt}
\caption{Snowflake Sudoku training set size: accuracy / wrong answers (\%)
on hard test puzzles of orders $4$ to $8$, trained on hard puzzles of the
same orders, mean over $3$ seeds. Training uses 8K steps unless noted.}
\label{tab:snowflake-data}
\begin{tabular}{lccc}
\toprule
\textbf{Training puzzles} & \textbf{LDT} & \textbf{LDT, one shot} & \textbf{Direct prediction} \\
\midrule
$1{,}000$ & 100.00 / 0.00 & 95.52 / 4.48 & 88.87 / 11.13 \\
$2{,}000$ & 100.00 / 0.00 & 96.77 / 3.23 & 97.63 / 2.37 \\
$2{,}000$, 16K steps & 100.00 / 0.00 & 98.42 / 1.58 & 96.05 / 3.95 \\
all $3{,}957$ & 100.00 / 0.00 & 98.29 / 1.71 & 99.28 / 0.72 \\
\bottomrule
\end{tabular}
\end{table}

In distribution (Table~\ref{tab:snowflake-data}), LDT solves every hard
test puzzle with as few as $1{,}000$ training puzzles. Direct prediction
needs more data: $88.9\%$ with $1{,}000$ puzzles, $97.6\%$ with $2{,}000$
and $99.3\%$ with all $3{,}957$. Training for 16K steps instead of 8K
improves one shot but not direct prediction.

\section{Fine-Tuned LLM Baseline on Sudoku-Extreme}
\label{app:llm-finetune}

The LLMs in Appendix~\ref{app:llm-prompts} are evaluated zero-shot. This
appendix measures how a language model performs when it is trained on
Sudoku-Extreme puzzles.

\paragraph{Model and training} We fine-tune all parameters of
Qwen3.5-0.8B, about $1{,}000\times$ the parameter count of LDT, in BF16.
The training set is $1{,}000$ puzzles drawn at random (seed $42$) from
the Sudoku-Extreme training split. LDT's $1{,}000$ training puzzles are
drawn with the same seed after a $10{,}000$-puzzle validation set is
removed from the split, so the two training sets share $16$ puzzles.
There is no data augmentation. The prompt gives the puzzle as an
$81$-character row-major string with \texttt{0} for blank cells and asks
for the solved $81$-digit string with no other text. The target is the
$81$-digit solution, with no reasoning trace, and the loss is computed
on the answer tokens only. Training runs for $16$ epochs ($512$ steps)
with AdamW, learning rate $2 \times 10^{-5}$, weight decay $0.1$, batch
size $32$, a cosine schedule and a linear warmup over the first $10\%$
of steps. Each run uses one seed ($0$).

\paragraph{Puzzle difficulty} We train seven models. One is trained on
the $1{,}000$ natural puzzles. The other six are trained on versions of
the same $1{,}000$ puzzles with exactly $K$ blank cells, for
$K \in \{1, 2, 4, 8, 16, 32\}$. Each such version is made from the
puzzle's solution by blanking $K$ cells chosen uniformly at random; the
chosen cells are fixed for the whole training run and are not limited to
the cells that are blank in the natural puzzle. Natural Sudoku-Extreme
puzzles have about $56$ blank cells and are selected to require long
chains of deduction. Puzzles with randomly placed blanks are much
easier at the same blank count.

\paragraph{Evaluation} Each model is evaluated on the first $256$
puzzles of the fixed test ordering of Appendix~\ref{app:eval-protocol}.
These puzzles have $45$ to $64$ blank cells ($55.8$ on average). The
model trained on $K$ blanks is evaluated on the same $256$ puzzles
rebuilt with $K$ randomly placed blanks. We draw $32$ samples per puzzle
(temperature $0.8$, top-$p$ $0.95$, at most $128$ new tokens). A sample
is correct if, after removing surrounding whitespace, it is exactly the
$81$-digit solution. With $c$ correct samples out of $n = 32$, pass@$k$
for a puzzle is $1 - \binom{n-c}{k} / \binom{n}{k}$, the probability that
at least one of $k$ samples drawn without replacement is correct; we
average it over the $256$ puzzles. Pass@$32$ is the fraction of puzzles
with at least one correct sample. It assumes that the correct sample
can be identified, which the model itself does not do. We evaluate every
second epoch and report the final checkpoint (epoch $16$).

\begin{table}[htbp]
\centering
\footnotesize
\caption{Fine-tuned Qwen3.5-0.8B on Sudoku-Extreme: pass@$k$ (\%) at
epoch $16$ on the first $256$ test puzzles, $32$ samples per puzzle,
$1$ seed. Each row is a separate model, trained on $1{,}000$ training
puzzles and evaluated on test puzzles of the same kind. In the first six
rows, the puzzles are rebuilt from their solutions with exactly $K$
randomly placed blank cells. The last row uses the natural puzzles
($45$--$64$ blanks in the evaluated set).}
\label{tab:llm-finetune}
\begin{tabular}{lrrr}
\toprule
\textbf{Blank cells} & \textbf{pass@1} & \textbf{pass@8} & \textbf{pass@32} \\
\midrule
$K = 1$  (random)   & $98.6$ & $99.5$ & $99.6$ \\
$K = 2$  (random)   & $97.7$ & $98.7$ & $99.2$ \\
$K = 4$  (random)   & $84.1$ & $90.4$ & $91.8$ \\
$K = 8$  (random)   & $78.6$ & $87.7$ & $89.8$ \\
$K = 16$ (random)   & $55.9$ & $68.6$ & $71.9$ \\
$K = 32$ (random)   & $8.7$  & $14.4$ & $16.4$ \\
\midrule
Natural puzzles     & $0.0$  & $0.0$  & $0.0$ \\
\bottomrule
\end{tabular}
\end{table}

\paragraph{Results} With $1$ or $2$ random blanks the fine-tuned model
solves almost every puzzle ($99.6\%$ and $99.2\%$ pass@$32$), so the
training and evaluation pipeline works. Accuracy falls as blanks are
added: pass@$32$ is $91.8\%$ at $4$ blanks, $71.9\%$ at $16$ and
$16.4\%$ at $32$. On the natural puzzles the model solves none of the
$256$ puzzles at epoch $16$. Over all eight evaluated checkpoints
(epochs $2, 4, \ldots, 16$), it solves one puzzle in one checkpoint
(epoch $10$) and none in the others. Its output format also degrades
with training: the share of samples that are not a single $81$-digit
string falls from $32\%$ at epoch $2$ to $0.6\%$ at epoch $6$, then
rises to $55\%$ at epoch $16$.

For comparison, LDT trained on $1{,}000$ natural puzzles without
augmentation solves $96.54\%$ of natural test puzzles, with one answer
per puzzle (Table~\ref{tab:Sudoku-extreme}).

\section{Frontier LLM Baselines}
\label{app:llm-prompts}

We evaluate frontier chain-of-thought LLMs zero-shot with the prompts below.

\paragraph{April 2026} Claude Opus 4.6~\cite{Anthropic2026Claude46}, DeepSeek
V4-Pro~\cite{DeepSeek2026V4Release} and GPT-5.4~\cite{OpenAI2026GPT54} were
each evaluated on $200$ test puzzles per benchmark and solved none of them.
Their Snowflake puzzles come from an earlier version of the generator
(\url{https://github.com/santolucito/snowflakesudoku}), with one fixed layout
per order and including puzzles with givens added back, not the puzzles of
Section~\ref{subsec:snowflake}.

\begin{center}
\footnotesize
\begin{tabular}{lrrr}
\toprule
\textbf{Model (April 2026)} & \textbf{Sudoku-Extreme} & \textbf{Snowflake} & \textbf{Maze-Hard} \\
\midrule
Claude Opus 4.6 & 0.0 & 0.0 & 0.0 \\
DeepSeek V4-Pro & 0.0 & 0.0 & 0.0 \\
GPT-5.4         & 0.0 & 0.0 & 0.0 \\
\bottomrule
\end{tabular}

\smallskip
Solve rate (\%) on $200$ test puzzles per benchmark.
\end{center}

\paragraph{October 2026} We evaluate Claude Opus 5.5, GPT-6 Astra and GPT-6.1
Sol through OpenRouter on the first $50$ puzzles of each benchmark's test
ordering: Sudoku-Extreme, the hard Snowflake puzzles of
Section~\ref{subsec:snowflake} and Maze-Hard. Each call uses reasoning effort ``high'', at most $100{,}000$
output tokens and at most one hour; an answer that reaches a cap, or a call
that ends in a provider error, counts as unsolved. Sudoku and Snowflake
answers must equal the unique solution; a Maze answer must be a valid
$30 \times 30$ grid with walls, start and goal unchanged and a shortest path
from start to goal. Table~\ref{tab:llm-current} reports the solve rate and
the mean output tokens and cost per puzzle.

\begin{table}[htbp]
\centering
\footnotesize
\caption{Frontier LLMs in October 2026, zero-shot, on the first $50$ test
puzzles of each benchmark: solve rate (\%), mean output tokens per puzzle
and mean cost per puzzle (US dollars, OpenRouter prices).}
\label{tab:llm-current}
\vspace{3pt}
\resizebox{\linewidth}{!}{%
\begin{tabular}{lrrrrrrrrr}
\toprule
& \multicolumn{3}{c}{\textbf{Sudoku-Extreme}} & \multicolumn{3}{c}{\textbf{Snowflake (hard)}} & \multicolumn{3}{c}{\textbf{Maze-Hard}} \\
\cmidrule(lr){2-4}\cmidrule(lr){5-7}\cmidrule(lr){8-10}
\textbf{Model} & \textbf{Solved \%} & \textbf{Tokens} & \textbf{\$} & \textbf{Solved \%} & \textbf{Tokens} & \textbf{\$} & \textbf{Solved \%} & \textbf{Tokens} & \textbf{\$} \\
\midrule
Claude Opus 5.5 & 54 & 50{,}794 & 1.02 & 100 & 3{,}797 & 0.08 & 26 & 14{,}319 & 0.29 \\
GPT-6 Astra     & 68 & 28{,}336 & 1.42 & 100 & 2{,}303 & 0.13 & 26 & 11{,}906 & 0.60 \\
GPT-6.1 Sol     & 44 & 18{,}797 & 0.19 & 96  & 2{,}447 & 0.03 & 12 & 11{,}457 & 0.12 \\
\bottomrule
\end{tabular}}
\end{table}

\paragraph{Observed failures} Hard Snowflake puzzles are solved almost
always, in under a minute and a few thousand output tokens; they are much
easier for these models than Sudoku-Extreme. On Sudoku-Extreme, many
failures return no answer: Claude Opus 5.5 reaches the output-token cap on
$14$ puzzles, and GPT-6 Astra and GPT-6.1 Sol end in a provider error on $6$
and $11$ puzzles, after up to $52$ minutes. Of the answers returned, most
wrong ones break a row, column or box constraint ($12$ for GPT-6.1 Sol, $4$
for GPT-6 Astra, $3$ for Claude Opus 5.5); a few change a given or are valid
grids that differ from the solution. On Maze-Hard, the most common failure is
a path from start to goal that is longer than the shortest path ($14$ to
$21$ per model). The next most common is a malformed grid ($16$ to $18$ per
model), almost always a row of $29$ characters instead of $30$. A few answers do not connect start and goal or
pass through a wall.

\subsection{Sudoku-Extreme}
\label{subsec:llm-sudoku-extreme}

\paragraph{System prompt}
\begin{quote}\ttfamily\small
You are a Sudoku solver. You will be given a $9 \times 9$ Sudoku
puzzle. Solve it and respond ONLY with the completed $9 \times 9$
grid, one row per line, digits only, no spaces or other characters.
Example response format:
\begin{verbatim}
534678912
672195348
...
345286179
\end{verbatim}
\end{quote}

\paragraph{User prompt}
\begin{quote}\ttfamily\small
Solve this Sudoku:\\[2pt]
\{puzzle\_text\}\\[2pt]
(\{puzzle\_text\} is the puzzle as 9 lines of 9 characters each,
with empty cells written as 0.)
\end{quote}

\subsection{Snowflake Sudoku}
\label{subsec:llm-snowflake}

\paragraph{System prompt}
\begin{quote}\ttfamily\small
You are solving a Snowflake Sudoku puzzle, a Sudoku variant played on
a snowflake-shaped topology of $N$ cells, each holding a digit from
$1$ to $6$. The puzzle is defined by a list of constraint groups;
each group is a set of cell IDs whose digits must all be distinct.
You will be given (1) the total number of cells, (2) the list of
constraint groups, and (3) the current cell values, where \texttt{?}
marks a cell you must solve and an integer $1$--$6$ marks a given.
Respond with ONLY the solution: one line per cell in order, formatted
as \texttt{CELL\_ID:DIGIT} with no spaces, comments, or other text.
Output exactly $N$ lines. Example:
\begin{verbatim}
0:3
1:5
2:1
3:6
4:2
5:4
\end{verbatim}
\end{quote}

\paragraph{User prompt}
\begin{quote}\ttfamily\small
Solve this Snowflake Sudoku:\\[2pt]
\{puzzle\_text\}\\[2pt]
(\{puzzle\_text\} encodes the puzzle size, the constraint topology,
and the current cell values, e.g.:)\\[4pt]
Puzzle (n=4, 24 cells):

Constraint groups (each group must contain distinct digits):\\
\hspace*{1em} Group 0: [0, 1, 2, 3, 4, 5]\\
\hspace*{1em} Group 1: [6, 7, 8, 9, 10, 11]\\
\hspace*{1em} Group 2: [12, 13, 14, 15, 16, 17]\\
\hspace*{1em} ...

Cell values (? = solve this):\\
\hspace*{1em} Cell 0: 3\\
\hspace*{1em} Cell 1: ?\\
\hspace*{1em} Cell 2: ?\\
\hspace*{1em} Cell 3: 6\\
\hspace*{1em} ...
\end{quote}

\subsection{Maze-Hard}
\label{subsec:llm-maze}

\paragraph{System prompt}
\begin{quote}\ttfamily\small
You are a maze solver. You will be given a $30 \times 30$ maze.
Every cell of the maze is exactly one of the following four
characters:
\begin{verbatim}
'#' - wall (impassable)
' ' - free cell
'S' - the start cell (exactly one)
'G' - the goal cell (exactly one)
\end{verbatim}
Find a path of orthogonally adjacent (up/down/left/right) free cells
that connects \texttt{S} to \texttt{G} without passing through any
wall. Diagonal moves are not allowed. Mark every free cell that lies
on the path with the character \texttt{o}. The cells \texttt{\#},
\texttt{S}, and \texttt{G} are unchanged, and free cells that are
not on the path remain \texttt{' '}.
Respond with ONLY the completed $30 \times 30$ grid: exactly $30$
lines, each containing exactly $30$ characters drawn from the set
\texttt{\{\#, ' ', S, G, o\}}. Do not include row numbers, column
markers, blank lines, code fences, prose, or any other text.
Whitespace inside each row is significant: a leading or trailing
space is itself a free cell, not formatting. Example response format
(shown for a $6 \times 6$ maze for brevity; your output must be $30$
rows of $30$ characters):
\begin{verbatim}
######
#Sooo#
#  #o#
#  #o#
## ooG
######
\end{verbatim}
\end{quote}

\paragraph{User prompt}
\begin{quote}\ttfamily\small
Solve this maze:\\[2pt]
\{puzzle\_text\}\\[2pt]
(\{puzzle\_text\} is the maze as 30 lines of 30 characters each,
drawn from \{\#, ' ', S, G\}.)
\end{quote}

\section{Snowflake Sudoku: Data and Encoding}
\label{app:snowflake-diagram}
\label{app:snowflake-data}

We generate puzzles using CVC5~\cite{DBLP:conf/tacas/BarbosaBBKLMMMN22}.
Every puzzle gets its own randomly grown, connected layout of order $n$; CVC5
finds a valid complete assignment, and a greedy minimization pass removes
givens while CVC5 verifies that the puzzle retains a unique solution, so the
puzzle is minimal. We rate each puzzle by the number of guesses a simple
solver needs: it applies naked and hidden singles until nothing changes, then
branches on a cell with the fewest candidates. A rating of $0$ means singles
alone solve the puzzle, which holds for $41\%$ (order $8$) to $96\%$ (order
$4$) of minimal puzzles. We keep only puzzles with rating at least $1$; the
median test puzzle needs $3$ guesses. Puzzles of orders $4$ to $10$ are split
into test ($10\%$), validation ($5\%$) and training sets, giving $7{,}801$
training, $476$ validation and $951$ test puzzles. Small orders are rare
because singles solve almost all of them: orders $4$ to $8$ have $3{,}957$
training and $506$ test puzzles, of which $19$ are of order $4$. Unless
noted, we train on $2{,}000$ puzzles of orders $4$ to $8$, and orders $9$ and
$10$ are held out for the order-transfer experiments in
Appendix~\ref{app:ood-snowflake}.

To handle the variable topology with a fixed-size transformer, we
embed every puzzle into a fixed covering grid of $25$ hexagon positions
($5 \times 5$ in axial coordinates, $150$ cells); layouts that do not fit
are regrown. Each hexagonal cell
maps to a deterministic position via axial coordinates $(q, r)$ and
triangular direction, with cells absent from a given puzzle zeroed
out and excluded from the loss via a per-cell mask; during training each
puzzle is translated by a random whole number of hexes within the grid.
Because every layout differs, a cell's position no longer determines its
constraint groups, so each cell also receives a multi-hot indicator of the
groups it belongs to. Group identifiers are reassigned at random for every
sample, so the model can use shared membership but cannot tie an
identifier to a location. A single model thus handles all puzzle sizes
without architecture changes. Training and inference follow the same solve
procedure as Sudoku-Extreme, with 8K training steps; unlike Sudoku-Extreme,
training also uses per-step augmentation and the random hex translation
(Appendix~\ref{app:snowflake-hparams}). Each snowflake puzzle
has a unique solution, so $\alpha$ runs at $K = 1$. Full hyperparameters are
in Appendix~\ref{app:snowflake-hparams}. TRM, HRM, and Sotaku are not
trained on snowflake Sudoku and are omitted.

\section{Hyperparameters}
\label{app:hparams}

\paragraph{Augmentation} Per-step augmentation wraps each solve step in a
random symmetry of the problem and inverts it after deduction and branching:
digit permutation composed with a dihedral grid symmetry for Sudoku, digit
permutation for Snowflake (the dihedral group does not preserve the hex
topology), and dihedral symmetries for Maze. All three benchmarks use it at
evaluation; Snowflake and Maze also use it in training, while for Sudoku it
did not help in training. Sudoku instead augments at the dataset level,
Snowflake translates each puzzle by a random whole number of hexes within the
covering grid, and Maze applies a random dihedral symmetry and a random swap
of start and goal to each training sample. The models in
Figure~\ref{fig:sudoku-train-test-compute}(a, b) were trained with per-step
augmentation, unlike the Sudoku rows of Table~\ref{tab:Sudoku-extreme};
panel (c) uses the portfolio models, which are trained as in
Table~\ref{tab:Sudoku-extreme}.

\subsection{Sudoku-Extreme}
\label{app:sudoku-hparams}

These settings are used for Sudoku-Extreme
(Section~\ref{subsec:Sudoku-extreme}) and are the base for the other
benchmarks.

\paragraph{Architecture} Looped transformer with embedding dimension
$d = 128$, $4$ layers, $4$ attention heads, $L = 16$ internal loops per
forward pass, FFN multiplier $4.0$, dropout $p = 0.1$ during training,
and learned 2D positional embeddings.
Total parameter count is $\approx 800{,}000$. Sudoku-Extreme uses
$9$ input channels per cell, one candidate sigmoid per digit.

\paragraph{Optimizer} AdamW with learning rate $3 \times 10^{-3}$,
weight decay $0.1$, $\beta = (0.9, 0.95)$, gradient clip $1.0$, cosine
schedule with a linear warmup over the first $10\%$ of total training
steps.

\paragraph{Solve-procedure training}
Batch size $B = 512$ with pool-to-batch multiplier $1$ (the pool
holds exactly one batch of states, all of which is consumed per step). Sudoku-Extreme is
trained for $T = 4{,}000$ steps over the $1{,}000$-puzzle train split
($\approx 2{,}048$ epochs; $T \in \{1{,}000, 2{,}000\}$ also reported
in Table~\ref{tab:Sudoku-extreme}, at $\approx 512$ and
$\approx 1{,}024$ epochs respectively). Maximum pool age
$\tau_{\text{age}} = 100$ training steps since insertion. Pool states are
advanced with elimination threshold $\theta_{\text{elim}} = 0.1$, conflict
threshold $\theta_{\text{CLS}}^{\text{train}} = 0.5$ and digits sampled at
decide temperature $\tau_{\text{decide}} = 1.5$. For Sudoku,
each puzzle is wrapped in a random digit permutation composed with a dihedral
grid symmetry before it enters the training pool, and training uses no
per-step augmentation; all Sudoku-Extreme rows of
Table~\ref{tab:Sudoku-extreme} except the no-augmentation row use this.
Training seeds $0$, $1$ and $2$ ($0$ to $14$ for the portfolio models).

\paragraph{No-augmentation variant} The no-aug row of
Table~\ref{tab:Sudoku-extreme} drops both augmentation channels:
training sees only the raw $1{,}000$-puzzle Sudoku-Extreme split with
no dataset-level expansion and no per-step augmentation. Inference keeps the
per-step augmentation below; without it, the model solves $31.6\%$ (mean over seeds). All other hyperparameters match the base
set.

\paragraph{Loss weights}
BCE class weights $w^{+} = 4.0$ on positive candidates and $w^{-} = 0.5$
on negatives; auxiliary softmax cross-entropy weight
$\lambda_{\text{ce}} = 0.2$; CLS BCE weight
$\lambda_{\text{cls}} = 0.1$.

\paragraph{Inference defaults (Section~\ref{subsec:step-and-solve})}
At inference we batch the parallel solve along two axes. A
\emph{slot} holds one puzzle from the moment it enters the batch
until that puzzle either self-accepts (a chain reaches a complete,
operator-accepted solution) or hits the per-puzzle round budget; we
run $M = 8$ slots concurrently, evicting and refilling slots from a
queue of remaining test puzzles. Within each slot, $H = 64$
\emph{chains} explore the same starting state in parallel but
diverge under the random choice of branching cell, per-step
augmentation and eval-time dropout -- effectively $H$ independent
restarts of the search for the same puzzle. One forward pass therefore
spans $M \cdot H = 512$ rows. Per-puzzle round budget $R = 1{,}000$,
elimination threshold $\theta_{\text{elim}} = 0.1$, eval-time dropout
$p_{\text{drop}} = 0.05$, conflict threshold
$\theta_{\text{CLS}}^{\text{eval}} = 0.6$, tuned on the
Sudoku-Extreme train set. The branching cell is drawn uniformly at random
among cells with at least two candidates, and the digit is the argmax of
the digit logits $s$ ($\tau_{\text{decide}}$ is used only in training).
We additionally apply \emph{per-step
augmentation} at inference: at each Solve step the lattice state is
wrapped in a fresh random digit-permutation composed with a dihedral
symmetry, the model performs deduction and branching in that
augmented frame, and we invert the symmetry before writing back to
the canonical state. This decorrelates the parallel chains and broadens the inference
search.

\paragraph{Sequential-cost estimation}
Figure~\ref{fig:sudoku-train-test-compute}(a, b) reports per-puzzle
forward-pass counts as a sequential (batch-1) search would have paid,
even though inference is run with the parallel slot/chain batch
described above. Within a slot, we number \emph{attempts} (a chain's run
from the original puzzle until it is accepted, or until a conflict
restarts it) in the order they start. Let attempt $w$ be the first to
reach an accepted solution. A sequential search would have run attempts
$0, \ldots, w$ one after another, so the cost we report is
$(w + 1)\,\bar{\ell}$, where $\bar{\ell}$ is the mean length in rounds of
the attempts that ended in that slot. To observe the attempts numbered
below $w$ that are still running when $w$ succeeds, we \emph{drain} the
slot after the winner: it keeps running until each of them has ended
(for at most $200$ further rounds). Puzzles answered wrongly or not
solved within the budget are charged the upper bound $R \cdot H$. This
yields sequential-cost estimates without re-running the solve at
batch $1$.

\subsection{Snowflake Sudoku}
\label{app:snowflake-hparams}

For Snowflake Sudoku (Section~\ref{subsec:snowflake}) we keep the
Sudoku-Extreme settings above except for the following changes. Each
puzzle is embedded in the fixed $150$-cell covering grid
(Appendix~\ref{app:snowflake-data}) with $50$ channels per cell: $6$
candidate sigmoids, a read-only \emph{in-puzzle} mask, and $43$ read-only
constraint-group channels; the deduction operator edits only the $6$
candidate channels ($\approx 816{,}000$ parameters). Training runs for
$T = 8{,}000$ steps on $2{,}000$ puzzles of the training orders
($1{,}000$ for the 1K column of Table~\ref{tab:snowflake}), with per-step
digit-permutation augmentation in training and at evaluation and a random
whole-hex translation of every training sample; there is no dihedral
augmentation. Direct prediction uses the same model without the CLS
token, trained only with the cross-entropy loss ($\lambda_{\text{ce}} = 1$,
$w^{+} = w^{-} = \lambda_{\text{cls}} = 0$) on puzzles as given, each used for a
single step (pool age $0$), and is evaluated in one shot without dropout.

\subsection{Maze}
\label{app:maze-hparams}

For Maze (Section~\ref{subsec:maze}) we keep the Sudoku-Extreme settings
above except for the following changes.

\paragraph{General} Augmentation is dihedral only --
digit-permutation is disabled because the channels (wall, free,
start, goal, path) carry distinct semantics. Besides per-step
augmentation in training and at evaluation, each training sample gets a
random dihedral symmetry and a random swap of start and goal.
Elimination threshold $\theta_{\text{elim}} = 0.1$ in training and at
evaluation.

\paragraph{$15 \times 15$ setting} Used for the multi-solution
supervision sweep (Figure~\ref{fig:maze-k-sweep-loss}). Embedding
dimension $d = 128$ as in the base set, batch size
$B = 256$, total training steps $T = 4{,}000$ ($\approx 102$ epochs
over a $10{,}000$-puzzle synthetic training set), inference conflict
threshold $\theta_{\text{CLS}}^{\text{eval}} = 0.6$ (matching the
base set), $M = 4$ slots of $64$ chains, and
$K \in \{1, 8, 32, 64, 128, 256, 512\}$ swept across runs. Each $K$
value is averaged over $9$ training seeds ($0$ to $8$); all seeds share
one procedurally generated training set (data seed $42$), so the seed
controls initialization and training randomness only.

\paragraph{$30 \times 30$ setting} Used for the headline Maze-Hard
results (Table~\ref{tab:maze-hard}). Embedding dimension $d = 192$
($\approx 1.8$M parameters total), with 2D RoPE inside attention in place
of the base set's learned 2D positional embedding. Batch size
$B = 192$, total training steps $T = 20{,}000$, pool-to-batch
multiplier $2.0$, eval-time dropout $p_{\text{drop}} = 0$ (vs.\
$0.05$ elsewhere), one slot ($M = 1$) of $64$ chains, inference conflict
threshold $\theta_{\text{CLS}}^{\text{eval}} = 0.53$, calibrated on a
held-out Maze-Hard set. We report two $K$ configurations, both trained on
the $1{,}000$-puzzle Maze-Hard split for $T = 20{,}000$ steps
($\approx 3{,}840$ epochs): $K = 1$ (head-to-head with TRM) and
$K = 512$.

\section{Evaluation Protocol}
\label{app:eval-protocol}

Every Sudoku-Extreme test evaluation walks one fixed random permutation of the
$422{,}786$ test puzzles (generated from a recorded seed and checked against
a recorded SHA-256), so every model and seed sees the same puzzles in the
same order. A puzzle is a timeout if no chain returns a solution within the
round budget ($1{,}000$ rounds unless noted), and wrong if the returned grid
differs from the solution. Full-split evaluations cover every test puzzle.
Sequential evaluations stop at $100$ failures (wrong answers or timeouts),
or, from $1{,}000$ puzzles on and checked every $100$ puzzles, once the
half-width of a $95\%$ confidence sequence for the accuracy is at most
$0.1$ percentage points. A confidence sequence holds uniformly over the
number of puzzles evaluated, so the interval stays valid under this
data-dependent stopping. Snowflake Sudoku and Maze-Hard evaluations cover
every test puzzle.

In Table~\ref{tab:Sudoku-extreme}, the 4K rows with training augmentation are
evaluated on all test puzzles and the others sequentially; HRM, TRM and Sotaku v2 are
evaluated on the first $10{,}000$ puzzles of the same ordering. The number of puzzles
evaluated per row, summed over seeds:

\begin{center}
\footnotesize
\begin{tabular}{lr}
\toprule
\textbf{Row} & \textbf{Puzzles evaluated} \\
\midrule
HRM, TRM, Sotaku v2 (each)              & $10{,}000$ \\
LDT, 1K train steps                     & $58{,}556$ \\
LDT, 2K train steps                     & $26{,}900$ \\
LDT, 4K train steps                     & $3 \times 422{,}786$ \\
\quad 10{,}000 search rounds            & $3 \times 376$ (reruns) \\
\quad portfolios                        & $15 \times 422{,}786$ plus reruns \\
LDT, 4K, 1K puzzles, no training augmentation & $12{,}914$ \\
\bottomrule
\end{tabular}
\end{center}

The $10{,}000$-round row reruns the 4K-step models' timeouts at $1{,}000$
rounds on the $376$ test puzzles that any of the $15$ portfolio models left
unsolved.

For Maze-Hard (Table~\ref{tab:maze-hard}), every method is evaluated on all
$1{,}000$ test mazes. LDT keeps a snapshot every $1{,}000$ training steps; we
evaluate them at $\theta_{\text{elim}} = 0.1$ and report the best per seed.
For HRM and TRM we evaluate every saved checkpoint of our retrains and
report the best; their exact-match accuracies (path identical to the
dataset's reference path) are $16.7\%$ (HRM, trained with augmentation) and
$84.5\%$ (TRM). Training compute in
Table~\ref{tab:maze-hard} counts each method up to its reported checkpoint;
for LDT, the best snapshots are at $12$K, $14$K and $18$K steps ($K = 1$)
and at $20$K, $18$K and $18$K steps ($K = 512$), and we report the mean over
seeds.

\section{Compute Accounting}
\label{app:compute}

\paragraph{B200 time} All times are on a single NVIDIA B200. LDT training
time is the measured training loop after compilation, excluding periodic
monitoring solves. We did not train HRM or TRM on Sudoku-Extreme. For their
training time we run each released checkpoint's published training
configuration for $12$ minutes on one B200, read the steady-state time per
step, and multiply by the checkpoint's step count:

\begin{center}
\footnotesize
\begin{tabular}{lrrr}
\toprule
\textbf{Model} & \textbf{s / step} & \textbf{Steps} & \textbf{B200 hours} \\
\midrule
HRM             & 0.059 & 52{,}083 & 0.85 \\
TRM (MLP)       & 0.524 & 32{,}550 & 4.7 \\
EqR             & 0.036 & 50{,}000 & 0.50 \\
Sotaku v2       & 0.294 & 50{,}000 & 4.1 \\
\bottomrule
\end{tabular}
\end{center}

HRM's step count is its $40{,}000$ epochs over $1{,}000$ puzzle groups at
global batch $768$; the TRM checkpoint was saved at step $32{,}550$ of a run
at global batch $1{,}536$. HRM's release does not record its step, so its
count is an upper bound. EqR
shows no training progress without its logging service, so we time it as the
difference between runs capped at $200$ and $1{,}200$ steps; Sotaku v2 is
timed the same way on its $50$K reference recipe (steps $200$ and $900$,
after an untimed run that fills its data caches). For
Maze-Hard we retrained HRM and TRM ourselves and derive their time to the
reported checkpoint the same way, from measured throughputs of $0.61$ (HRM)
and $1.46$ (TRM) seconds per step. These are the costs of the published
recipes on our hardware; none of the methods, including ours, is tuned for
B200 utilization. Inference cost is the evaluation's GPU time divided by the
number of puzzles evaluated.

\begin{center}
\footnotesize
\begin{tabular}{llrr}
\toprule
\textbf{Benchmark} & \textbf{Method} & \textbf{Training} & \textbf{Inference (s / puzzle)} \\
\midrule
Sudoku-Extreme & HRM                        & 51m$^\dagger$   & 0.014 \\
               & TRM, MLP                   & 4.7h$^\dagger$  & 0.011 \\
               & Sotaku v2                  & 4.1h$^\dagger$  & 0.0091 \\
               & EqR                        & 30m$^\dagger$   & 0.61 \\
               & LDT, 1K train steps        & 3m              & 0.26 \\
               & LDT, 2K train steps        & 7m              & 0.094 \\
               & LDT, 4K train steps        & 13m             & 0.042 \\
               & \quad $10{,}000$ search rounds & 13m         & 0.044 \\
               & \quad no training augmentation & 11m         & 0.024 \\
\midrule
Maze-Hard      & HRM                        & 20.7h$^\dagger$ & --- \\
               & TRM                        & 7.9h$^\dagger$  & --- \\
               & LDT, $K = 1$               & 6.9h            & --- \\
               & LDT, $K = 512$             & 9.1h            & --- \\
\bottomrule
\end{tabular}

{\scriptsize $^\dagger$Estimated from measured throughput; LDT times are measured.}
\end{center}

\paragraph{Inference FLOPs}
We count a multiply-add as $2$ FLOPs and include, for every layer
application, the attention projections ($4d^2$ per token), the attention
scores and values ($2t^2d$ over $t$ tokens) and the feed-forward block;
embeddings, normalization and output heads are omitted (under $2\%$ at these
sizes). One LDT model call runs $L = 16$ iterations of $4$ layers ($d = 128$,
$81$ cells plus the CLS token) for each of the $64$ chains of a puzzle; the
cost per puzzle is the mean number of model calls per puzzle times $64$ times
the cost of one chain's call. HRM and TRM run their full $16$ ACT steps at
evaluation, each of $24$ (HRM) or $42$ (TRM) layer applications at $d = 512$
with $16$ (TRM) or $1$ (HRM) extra puzzle-embedding positions; TRM's MLP
variant replaces attention with a SwiGLU across positions. EqR uses TRM's
MLP backbone and, at its reported setting, runs $64$ ACT steps for each of
$128$ restarts. Sotaku v2 runs $1{,}024$ iterations of $4$ layers ($d = 128$,
feed-forward width $512$). These are estimates of arithmetic, not
measurements, and ignore early stopping: LDT stops when a puzzle is solved,
which the mean call count reflects, while the others run their full budget.

\begin{center}
\footnotesize
\begin{tabular}{lrr}
\toprule
\textbf{Method (setting)} & \textbf{GFLOPs per step} & \textbf{TFLOPs per puzzle} \\
\midrule
Sotaku v2 ($1{,}024$ iterations)                 & $0.14$ / iteration & $0.14$ \\
HRM ($16$ ACT steps)                             & $13.7$ / ACT step  & $0.22$ \\
TRM, MLP ($16$ ACT steps)                        & $25.6$ / ACT step  & $0.41$ \\
EqR (depth $64$, breadth $128$)                  & $25.6$ / ACT step  & $210$ \\
LDT, 1K train steps ($65.4$ calls $\times$ $64$ chains) & $2.28$ / chain call & $9.6$ \\
LDT, 2K train steps ($22.2$ calls)               & $2.28$ / chain call & $3.2$ \\
LDT, 4K train steps ($10.1$ calls)               & $2.28$ / chain call & $1.5$ \\
\bottomrule
\end{tabular}
\end{center}

\paragraph{Training FLOPs}
A training step of LDT costs $4$ forward passes per sample: the forward and
backward pass through all $16$ iterations ($3$ forward-pass equivalents) and
the no-grad forward that advances the pool states. HRM and TRM backpropagate
only through the last of their recursion cycles: TRM runs $3$ H-cycles and
backpropagates through one ($1\tfrac{2}{3}$ forward-pass equivalents per
sample); HRM backpropagates through its last L- and H-module calls, $8$ of
its $24$ layer applications, and runs one more no-grad forward for its
halting target ($2\tfrac{2}{3}$). Sotaku v2 trains $16$ iterations with
gradients, preceded on $20\%$ of batches by $32$ to $512$ iterations without.
We do not count LDT's periodic monitoring solves, which change no weights;
they would add up to about $30\%$ to its training FLOPs. Model FLOPs
utilization (MFU) is the training FLOPs divided by the measured or estimated
B200 time, relative to the B200's dense BF16 peak of $2.25$ PFLOP/s.

\begin{center}
\footnotesize
\begin{tabular}{lrrrr}
\toprule
\textbf{Method} & \textbf{Steps} & \textbf{Batch} & \textbf{PFLOPs} & \textbf{MFU} \\
\midrule
HRM               & $52{,}083$ & $768$   & $1{,}466$ & $21\%$ \\
TRM, MLP          & $32{,}550$ & $1{,}536$ & $2{,}136$ & $5.6\%$ \\
EqR               & $50{,}000$ & $768$   & $1{,}640$ & $40\%$ \\
Sotaku v2         & $50{,}000$ & $2{,}048$ & $1{,}265$ & $3.8\%$ \\
LDT, 1K train steps & $1{,}000$ & $512$  & $4.7$ & $1.0\%$ \\
LDT, 2K train steps & $2{,}000$ & $512$  & $9.4$ & $1.1\%$ \\
LDT, 4K train steps & $4{,}000$ & $512$  & $18.7$ & $1.1\%$ \\
\bottomrule
\end{tabular}
\end{center}

LDT trains with $78$ to $114$ times fewer FLOPs than HRM and TRM, but its
small model uses the GPU far less fully, so its advantage in B200 time
(Table~\ref{tab:Sudoku-extreme}) is smaller.
EqR uses TRM's architecture and the same one-step gradient, so it costs
the same FLOPs per step, yet its released code runs about $7$ times as fast
on the B200 as the TRM implementation we measured; MFU depends as
much on the implementation as on the method.

\end{document}